\documentclass{ecai}
\usepackage{subfigure}
\usepackage{graphicx}
\usepackage{latexsym}
\usepackage{multirow}
\usepackage{mathtools}
\usepackage{amssymb}
\usepackage{booktabs}
\usepackage[linesnumbered, ruled]{algorithm2e}

\usepackage{caption2}

\usepackage{xcolor}
\usepackage{xspace}
\usepackage{url}
\usepackage{longtable}
\usepackage{hhline}
\usepackage{color}
\usepackage{listings}
\usepackage[ruled]{algorithm2e}

\definecolor{myblue}{HTML}{0072C6}
\definecolor{myyellow}{HTML}{FFFADF}
\definecolor{myred}{HTML}{FF0000}

\begin{document}

\begin{frontmatter}

\title{Diffusion Model for Camouflaged Object Detection}  

\author[A]{\fnms{Zhennan}~\snm{Chen}}
\author[B]{\fnms{Rongrong}~\snm{Gao}}
\author[C{;$\ast$}]{\fnms{Tian-Zhu}~\snm{Xiang}}
\author[A]{\fnms{Fan}~\snm{Lin}\thanks{Corresponding Authors. Email: tianzhu.xiang19@gmail.com; 	iamafan@xmu.edu.cn.}}


\address[A]{School of Informatics, Xiamen University, Xiamen, China} 
\address[B]{Department of Computer Science and Engineering, HKUST, Hong Kong, China}
\address[C]{G42, Abu Dhabi, UAE}

\begin{abstract}
Camouflaged object detection is a challenging task that aims to identify objects that are highly similar to their background. Due to the powerful noise-to-image denoising capability of denoising diffusion models, in this paper, we propose a diffusion-based framework for camouflaged object detection, termed diffCOD, a new framework that considers the camouflaged object segmentation task as a denoising diffusion process from noisy masks to object masks. Specifically, the object mask diffuses from the ground-truth masks to a random distribution, and the designed model learns to reverse this noising process. To strengthen the denoising learning, the input image prior is encoded and integrated into the denoising diffusion model to guide the diffusion process. Furthermore, we design an injection attention module (IAM) to interact conditional semantic features extracted from the image with the diffusion noise embedding via the cross-attention mechanism to enhance denoising learning. Extensive experiments on four widely used COD benchmark datasets demonstrate that the proposed method achieves favorable performance compared to the existing 11 state-of-the-art methods, especially in the detailed texture segmentation of camouflaged objects. Our code will be made publicly available at: \textcolor{blue}{{https://github.com/ZNan-Chen/diffCOD}}.

\end{abstract}

\end{frontmatter}

\section{Introduction}
Camouflage is to use any combination of coloration, illumination, or materials to hide organisms in their surroundings, or disguise them as something else, for deception and paralysis purposes. Camouflaged object detection (COD)~\cite{fan2020camouflaged}, that is, segmenting camouflaged objects from the background, is a challenging vision topic that has emerged in recent years, due to the high similarity of camouflaged objects to the background. COD has also attracted growing research interest from the computer vision community, because of its wide range of real-world applications, such as agricultural pest detection~\cite{kumar2021early}, medical image segmentation~\cite{li2022trichomonas}, and industrial defect detection~\cite{tabernik2020segmentation}. 


With the advent of large-scale camouflaged object detection datasets in recent years, such as CAMO~\cite{le2019anabranch} and COD10K~\cite{fan2020camouflaged} datasets, numerous deep learning-based methods have been proposed and achieved great progress. Some methods are inspired by human visual mechanisms and adopt convolutional neural networks to imitate predation behavior, thus designing a series of models for COD, such as search identification network~\cite{fan2021concealed}, positioning and focus network~\cite{mei2021camouflaged}, zoom in and out~\cite{pang2020multi}, and PreyNet~\cite{zhang2022preynet}. Some methods adopt auxiliary cues to improve network discrimination, or branch tasks to jointly learn camouflage features. The former typically employ frequency domain~\cite{zhong2022detecting}, edge/texture~\cite{ji2022gradient, zhu2021inferring}, or motion information~\cite{cheng2022implicit} to improve feature representation, and the latter usually introduces boundary detection~\cite{sun2022boundary}, classification~\cite{le2019anabranch}, fixation~\cite{lv2021simultaneously}, or saliency detection~\cite{li2021uncertainty} for multi-task collaborative learning. More recently, to improve global contextual exploration, transformer-based approaches have also been proposed, such as HitNet~\cite{hu2022high} and FSPNet~\cite{huang2023feature}. Although these methods have greatly improved the performance of camouflaged object detection, the existing methods still struggle to achieve accurate location and segmentation in most complex scenarios, due to the interference of highly similar backgrounds and the complexity of the appearance of camouflaged objects.


In recent years, diffusion models~\cite{ho2020denoising} have demonstrated impressive performance in the generative modeling of images and videos~\cite{dhariwal2021diffusion}, opening up a new era of generative models. Diffusion models are a class of generative models that consist of Markov chains trained using variational inference, to denoise noisy images blurred by Gaussian noise via learning the reverse diffusion process. Because of its powerful noise-to-image denoising pipeline, the computer vision community is curious about its variants for discriminative tasks~\cite{croitoru2023diffusion}. More recently, diffusion models have been found to be highly effective in other computer vision tasks, such as image editing~\cite{hertz2022prompt}, super-resolution~\cite{li2022srdiff}, instance segmentation~\cite{gu2022diffusioninst}, semantic segmentation~\cite{baranchuk2021label, brempong2022denoising} and medical image segmentation~\cite{aimon2023ambiguous, wu2022medsegdiff}. However, despite their great potential, diffusion models for challenging camouflaged object detection have still not been well explored. 


\begin{figure}[t]
    \centering
    \subfigure[Mainstream COD paradigm.]{
		\begin{minipage}[t]{0.99\columnwidth}
		\centering
            \includegraphics[width=0.75\linewidth,height=0.2\columnwidth]{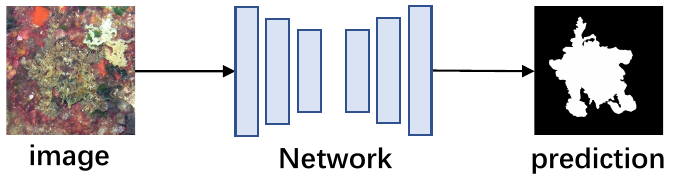}\\
            \label{fig:DGNet_FDCOD_BGNet:FDCOD}
		\end{minipage}
	}
    \subfigure[Diffusion-based COD paradigm.]{
		\begin{minipage}[t]{0.99\columnwidth}
		\centering
		\includegraphics[width=\linewidth,height=0.2\columnwidth]{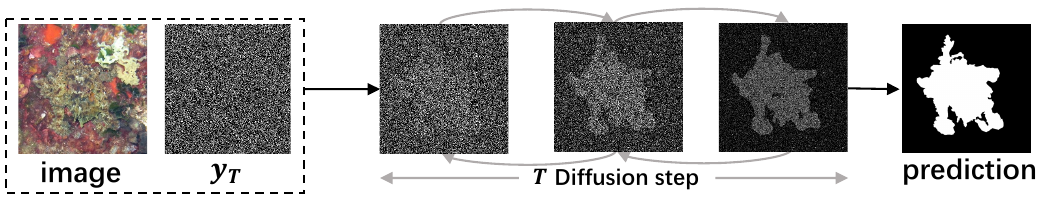}\\
            \label{fig:DGNet_FDCOD_BGNet:DGNet}
		\end{minipage}
	}
  \vspace{-12pt}
  \caption{(a) The current mainstream COD paradigm inputs images into the network for prediction in a single direction, generating a deterministic segmentation mask. (b) Our proposed diffCOD provides a novel paradigm that decomposes COD into a series of forward-and-reverse diffusion processes.}
  \captionstyle{normal}
  \label{fig:introdcution}
  \vspace{-8pt}
\end{figure}

In this paper, we propose to formulate the camouflaged object detection as a generative task, through a denoising diffusion process from the noisy mask to the object mask in the image. Specifically, in the training stage, Gaussian noise is added to the ground-truth masks to obtain noisy masks, and then the model learns to reverse this noising process. In the inference stage, the model progressively refines a set of randomly generated noisy masks from the image through the learned denoising model, until they perfectly cover the targeted object without noise.
We can see that the denoising diffusion model is the process of recovering the ground-truth mask from the random noisy distribution to the learned distribution over object masks. As shown in Figure~\ref{fig:introdcution}, unlike previous deterministic network solutions that produce a single output for an input image, we decouple the detection of the object into a novel noise-to-mask paradigm with a series of forward-and-reverse diffusion steps, which can output masks from single or multi-step denoising, thereby generating multiple object segmentation masks from a single input image.

To this end, we propose a denoising diffusion-based model, termed diffCOD, which approaches camouflaged object tasks from the perspective of the noise-to-mask denoising diffusion process. 
The proposed model adopts a denoising network conditioned on the input image prior. The semantic features extracted from the image by a Transformer encoder are integrated into the denoising diffusion model to guide the diffusion process at each step. 
To effectively bridge the gap between the diffusion noise embedding and the conditional semantic features, an injection attention module (IAM) is designed to enhance the denoising diffusion learning by aggregating conditional semantic features and diffusion model encoder through a cross-attention mechanism. 
Our contributions are summarized as follows: 
\begin{itemize}
    \item We extend the denoising diffusion models to the task of camouflaged object detection, and propose a diffusion-based object segmentation model, called diffCOD, a novel framework that views camouflaged object detection as a denoising diffusion process from noisy masks to object masks. 
  
    
    \item We design an injection attention module (IAM) to model the interaction between noise embeddings and image features. The proposed module adopts the cross-attention mechanism to integrate the conditional semantic feature extracted from the image into the diffusion model encoder to guide and enhance denoising learning. 
    
    
    \item Extensive quantitative and qualitative experiments demonstrate that the proposed diffCOD achieves superior performance over the recent 11 state-of-the-art (SOTA) methods by a large margin, especially in object detail texture segmentation, indicating the effectiveness of the proposed method. 
    
\end{itemize}


\section{Related Work}
\subsection{Camouflaged Object Detection}
Existing COD methods~\cite{Dong2023a, fan2021concealed, fan2020camouflaged} are based on a non-generative approach to segment the objects from the background. The approaches in COD can be broadly categorized into the following strategies:
a) Introducing additional cues to facilitate the exploration of camouflage features. BGNet~\cite{sun2022boundary} uses edge semantic information to enable the model to extract features that highlight the structure of the object and thus pinpoint the object boundary. TINet~\cite{zhu2021inferring} designs a texture label to find boundaries and texture differences through progressive interactive guidance. FDCOD~\cite{zhong2022detecting} incorporates frequency domain features into CNN models to better detect objects from the background. DGNet~\cite{ji2022gradient} utilizes gradient edge information to facilitate the generation of contextual and texture features. b) Multi-task learning strategies are used to improve segmentation capabilities. ANet~\cite{le2019anabranch} proposed joint learning of classification and segmentation tasks to help the model improve recognition accuracy. UJSC~\cite{li2021uncertainty} detects both salient and camouflaged objects to improve the model performance. Rank-Net~\cite{lv2021simultaneously} proposes to use the localization model to find the obvious discriminative region of the camouflaged object, and the segmentation model to segment the full range of the camouflaged object. c) Coarse-to-fine feature learning strategy is utilized to explore and integrate multi-scale features. SegMaR~\cite{jia2022segment} uses multi-stage detection to focus on the region where the goal is located. ZoomNet~\cite{pang2022zoom} learns multi-scale semantic information through multi-scale integration and hierarchical hybrid strategies to promote models that produce predictions with higher confidence. PreyNet~\cite{zhang2022preynet} imitates the predation process for stepwise aggregation and calibration of features. PFNet~\cite{mei2021camouflaged} mimics nature's predation process by first locating potential targets from a global perspective and then gradually refining the fuzzy regions. SINet~\cite{fan2020camouflaged} is designed to improve segmentation performance by locating the object first and then differentiating the details. $C^{2}$FNet~\cite{sun2021context} proposes to use global contextual information to fuse on high-level features in a cascading manner to obtain better performance. HitNet~\cite{hu2022high} and FSPNet~\cite{huang2023feature} propose to explore global context cues by transformers.
In this paper, we introduce generative models, \textit{i.e.}, denoising diffusion models, into the COD task to gradually refine the object masks from the noisy image, which achieve excellent performance, especially for objects with fine textures.  


\subsection{Diffusion Model}
The diffusion model~\cite{ho2020denoising,song2019generative} is a generative model that uses a forward Gaussian diffusion process to sample a noisy image, and then iteratively refines it using a backward generative process to obtain a denoised image. Diffusion models have shown strong potential in several fields, such as image synthesis~\cite{dhariwal2021diffusion,ho2020denoising}, image editing~\cite{hertz2022prompt}, and image super-resolution~\cite{daniels2021score}. Moreover, the learning process of diffusion models is able to capture high-level semantic information that is valuable for segmentation tasks~\cite{baranchuk2021label}, which has led to a growing interest in diffusion models for image segmentation including medical image segmentation~\cite{wu2022medsegdiff,wu2023medsegdiff}, semantic segmentation~\cite{brempong2022denoising,ji2023ddp,wu2023diffumask,xu2023open}, and instance segmentation~\cite{amit2021segdiff,gu2022diffusioninst}. MedSegDiff~\cite{wu2022medsegdiff} proposes the first DPM-based medical segmentation model, and MedSegDiff-V2~\cite{wu2023medsegdiff} further improves the performance based on it using transformer. DDeP~\cite{brempong2022denoising} finds that pre-training a semantic segmentation model as a denoising self-encoder is beneficial for performance improvement. DDP~\cite{ji2023ddp} designs a dense prediction framework with stepwise denoising refinement guided by image features. ODISE~\cite{xu2023open} combines a trained text image diffusion model with a discriminative model to achieve open-vocabulary panoptic segmentation. DiffuMask~\cite{wu2023diffumask} uses a model for the automatic generation of image and pixel-level semantic annotations, and it also shows superiority in open vocabulary segmentation. DiffusionInst~\cite{gu2022diffusioninst} proposes the first instance segmentation model based on a diffusion process to achieve global instance mask reconstruction. Segdiff~\cite{amit2021segdiff} uses a diffusion probabilistic approach to design an end-to-end segmentation model that does not rely on a pre-trained backbone. 
However, there are no studies that demonstrate the effectiveness of diffusion models in COD tasks. In this work, we present the first diffusion model for the COD segmentation task. 

\begin{figure*}[t]
\includegraphics[width=\textwidth]{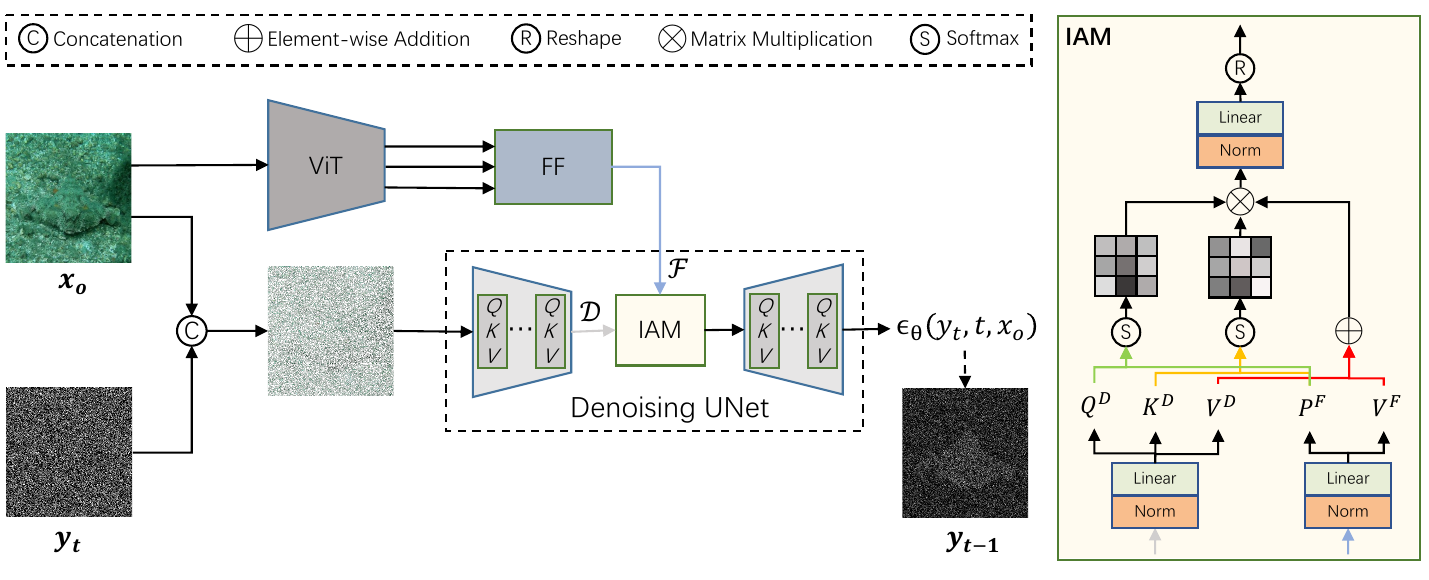}
\caption{Our proposed diffCOD framework for COD, which feeds a given image into a denoising diffusion model with UNet architecture as the core component for denoising. An injection attention module (IAM) is designed to implicitly guide the diffusion process with the conditional semantic features that have gone through the backbone and feature fusion module (FF), allowing the model to take full advantage of the correspondence between image features and diffusion information.}
\label{fig:overall}
\end{figure*}

\section{Methodology}
In this section, we first review the diffusion model (Sec.~\ref{diffusion_model}). Then we introduce the architecture of diffCOD  (Sec.~\ref{diffCOD}). Finally, we describe the specific process of training and inference of diffCOD (Sec.~\ref{training} \& Sec.~\ref{Inference}).

\subsection{Diffusion Model}
\label{diffusion_model}
The diffusion probability model has reaped plenty of attention due to its simple training process and excellent performance. It is mainly divided into forward process and reverse process. In the forward process, noise is added to the target image to make it closer to the Gaussian distribution. The reverse process learns to map the noise to the real image.

The forward process refers to the gradual incorporation of Gaussian noise with variance $\beta_{t} \in(0,1)$ into the original image $x_{0} \sim p\left(x_{0}\right)$ at time $t$ until it converges to isotropic Gaussian distribution. The forward process is described by the formulation:
\begin{equation}
    q\left(x_{t} \mid x_{t-1}\right)=\mathcal{N}\left(x_{t} ; \sqrt{1-\beta_{t}} x_{t-1}, \beta_{t}\mathbf{I}\right)
\end{equation}
where $t \in [1,T]$. We can obtain the latent variable $x_{t}$ directly by using $x_{0}$ by the following equation:
\begin{equation}
    q\left(x_{t} \mid x_{0}\right)=\mathcal{N}\left(x_{t} ; \sqrt{\bar{\alpha}_{t}} x_{0},\left(1-\bar{\alpha}_{t}\right) \mathbf{I}\right)
\end{equation}
where $\alpha_{t}:=1-\beta_{t}$, $\bar{\alpha}_{t}:=\prod_{s=0}^{t} \alpha_{s}$ and $\epsilon \sim \mathcal{N}(0,\mathbf{I})$.

The reverse process converts the latent variable distribution $p(x_{T})$ to $p(x_{0})$ through a Markov chain, and the reverse process can be denoted as follows:
\begin{equation}
    p_{\theta}\left(x_{t-1} \mid x_{t}\right)=\mathcal{N}\left(x_{t-1} ; \mu_{\theta}\left(x_{t}, t\right), \Sigma_{\theta}\left(x_{t}, t\right)\right)
\end{equation}

The combination of $q$ and $p$ is a variational auto-encoder, and the variational lower bound (VLB) is defined as follows:
\begin{equation}
    L_{\mathrm{vlb}}:=L_{0}+L_{1}+\ldots+L_{T-1}+L_{T}
    \label{Loss1}
\end{equation}
\begin{equation}
    L_{0}:=-\log p_{\theta}\left(x_{0} \mid x_{1}\right)
    \label{Loss2}
\end{equation}
\begin{equation}
    L_{t-1}:=D_{K L}\left(q\left(x_{t-1} \mid x_t, x_0\right) \|\ p_\theta\left(x_{t-1} \mid x_t\right)\right)
    \label{Loss3}
\end{equation}
\begin{equation}
    L_T:=D_{K L}\left(q\left(x_T \mid x_0\right) \|\ p\left(x_T\right)\right)
    \label{Loss4}
\end{equation}

\subsection{Architecture}
\label{diffCOD}
As shown in Figure~\ref{fig:overall}, the proposed diffCOD aims to solve the COD task by the diffusion model. The denoising network of diffCOD is based on the UNet architecture~\cite{ronneberger2015u}. To get effective conditional semantic features, we obtain multi-scale features by ViT-based backbone and feature fusion (FF) to yield features containing rich multi-scale details. In addition, to let the texture patterns and localization information in the conditional semantic features guide the denoising process, we propose an injection attention module (IAM) based on cross-attention. This allows the network to reduce the difference between diffusion features and image features and to combine the advantages of both.

\vspace{10pt}
\noindent\textbf{Feature Fusion (FF).}
Given an initial input image $x_{o} \in \mathbb{R}^{H\times W \times 3}$, we adopt the top-three high-level features of the visual backbone as our multi-scale backbone features, denoted as $\mathcal{X}^{p}_{i}$, $i \in \{1, 2, 3\}$ whose resolution is $\frac{H}{k} \times \frac{W}{k}$, $k \in \{8, 16, 32\}$. Here we use PVTv2~\cite{wang2021pvtv2} as the backbone. Then FF is used to aggregate these multiscale features. Specifically, FF contains three branches to process $\mathcal{X}^{p}_{i}$, each branch uses two convolution operations with 3$\times$3 kernel for feature enhancement, and finally the three branches are coalesced by a single convolution to obtain $\mathcal{F} \in \mathbb{R}^{\frac{H}{32} \times \frac{W}{32} \times C}$.

\vspace{10pt}
\noindent\textbf{Injection Attention Module (IAM).} To introduce texture and location information of the original features in the noise prediction process, we employ a cross-attention-based IAM, which is embedded in the middle of the UNet-based denoising network. Given the multiscale fusion feature $\mathcal{F}$ from FF and the deepest feature $\mathcal{D} \in \mathbb{R}^{\frac{H}{32} \times \frac{W}{32} \times C}$ from the diffusion model as the common input to the IAM. Specifically, $\mathcal{D}$ is transformed by linear projection to generate the query $\mathbf{Q^{D}}$, the key $\mathbf{K^{D}}$ and the value $\mathbf{V^{D}}$. $\mathcal{F}$ generates $\mathbf{P^{F}}$, $\mathbf{V^{F}}$ by linear projection, and it is noteworthy that $\mathcal{F}$ does not generate the query and the key for similarity comparison, but uses the generated $\mathbf{P^{F}}$ to act as an intermediary for similarity comparison with $\mathcal{D}$. This process is defined as follows:
\begin{equation}
    \begin{split}
        \mathbf{Q^{D}}=\mathcal{D} \cdot \mathcal{W_{Q}^{D}}&, \quad \mathbf{K^{D}}= \mathcal{D} \cdot \mathcal{W_{K}^{D}}, \quad  \mathbf{V^{D}}=\mathcal{D} \cdot \mathcal{W_{V}^{D}} \\
        \mathbf{P^{F}}&=\mathcal{F} \cdot \mathcal{W_{P}^{F}}, \quad \mathbf{V^{F}}= \mathcal{F} \cdot\mathcal{W_{V}^{F}}
    \end{split}
\end{equation}
where $\mathcal{W_{Q}^{D}}$, $\mathcal{W_{K}^{D}}$, $\mathcal{W_{V}^{D}}$, $\mathcal{W_{P}^{F}}$, $\mathcal{W_{V}^{F}}$ $\in \mathbb{R}^{d \times d}$. $d$ is the dimensionality.

Thus the IAM operation is defined as follows:
\begin{equation}
    \mathbf{M}^{att}_{1}=\operatorname{Softmax}\left(\frac{ \mathbf{Q^{D}} \cdot ( \mathbf{P^{F}})^{T} }{\sqrt{d}}\right)
\end{equation}
\begin{equation}
    \mathbf{M}^{att}_{2}=\operatorname{Softmax}\left(\frac{ \mathbf{K^{D}} \cdot (\mathbf{P^{F}})^{T} }{\sqrt{d}}\right)
\end{equation}
\begin{equation}
    O^{I} = \mathbf{M}^{att}_{1} \cdot \mathbf{M}^{att}_{2} \cdot (\mathbf{V}^{D} + \mathbf{V}^{F})
\end{equation}
where $\mathbf{M}^{att}_{1}$ and $\mathbf{M}^{att}_{2}$ represent the attention maps of $\mathbf{Q^{D}}$-$\mathbf{P^{F}}$ and $\mathbf{K^{D}}$-$\mathbf{P^{F}}$, respectively. $O^{I} \in \mathbb{R}^{\frac{H}{32} \times \frac{W}{32} \times C}$ denotes the final generated cross-attention fusion feature.

\subsection{Training}
\label{training}
In the forward process, the Gaussian noise $\epsilon_{t}$ is added to the ground truth $y_{0}$ to obtain the noise mapping ${y}_t \sim q\left({y}_t \mid {y}_0\right)$ by $T$-steps. The intensity of the noise is controlled by $\alpha_{t}$ and conforms to the standard normal distribution. This process can be defined as follows:
\begin{equation}
    y_{t}=\sqrt{\alpha_{t}} y_{t-1}+\left(1-\alpha_{t}\right) \epsilon_{t}
\end{equation}
where $t=[1, \cdots, T]$ and $\epsilon_{t} \sim \mathcal{N}(0, \mathbf{I})$. 

By iterative computation, we can directly obtain $y_{t}$. This process can be further marginalized as:
\begin{equation}
    y_{t}=\sqrt{\bar{\alpha}_{t}} y_{0}+\left(1-\bar{\alpha}_{t}\right) \epsilon_{t}
\end{equation}
where $\bar{\alpha}_t=\prod_{i=1}^t \alpha_i$. 

In the reverse process, we map from $y_{t}$ to $y_{t-1}$ until the segmented image is acquired step by step. The mathematics is defined as follows:
\begin{equation}
    \label{Equation_reverse}
    y_{t-1}=\mu_\theta\left(y_t, t, x_{o}\right)+\Sigma_\theta\left(y_t, t, x_{o}\right) \epsilon_t
\end{equation}

We train a denoising UNet model to predict ${\epsilon}_\theta\left({y}_t, t, {x}_o\right)$:
\begin{equation}
    \mu_\theta\left(y_t, t,x_{o}\right)=\frac{\left(y_t-\left(\frac{1-\alpha_t}{\sqrt{1-\bar{\alpha}_t}}\right) \epsilon_\theta\left(y_t, t,x_{o}\right)\right)}{\sqrt{\alpha_t}}  
\end{equation}

We follow the improved DDPM~\cite{nichol2021improved} to simplify Eq.~(4)-(7) to define the hybrid objective $L_{\mathrm{hybrid}}=L_{\mathrm{simple}}+ L_{\mathrm{vlb}}$. $L_{\mathrm{vlb}}$ learns the term $\Sigma_\theta\left(y_t, t,x_o\right)$. Furthermore, inspired by~\cite{wu2023medsegdiff}, we use FF and a convolution layer to provide an initial static mask $y_{m}$ to reduce the diffusion variance, and its mean square loss is defined as $L_{\mathrm{static}}$. Total loss function $L_{total}$ is defined as follows:
\begin{equation}
\left\{
\begin{array}{ll}
L_{\mathrm{simple}} &=\mathbb{E}_{t \sim[1, T], y_0 \sim q\left(y_0\right), \epsilon}\left\|\epsilon-\epsilon_\theta\left(y_t, t, x_o \right)\right\|^2 \\
L_{\mathrm{static}} &=\mathbb{E}_{y_0 \sim q\left(y_0\right), y_m}\left\|y_{0} - y_m\right\|^2 \\
L_{\mathrm{total}} &=L_{\mathrm{simple}}+ L_{\mathrm{vlb}}+L_{\mathrm{static}}
\end{array}
\right.
\end{equation}
Algorithm~\ref{algo:training} provides the training procedure for diffCOD.

\begin{algorithm}[ht]
\caption{diffCOD Training}
\label{algo:training}
\definecolor{codeblue}{HTML}{2E8B57} 
\definecolor{codekw}{HTML}{DC143C} 
\lstset{
  backgroundcolor=\color{white},
  columns=fullflexible,
  breaklines=true,
  captionpos=b,
  commentstyle=\fontsize{7.2pt}{7.2pt}\color{codeblue},
  keywordstyle=\fontsize{7.2pt}{7.2pt}\color{codekw},
  escapechar={|}, 
}
\lstset{language=Python}
\begin{lstlisting}[xleftmargin=-1em]
def training_loss(images, masks):
  """images: [b, h, w, 3], masks: [b, h, w, 1]"""

  # Encode images
  X_p = ViT(images) 
  F = FF(X_p) 
    
  # corrupt groundtruth
  t = uniform(0, 1)
  eps = normal(mean=0, std=1)
  mask_crpt = sqrt(gamma(t)) * masks +
              sqrt(1 - gamma(t)) * eps

  # predict and backward
  D = UNet_1(images, mask_crpt, t)
  O = IAM(F, D)
  preds = UNet_2(O)

  # compute loss
  loss = loss_function(preds, masks)
  return loss
\end{lstlisting}
\end{algorithm}

\subsection{Inference}
\label{Inference}
In the inference stage, we step-by-step apply Eq.~(\ref{Equation_reverse}) to sample a pure Gaussian noise $y_{t} \sim \mathcal{N}(0, I)$. In addition, we add conditional information related to the image features to guide the inference process. After performing $T$ iterations, we can obtain the segmentation image of the camouflaged object. Using the setting of \cite{nichol2021improved} for the sampling, the inference process of diffCOD is shown in Algorithm~\ref{algo:inference}.

\begin{algorithm}[ht]
\caption{diffCOD Inference}
\label{algo:inference}
\definecolor{codeblue}{HTML}{2E8B57} 
\definecolor{codekw}{HTML}{DC143C} 
\lstset{
  backgroundcolor=\color{white},
  columns=fullflexible,
  breaklines=true,
  captionpos=b,
  commentstyle=\fontsize{7.2pt}{7.2pt}\color{codeblue},
  keywordstyle=\fontsize{7.2pt}{7.2pt}\color{codekw},
  escapechar={|}, 
}
\lstset{language=Python}
\begin{lstlisting}[xleftmargin=-1em]
def inference(images, steps):
  """images: [b, h, w, 3], steps: sample steps"""

  # Encode images
  X_p = ViT(images) 
  F = FF(X_p) 
    
  m_t = normal(mean=0, std=1)
  
  # time intervals
  for step in range(steps):
    out = p_sample(images, F, m_t, step)

  return out
\end{lstlisting}
\end{algorithm}

\begin{table*}[t]
\resizebox{\textwidth}{!}{
\renewcommand{\arraystretch}{1.3}
\begin{tabular}{c|cccccccccccccccccccc}
\toprule[1pt]
\multirow{2}{*}{\textbf{Method}} & \multicolumn{5}{c|}{\textbf{COD10K}}                                                                       & \multicolumn{5}{c|}{\textbf{NC4K}}                                                                  & \multicolumn{5}{c|}{\textbf{CAMO}}                                                                     & \multicolumn{5}{c}{\textbf{CHAMELEON}}                                                 \\ \cline{2-21} 
                                 & $S_{\alpha}\uparrow$              & $F_{\beta}^{\omega}\uparrow$             & $F_{m}\uparrow$             & $E_{m}\uparrow$             & \multicolumn{1}{c|}{$MAE\downarrow$}                                 & $S_{\alpha}$              & $F_{\beta}^{\omega}\uparrow$             & $F_{m}\uparrow$             & $E_{m}\uparrow$             & \multicolumn{1}{c|}{$MAE\downarrow$}                                 & $S_{\alpha}$              & $F_{\beta}^{\omega}\uparrow$             & $F_{m}\uparrow$             & $E_{m}\uparrow$             & \multicolumn{1}{c|}{$MAE\downarrow$}                                 & $S_{\alpha}\uparrow$              & $F_{\beta}^{\omega}\uparrow$             & $F_{m}\uparrow$             & $E_{m}\uparrow$            & $MAE\downarrow$            \\ \hline
\multicolumn{1}{l|}{2019 CPD \cite{wu2019cascaded}}          & 0.736          & 0.547          & 0.607          & 0.801          & \multicolumn{1}{c|}{0.053}          & 0.769          & 0.652          & 0.713          & 0.822          & \multicolumn{1}{c|}{0.072}          & 0.688          & 0.552          & 0.623          & 0.728          & \multicolumn{1}{c|}{0.114}          & 0.876          & 0.809          & 0.821          & 0.914          & 0.036          \\
\multicolumn{1}{l|}{2019 EGNet \cite{Zhao_2019_ICCV}}        & 0.746          & 0.560           & 0.591          & 0.789          & \multicolumn{1}{c|}{0.053}          & 0.804          & 0.727          & 0.731          & 0.834          & \multicolumn{1}{c|}{0.066}          & 0.730          & 0.579          & 0.693          & 0.762          & \multicolumn{1}{c|}{0.104}          & 0.851          & 0.705          & 0.747          & 0.869          & 0.049          \\
\multicolumn{1}{l|}{2020 SINet \cite{fan2020camouflaged}}    & 0.772          & 0.543          & 0.640          & 0.810          & \multicolumn{1}{c|}{0.051}          & 0.810          & 0.665          & 0.741          & 0.841          & \multicolumn{1}{c|}{0.066}          & 0.753          & 0.602          & 0.676          & 0.774          & \multicolumn{1}{c|}{0.097}          & 0.867          & 0.727          & 0.792          & 0.889          & 0.044          \\
\multicolumn{1}{l|}{2020 MINet \cite{pang2020multi}}        & 0.780          & 0.628          & 0.677          & 0.838          & \multicolumn{1}{c|}{0.040}          & 0.810          & 0.717          & 0.764          & 0.856          & \multicolumn{1}{c|}{0.057}          & 0.741          & 0.629          & 0.682          & 0.783          & \multicolumn{1}{c|}{0.096}          & 0.853          & 0.768          & 0.803          & 0.902          & 0.035          \\
\multicolumn{1}{l|}{2020 PraNet \cite{fan2020pranet}}       & 0.800          & 0.656          & 0.699          & 0.869          & \multicolumn{1}{c|}{0.041}          & 0.826          & 0.739          & 0.780          & 0.878          & \multicolumn{1}{c|}{0.056}          & 0.769          & 0.664          & 0.716          & 0.812          & \multicolumn{1}{c|}{0.091}          & 0.870          & 0.790           & 0.816          & 0.915          & 0.039          \\
\multicolumn{1}{l|}{2021 PFNet \cite{mei2021camouflaged}}        & 0.797          & 0.656          & 0.698          & 0.875          & \multicolumn{1}{c|}{0.039}          & 0.826          & 0.743          & 0.783          & 0.884          & \multicolumn{1}{c|}{0.054}          & 0.774          & 0.683          & 0.737          & 0.832          & \multicolumn{1}{c|}{0.087}          & 0.889          & 0.823          & \textbf{0.840}          & \textbf{0.946}          & \textbf{0.030}          \\
\multicolumn{1}{l|}{2021 LSR \cite{lv2021simultaneously}}          & 0.805          & 0.660           & 0.703          & 0.876          & \multicolumn{1}{c|}{0.039}          & 0.832          & 0.743          & 0.785          & 0.888          & \multicolumn{1}{c|}{0.053}          & 0.793          & 0.703          & 0.753          & 0.850           & \multicolumn{1}{c|}{0.083}          & 0.890          & 0.824          & 0.834          & 0.932          & 0.034          \\
\multicolumn{1}{l|}{2022 ERRNet \cite{ji2022fast}}      & 0.780          & 0.629          & 0.679          & 0.867          & \multicolumn{1}{c|}{0.044}          & ---          & ---          & ---          & ---          & \multicolumn{1}{c|}{---}          & 0.761          & 0.660          & 0.719          & 0.817          & \multicolumn{1}{c|}{0.088}          & 0.877          & 0.805          & 0.821          & 0.927          & 0.036          \\
\multicolumn{1}{l|}{2022 NCHIT \cite{zhang2022camouflaged}}      & 0.790          & 0.608          & 0.689          & 0.817          & \multicolumn{1}{c|}{0.046}          & ---          & ---          & ---          & ---          & \multicolumn{1}{c|}{---}          & 0.780          & 0.671          & 0.733          & 0.803          & \multicolumn{1}{c|}{0.088}          & 0.874          &  0.793         & 0.812          & 0.891          & 0.041          \\
\multicolumn{1}{l|}{2022 CubeNet \cite{zhuge2022cubenet}}      & 0.795          & 0.644          & 0.681          & 0.864          & \multicolumn{1}{c|}{0.041}          & ---          & ---          & ---          & ---          & \multicolumn{1}{c|}{---}          & 0.788          & 0.682          & 0.743          & 0.838          & \multicolumn{1}{c|}{0.085}          & 0.873          & 0.787          & 0.823          & 0.928          & 0.037          \\
\multicolumn{1}{l|}{2023 CRNet \cite{he2022weakly}}      & 0.733          & 0.576          & 0.627          & 0.832          & \multicolumn{1}{c|}{0.049}          & ---          & ---          & ---          & ---          & \multicolumn{1}{c|}{---}          & 0.735          & 0.641          & 0.702          & 0.815          & \multicolumn{1}{c|}{0.092}          & 0.818          & 0.744          & 0.756          & 0.897          & 0.046          \\ \hline
\multicolumn{1}{c|}{diffCOD}        & \textbf{0.812} & \textbf{0.684} & \textbf{0.723} & \textbf{0.892} & \multicolumn{1}{c|}{\textbf{0.036}} & \textbf{0.837} & \textbf{0.761} & \textbf{0.802} & \textbf{0.891} & \multicolumn{1}{c|}{\textbf{0.051}} & \textbf{0.795} & \textbf{0.704} & \textbf{0.758} & \textbf{0.852} & \multicolumn{1}{c|}{\textbf{0.082}} & \textbf{0.893} & \textbf{0.826} & 0.837 & 0.933 & \textbf{0.030} \\

\bottomrule[1pt]
\end{tabular}
}
\caption{Quantitative comparisons of our proposed method and other 11 state-of-the-art methods on four widely used benchmark datasets. The higher the $S_{\alpha}$, $F_{\beta}^{\omega}$, $F_{m}$, and $E_{m}$, the better the performance. The smaller the $MAE$, the better. The best results are marked in $\mathbf{bold}$.}
\label{tab1}
\end{table*}

\section{Experiments}
\subsection{Experimental Setup}

\noindent\textbf{Datasets.} 
We conduct experiments on four widely used benchmark datasets of COD task, \textit{i.e.}, CAMO~, CHAMELEON, COD10K and NC4K. The details of each dataset
are as follows: 
\begin{itemize}
    \item CAMO contains 1,250 camouflaged images and 1,250 non-camouflaged images, covering eight categories.
    \item CHAMELEON has a total of 76 camouflaged images. 
    \item COD10K consists of 5,066 camouflaged, 1,934 non-camouflaged, and 3,000 background images. It is currently the largest dataset which covers 10 superclasses and 78 subclasses.
    \item NC4K is a newly published dataset that has a total of 4,121 camouflaged images.
\end{itemize}


Following the standard practice of COD tasks, we use 3,040  images from COD10K and 1,000 images from CAMO as the training set and the remaining data as the test set.
\\ \hspace*{\fill} \\
\noindent\textbf{Evaluation metrics.} 
According to the standard evaluation protocol of COD, we employ the five common metrics to evaluate our model, \textit{i.e.}, 
structure-measure ($S_{\alpha}$), weighted F-measure ($F_{\beta}^{\omega}$), mean F-measure ($F_{m}$), mean E-measure ($E_{m}$) and mean absolute error ($MAE$).
The purpose of structure-measure ($S_{\alpha}$) is to evaluate the structural information of the result and ground truth, including object perception and region perception. Weighted F-measure $F_{\beta}^{\omega}$ is the weighted information of the mean F-measure ($F_{m}$) metric, and these two metrics are a combined assessment of the accuracy and recall of the result. Mean E-measure ($E_{m}$) is able to perform both pixel-level matching and image-level statistics, and is used to calculate the overall and local accuracy of the segmentation results. The mean absolute error ($MAE$) metric is often used to evaluate the average pixel-level relative error between the result and ground truth.
\\ \hspace*{\fill} \\
\noindent\textbf{Implementation details.}
The proposed method is implemented with the PyTorch toolbox. 
We set the time step as $T$ = 1000 with a linear noise schedule for all the experiments.
We use Adam as our model optimizer with a learning rate of 1e-4. The batch size is set to 64.
During the training, the input images are resized to 256$\times$256 via bilinear interpolation and augmented by random flipping, cropping, and color jittering.
\\ \hspace*{\fill} \\
\noindent\textbf{Baselines.} 
Our diffCOD is compared with 11 recent state-of-the-art methods, including CPD \cite{wu2019cascaded}, EGNet \cite{Zhao_2019_ICCV}, SINet \cite{fan2020camouflaged}, MINet \cite{pang2020multi}, PraNet \cite{fan2020pranet}, PFNet \cite{mei2021camouflaged}, LSR \cite{lv2021simultaneously}, ERRNet \cite{ji2022fast}, NCHIT \cite{zhang2022camouflaged}, CubeNet \cite{zhuge2022cubenet}, CRNet \cite{he2022weakly}. For a fair comparison, all results are either provided by the authors or reproduced by an open-source model re-trained on the same training set with the recommended setting.

\subsection{Quantitative Evaluation}
The quantitative comparison of our proposed diffCOD with 11 state-of-the-art methods is shown in Table~\ref{tab1}. Our method achieves superior performance over other competitors, indicating that our model can generate high-quality camouflaged segmentation masks compared to previous methods. 
For the largest COD10K dataset, our method shows a substantial performance jump, with an average increase of 4.8\%, 12.8\%, 9.5\%, 6.4\% and 19.1\% for $S_{\alpha}$, $F_{\beta}^{\omega}$, $F_{m}$, $E_{m}$ and $MAE$, respectively. 
For another recent large-scale NC4K dataset, diffCOD also outperforms all methods, increasing by 3.4\%, 7.1\%, 6.1\%, 4.0\% and 14.8\% on average for $S_{\alpha}$, $F_{\beta}^{\omega}$, $F_{m}$, $E_{m}$ and $MAE$, respectively.
In addition, the most significant increases in the CAMO dataset were seen in the $F_{\beta}^{\omega}$ and $MAE$, with improvements of 10.2\% and 11.3\%, respectively. 
CHAMELEON is the smallest COD dataset, therefore most of the methods perform inconsistently on this dataset, our method increases 3.0\%, 6.2\%, 4.0\%, 2.6\% and 21.2\% for $S_{\alpha}$, $F_{\beta}^{\omega}$, $F_{m}$, $E_{m}$ and $MAE$, respectively.

\begin{figure}[t]
	\centering
	\subfigure[Image]{
		\begin{minipage}[t]{0.32\columnwidth}
			\centering
			\includegraphics[width=1.01\linewidth,height=0.84\columnwidth]{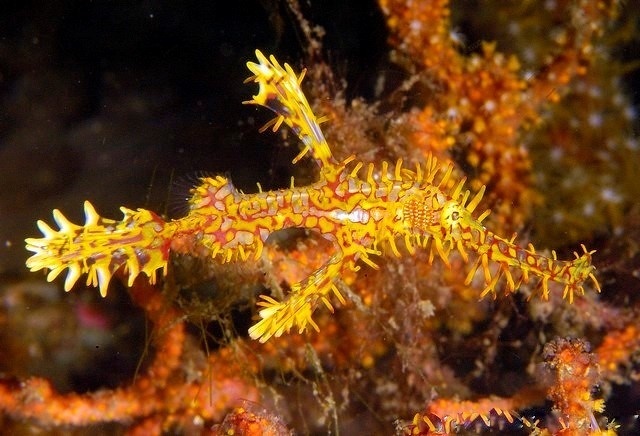}\\
			\vspace{0.01\linewidth}
                \includegraphics[width=1.01\linewidth,height=0.84\columnwidth]{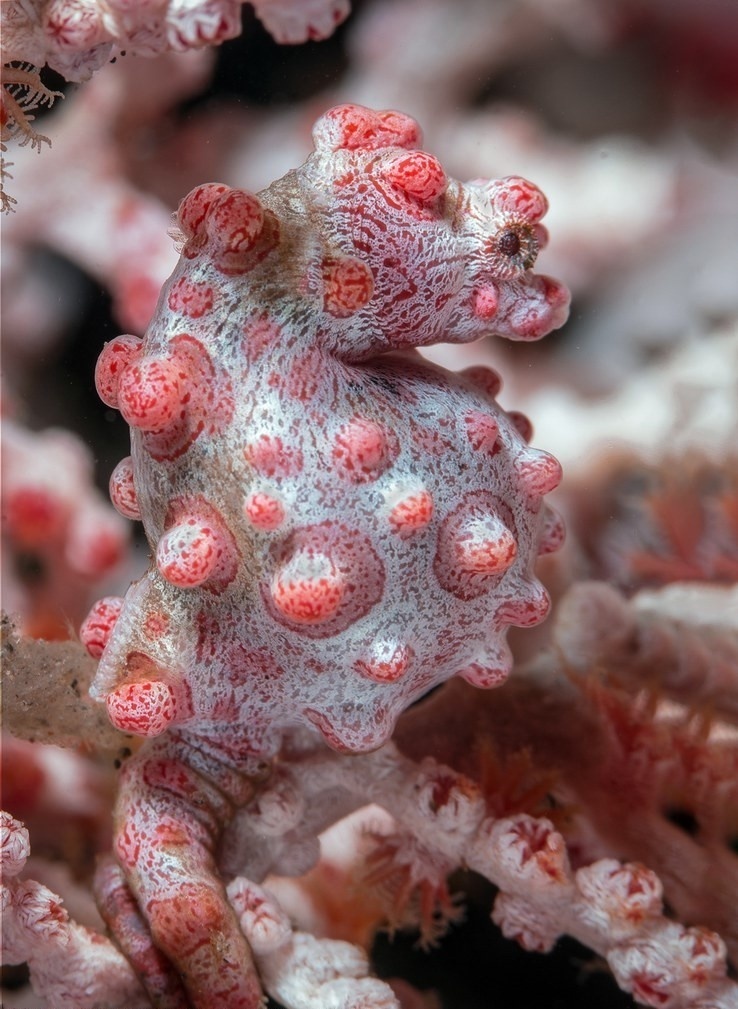}\\
			\vspace{0.01\linewidth}
                \includegraphics[width=1.01\linewidth,height=0.84\columnwidth]{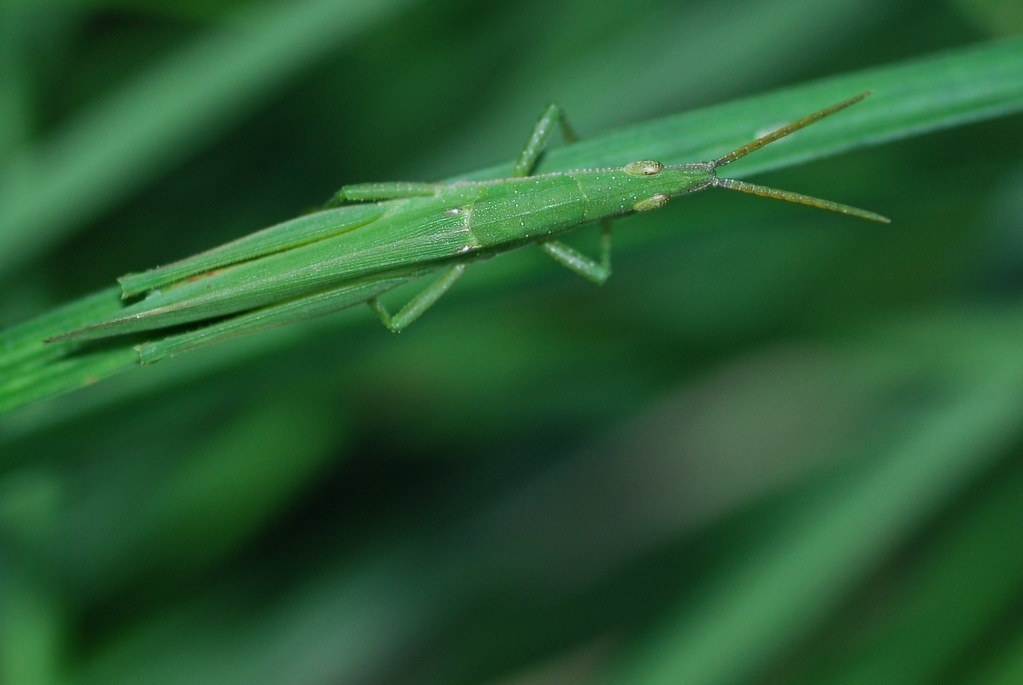}\\
                \vspace{0.01\linewidth}
			\includegraphics[width=1.01\linewidth,height=0.84\columnwidth]{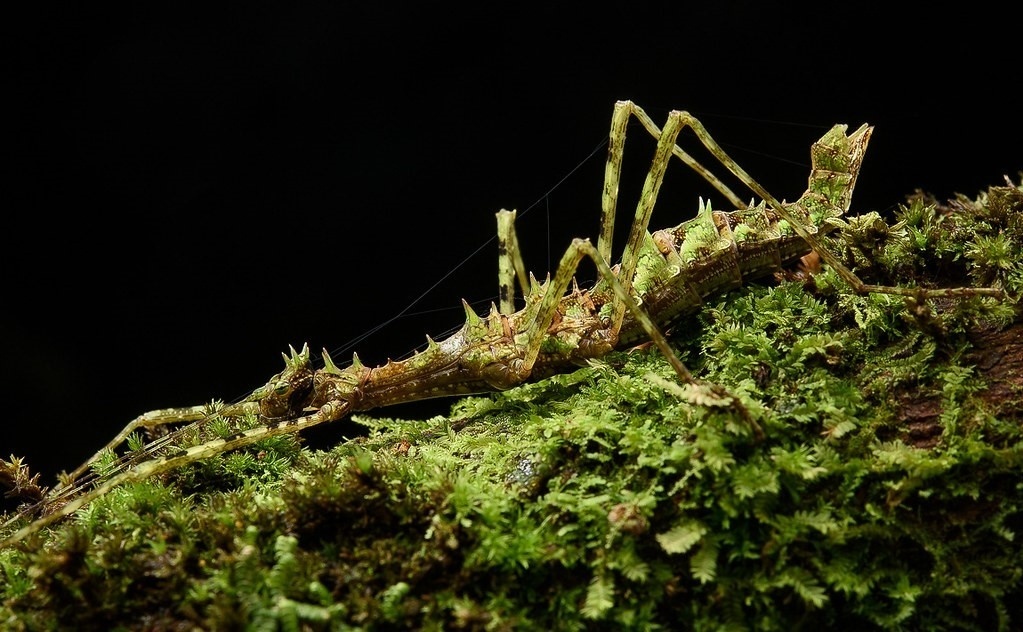}\\
			\vspace{0.11\linewidth}
		\end{minipage}%
	}\hspace{-0.012\columnwidth}
	\subfigure[GT]{
		\begin{minipage}[t]{0.32\columnwidth}
			\centering
			\includegraphics[width=1.01\linewidth,height=0.84\columnwidth]{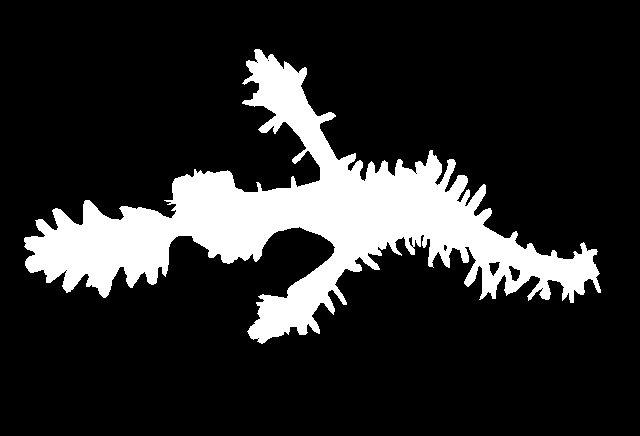}\\
			\vspace{0.01\linewidth}
                \includegraphics[width=1.01\linewidth,height=0.84\columnwidth]{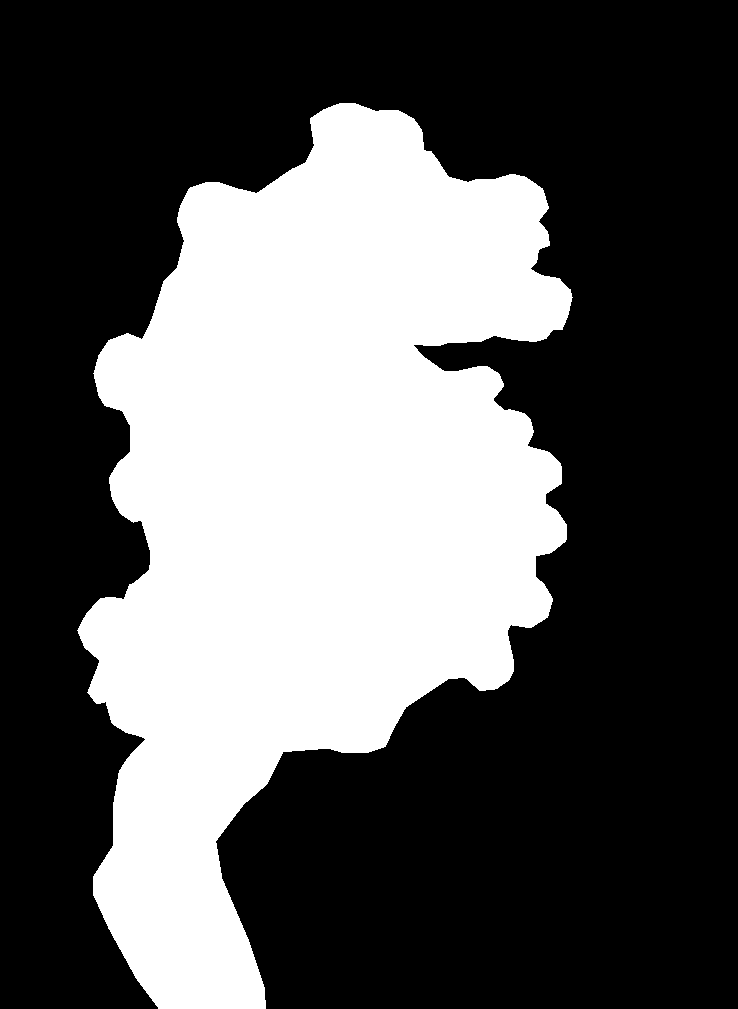}\\
			\vspace{0.01\linewidth}
                \includegraphics[width=1.01\linewidth,height=0.84\columnwidth]{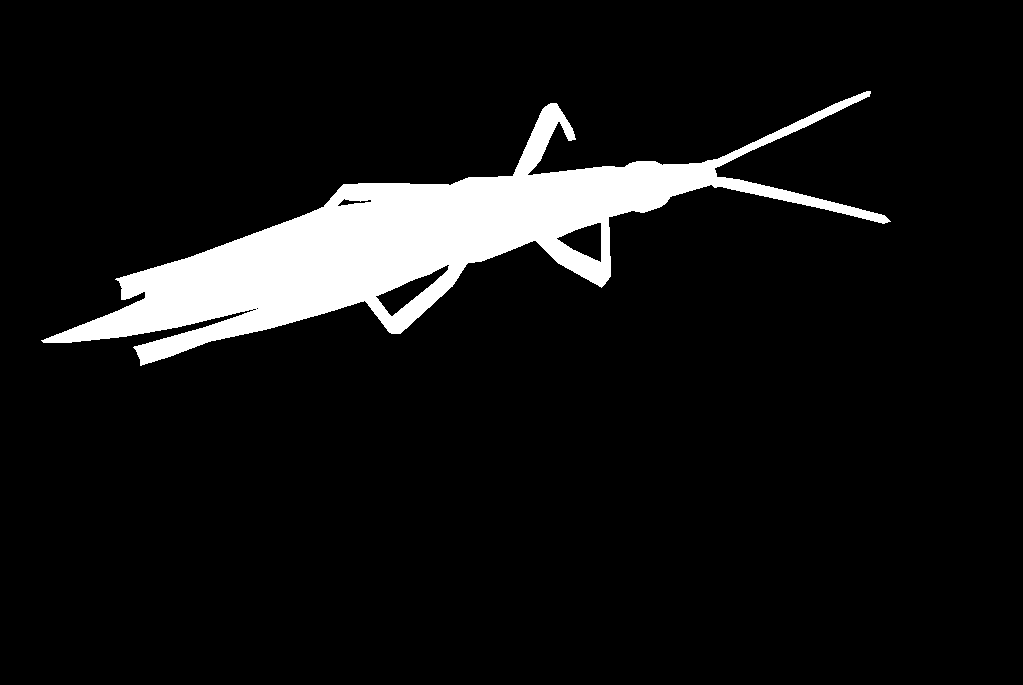}\\
                \vspace{0.01\linewidth}
			\includegraphics[width=1.01\linewidth,height=0.84\columnwidth]{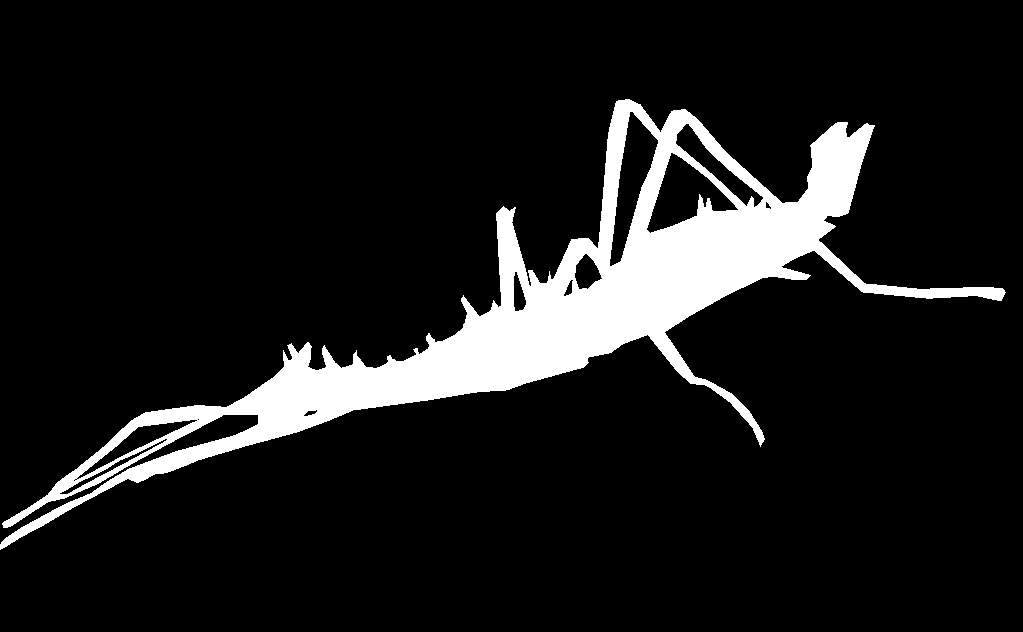}\\
			\vspace{0.11\linewidth}
		\end{minipage}%
	}\hspace{-0.012\columnwidth}
	\subfigure[diffCOD]{
		\begin{minipage}[t]{0.32\columnwidth}
			\centering
			\includegraphics[width=1.01\linewidth,height=0.84\columnwidth]{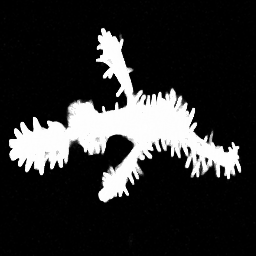}\\
			\vspace{0.01\linewidth}
                \includegraphics[width=1.01\linewidth,height=0.84\columnwidth]{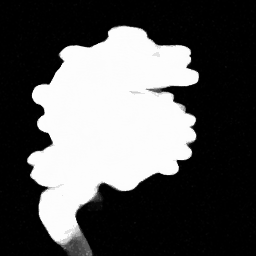}\\
			\vspace{0.01\linewidth}
                \includegraphics[width=1.01\linewidth,height=0.84\columnwidth]{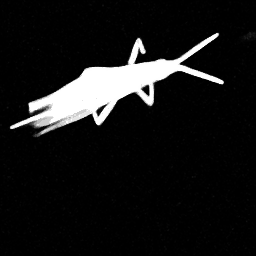}\\
                \vspace{0.01\linewidth}
			\includegraphics[width=1.01\linewidth,height=0.84\columnwidth]{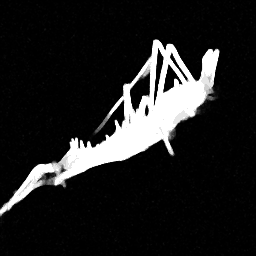}\\
			\vspace{0.11\linewidth}
		\end{minipage}%
	}\hspace{-0.012\columnwidth}
	\centering
	\caption{Visual results of our proposed model in terms of detailed textures.}
    \label{fig:detail}

\end{figure}

\begin{figure*}[t]
	\centering
	\subfigure[{\scriptsize Image}]{
		\begin{minipage}[t]{0.1\textwidth}
			\centering
			\includegraphics[width=1.15\textwidth,height=0.840\textwidth]{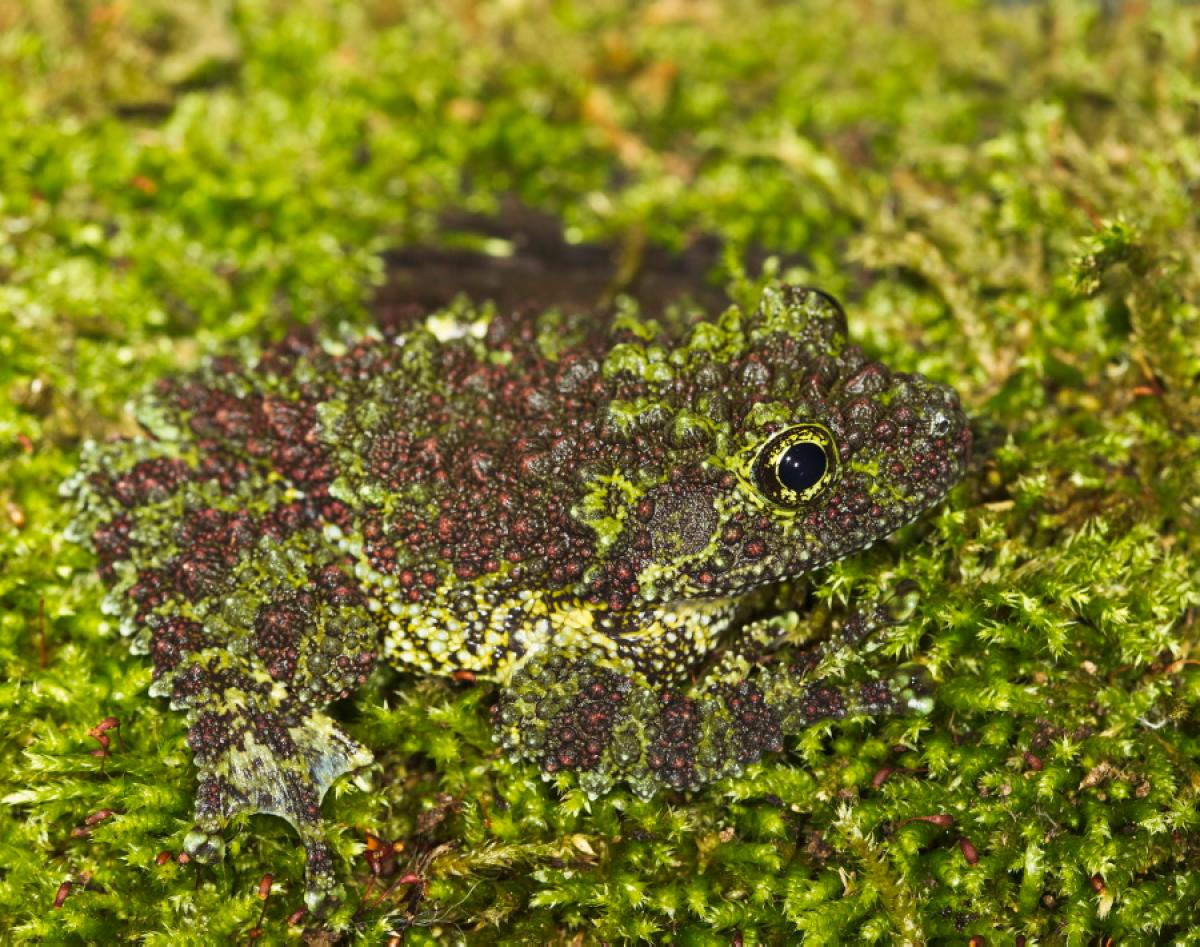}\\
			\vspace{0.01\linewidth}
                \includegraphics[width=1.15\textwidth,height=0.840\textwidth]{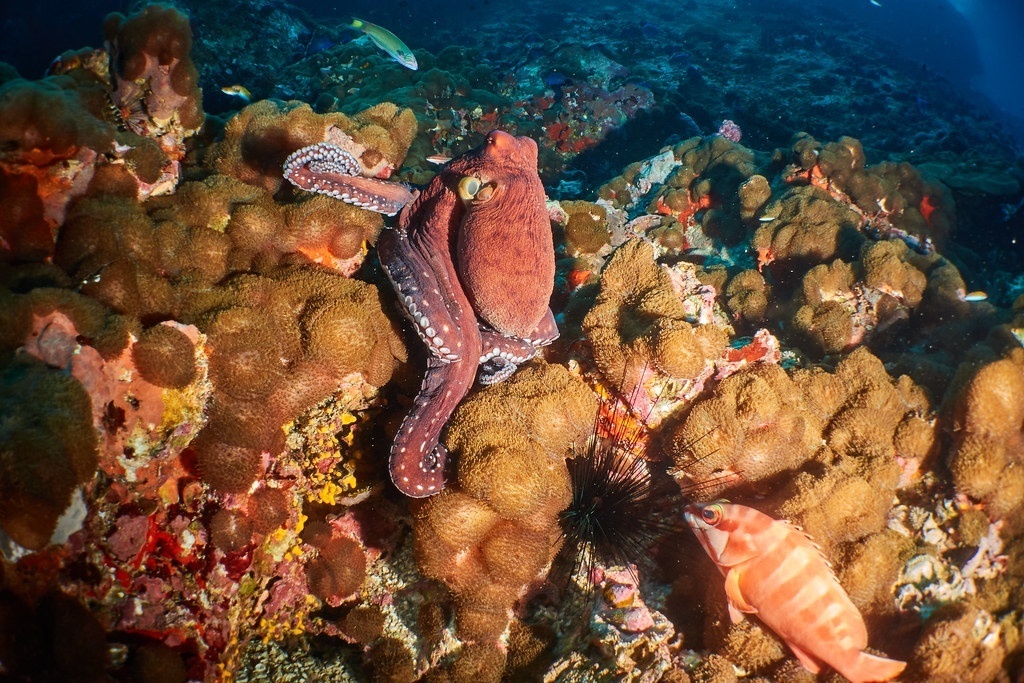}\\
			\vspace{0.01\linewidth}
                \includegraphics[width=1.15\textwidth,height=0.840\textwidth]{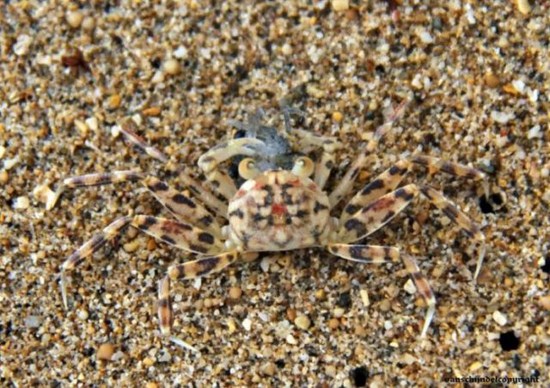}\\
                \vspace{0.01\linewidth}
			\includegraphics[width=1.15\linewidth,height=0.840\textwidth]{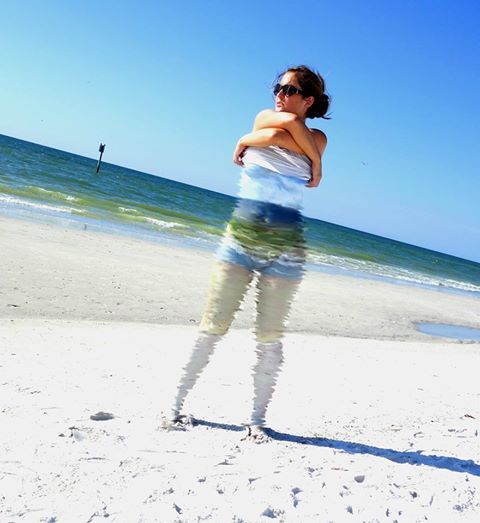}\\
			\vspace{0.01\linewidth}
                \includegraphics[width=1.15\linewidth,height=0.840\textwidth]{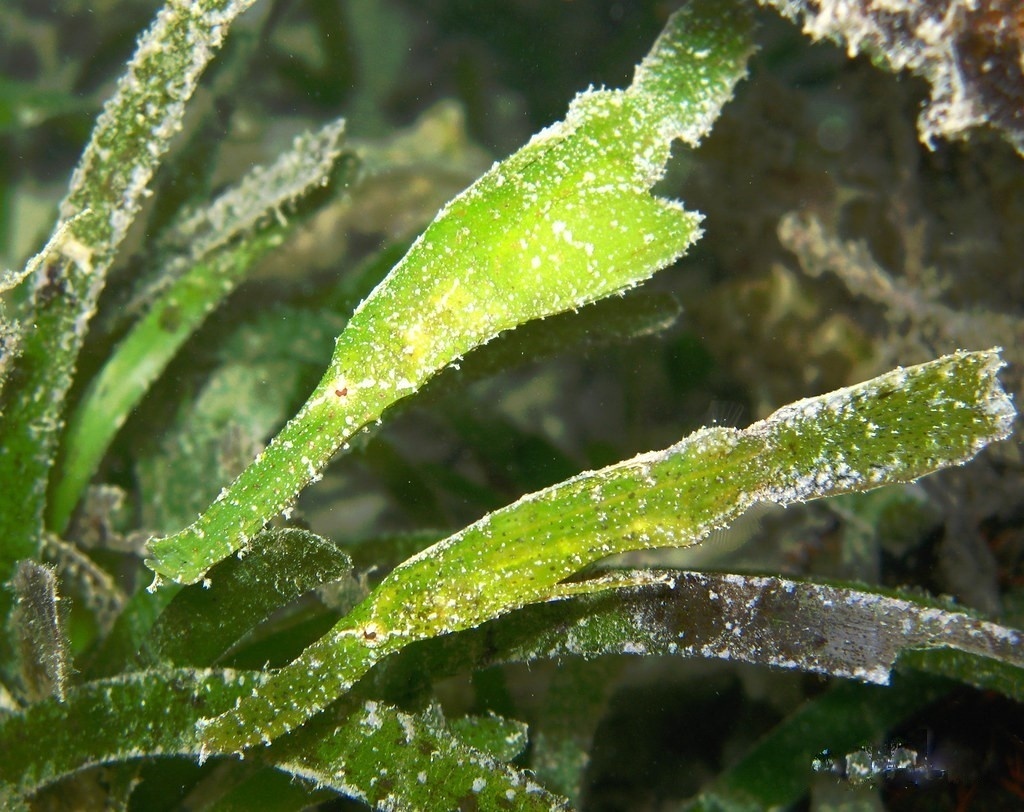}\\
			\vspace{0.01\linewidth}
                \includegraphics[width=1.15\linewidth,height=0.840\textwidth]{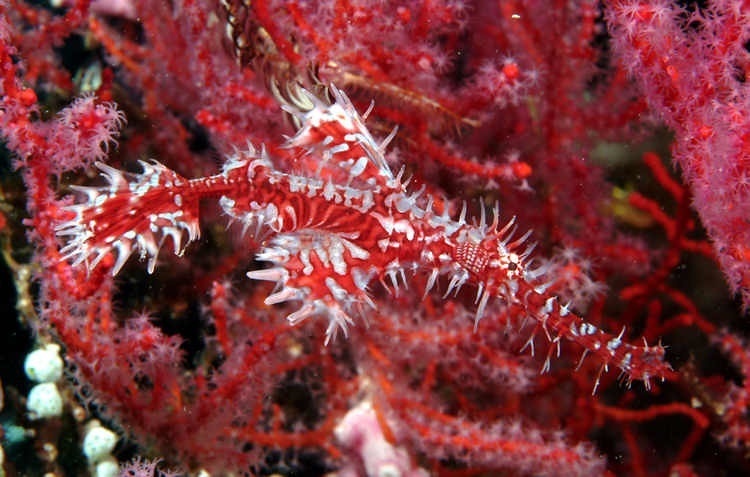}\\
			\vspace{0.01\linewidth}
                \includegraphics[width=1.15\linewidth,height=0.840\textwidth]{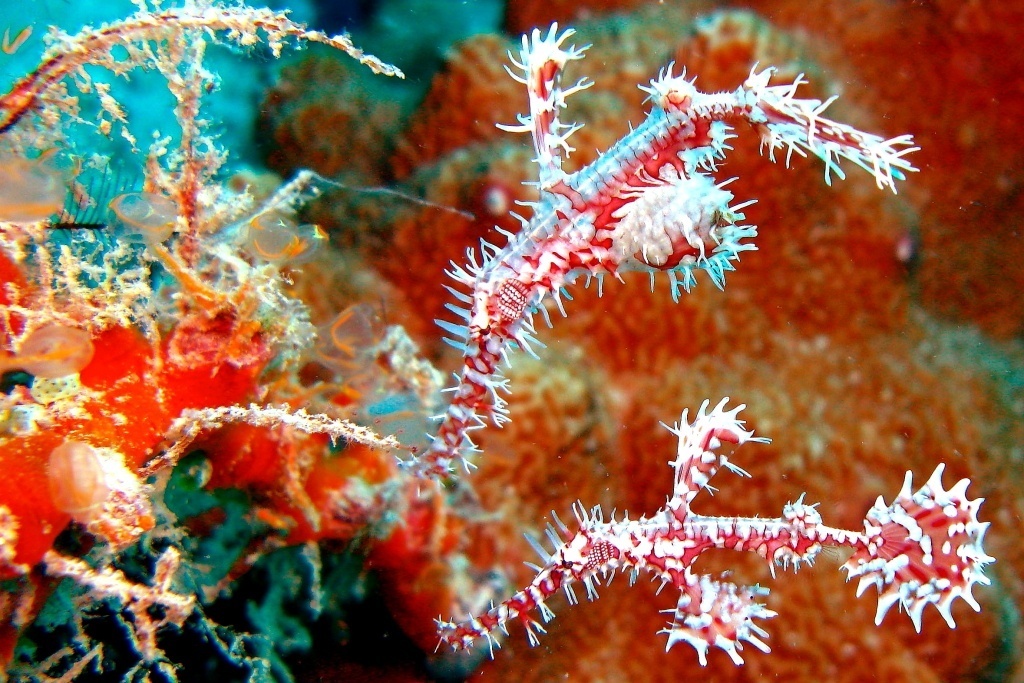}\\
			\vspace{0.01\linewidth}
                \includegraphics[width=1.15\linewidth,height=0.840\textwidth]{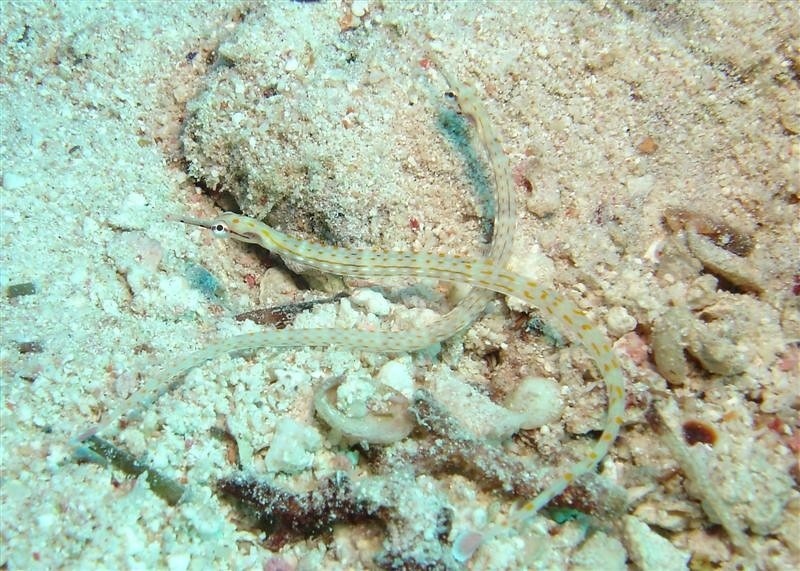}\\
			\vspace{0.01\linewidth}
			\includegraphics[width=1.15\linewidth,height=0.840\textwidth]{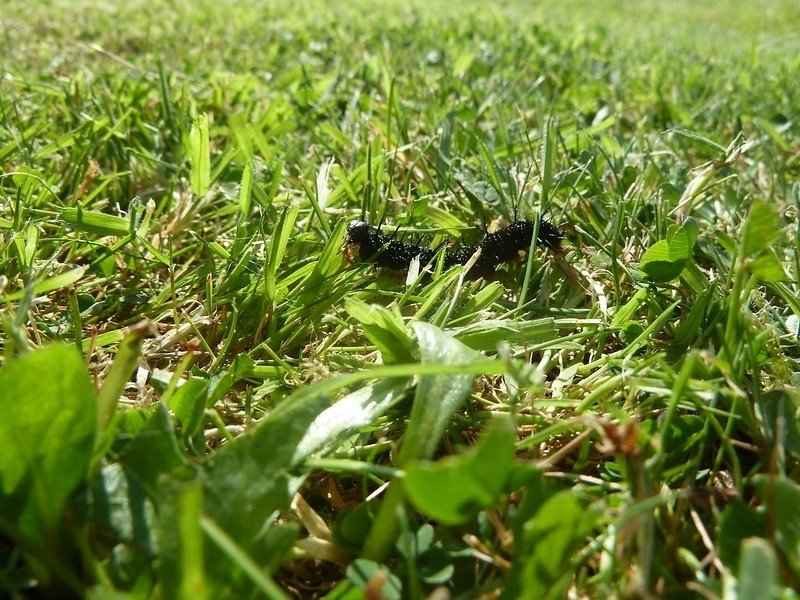}\\
			\vspace{0.01\linewidth}
                \includegraphics[width=1.15\textwidth,height=0.840\textwidth]{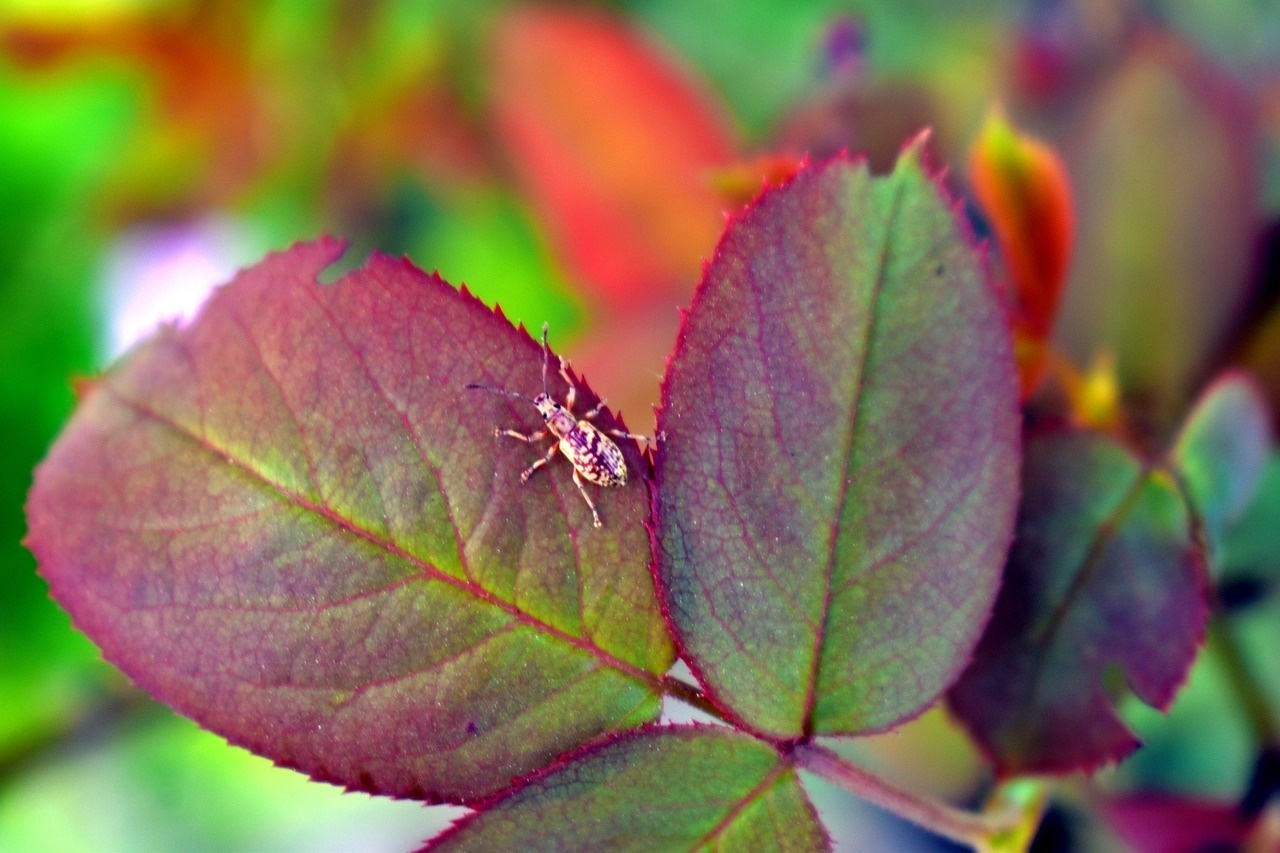}\\
                \vspace{0.01\linewidth}
			\includegraphics[width=1.15\linewidth,height=0.840\textwidth]{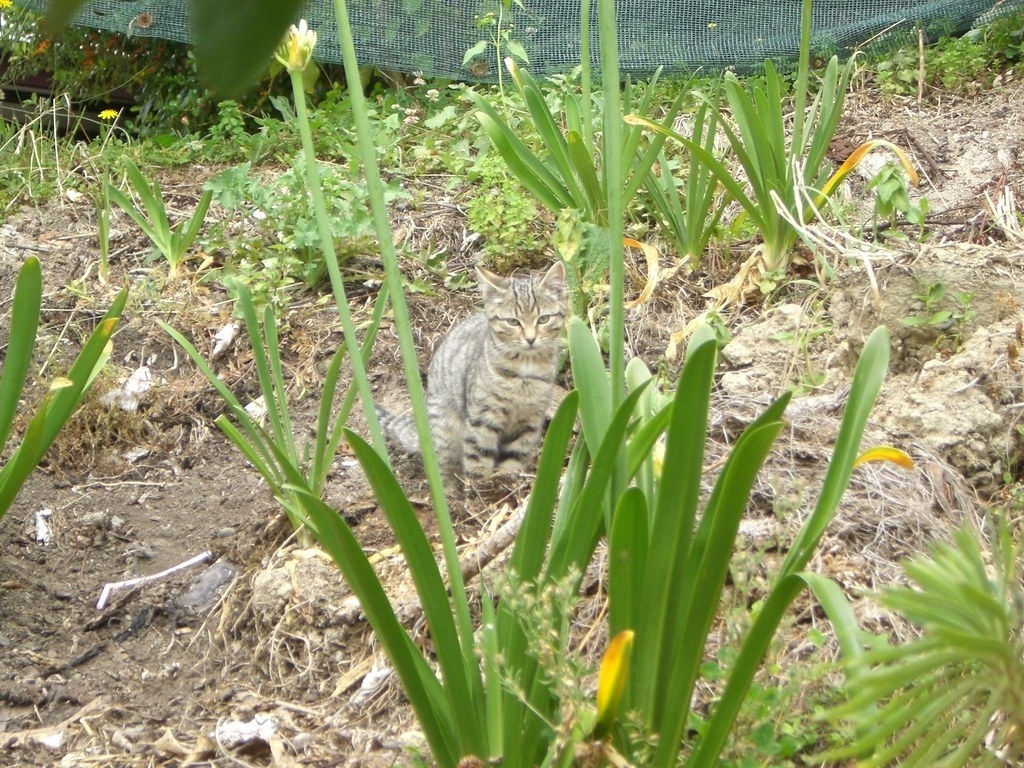}\\
			\vspace{0.08\linewidth}
		\end{minipage}%
	}\hspace{0.018\columnwidth}
	\subfigure[{\scriptsize GT}]{
		\begin{minipage}[t]{0.1\textwidth}
			\centering
			\includegraphics[width=1.15\linewidth,height=0.84\textwidth]{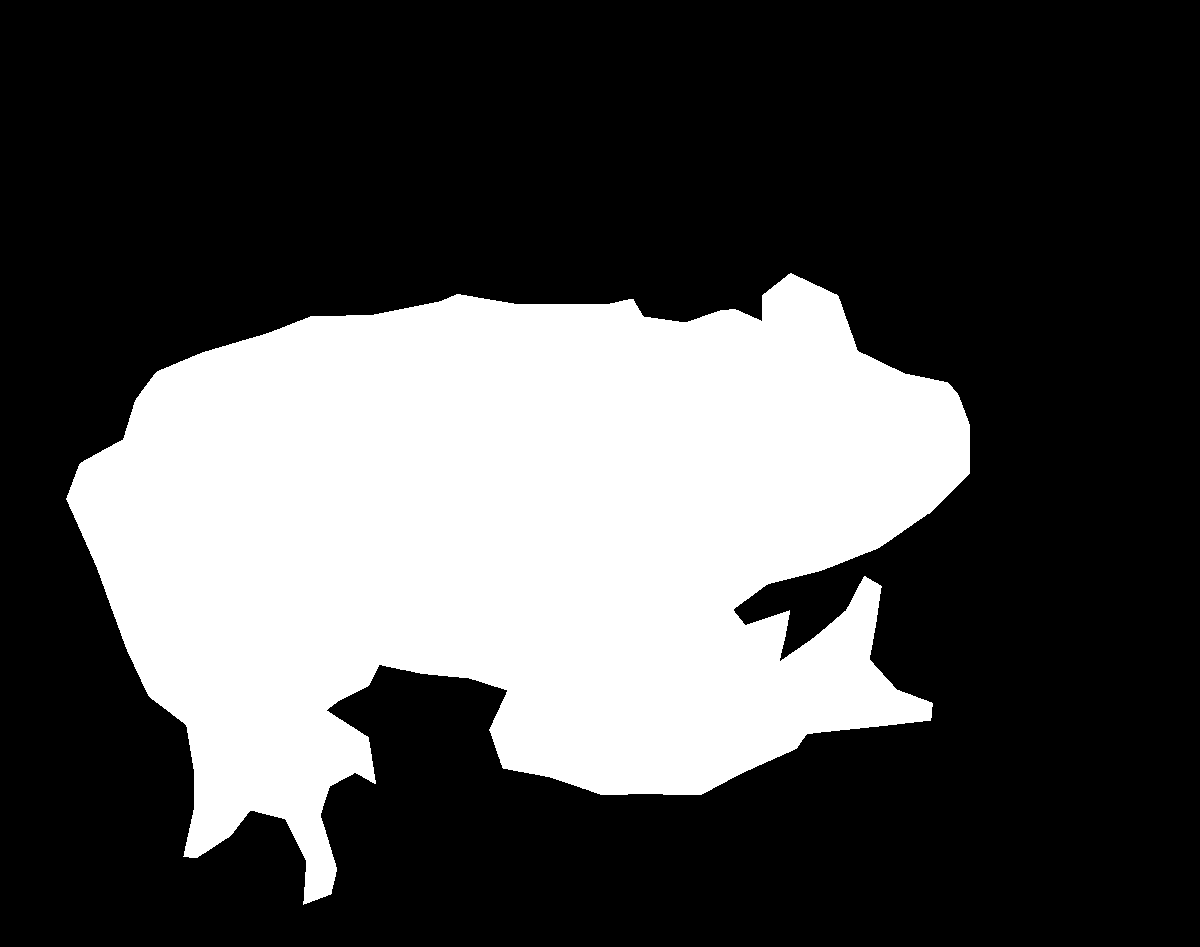}\\
			\vspace{0.01\linewidth}
                \includegraphics[width=1.15\textwidth,height=0.840\textwidth]{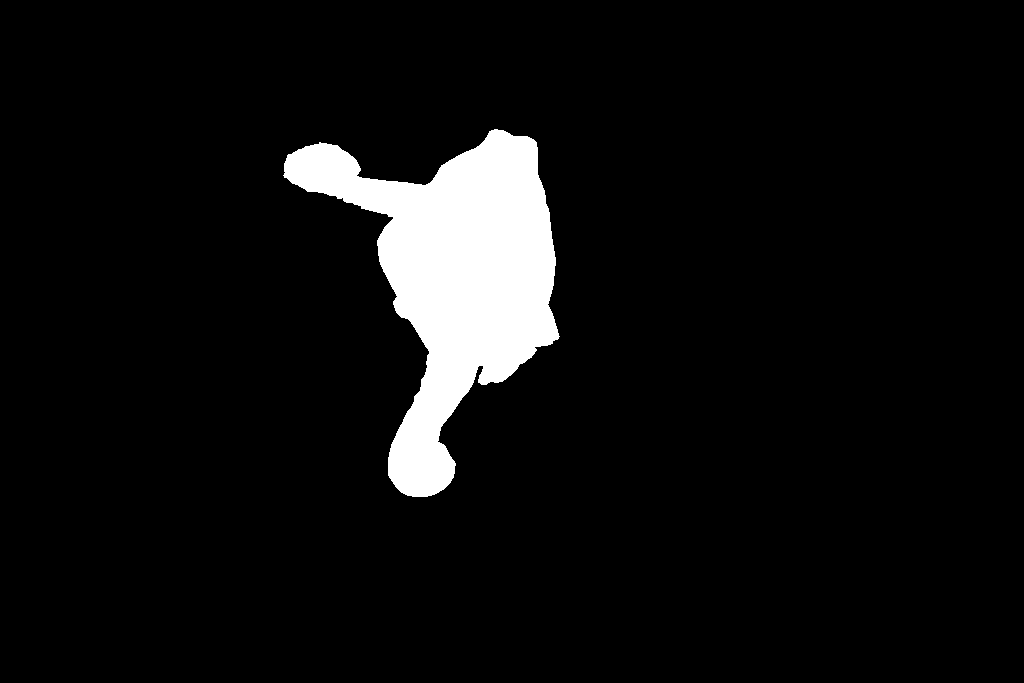}\\
			\vspace{0.01\linewidth}
                \includegraphics[width=1.15\linewidth,height=0.84\textwidth]{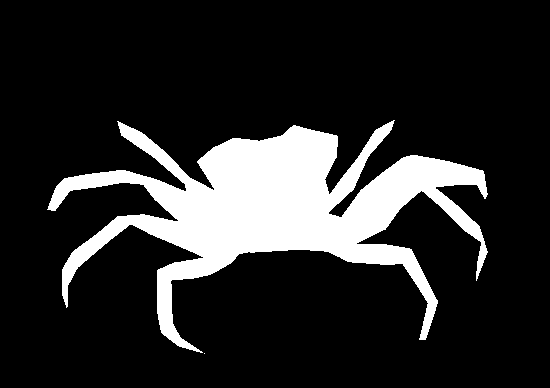}\\
                \vspace{0.01\linewidth}
			\includegraphics[width=1.15\linewidth,height=0.84\textwidth]{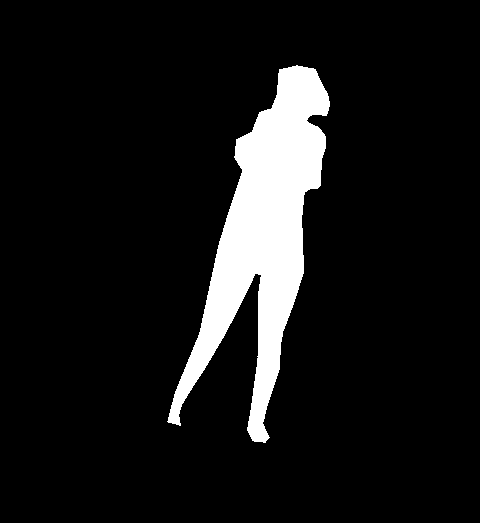}\\
			\vspace{0.01\linewidth}
                \includegraphics[width=1.15\linewidth,height=0.840\textwidth]{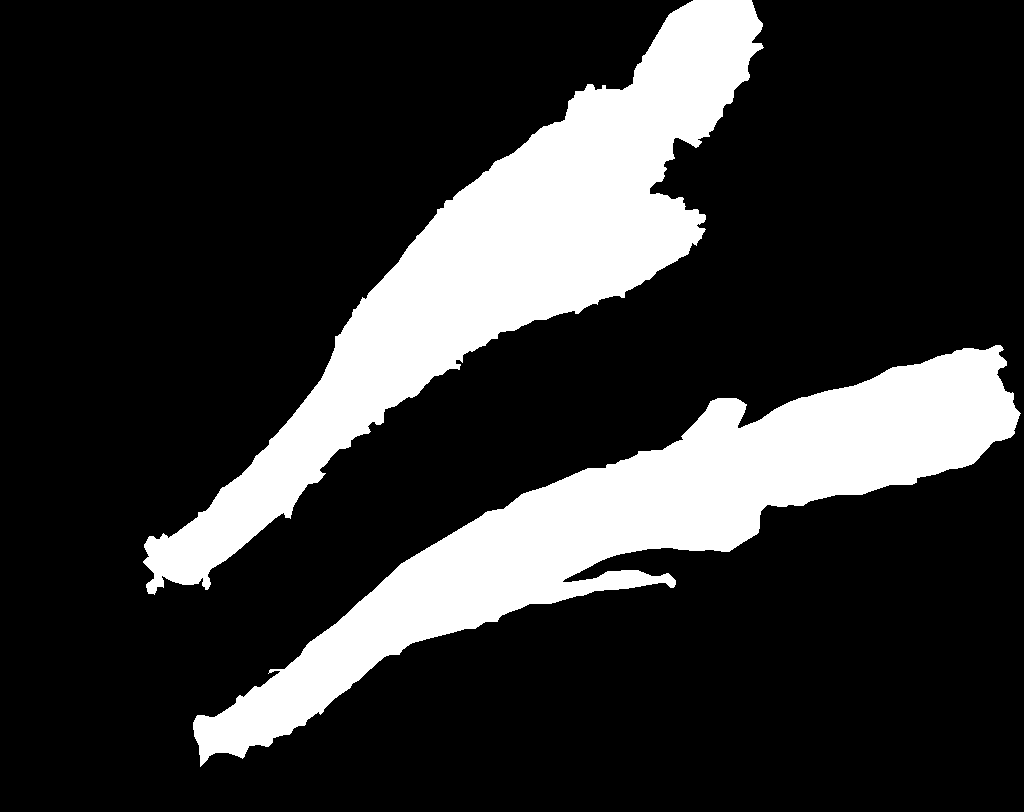}\\
			\vspace{0.01\linewidth}
                \includegraphics[width=1.15\linewidth,height=0.840\textwidth]{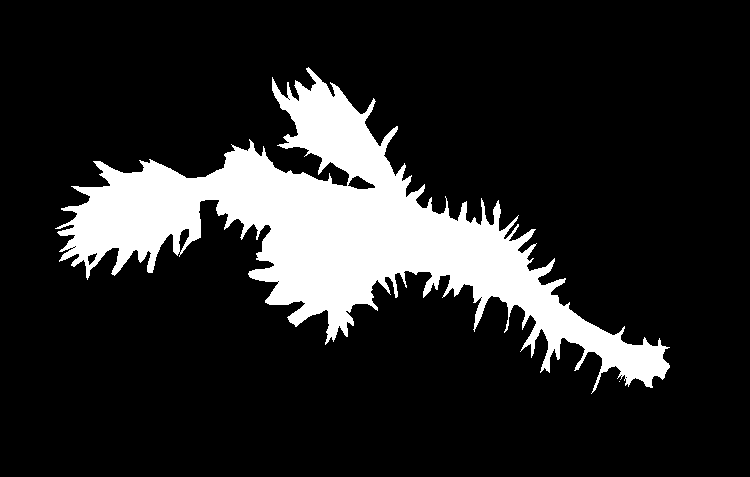}\\
			\vspace{0.01\linewidth}
                \includegraphics[width=1.15\linewidth,height=0.840\textwidth]{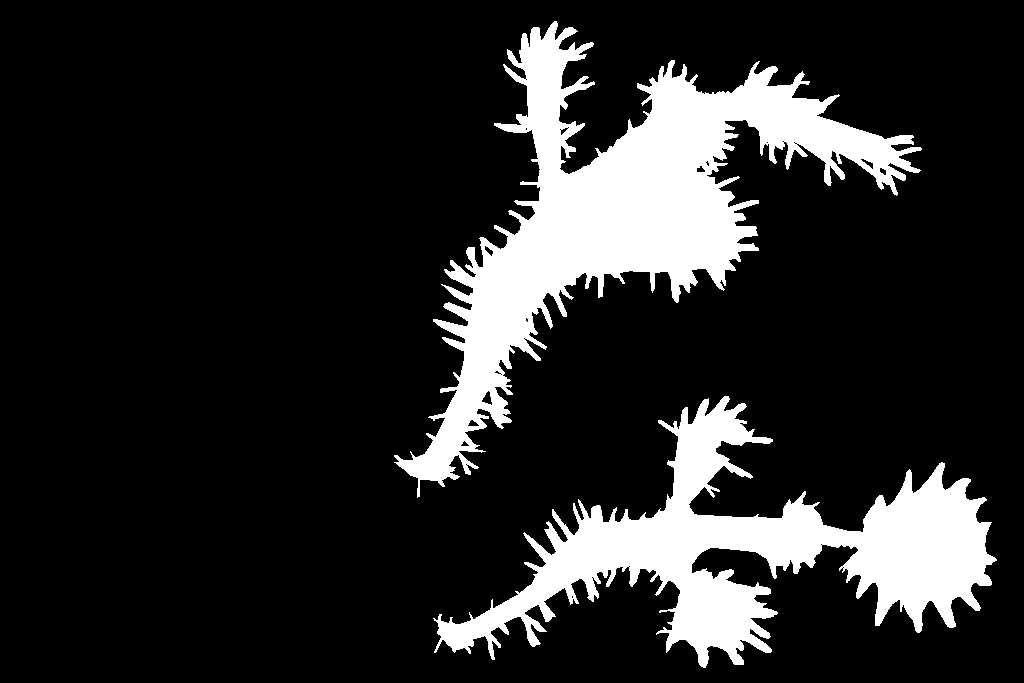}\\
			\vspace{0.01\linewidth}
			\includegraphics[width=1.15\linewidth,height=0.840\textwidth]{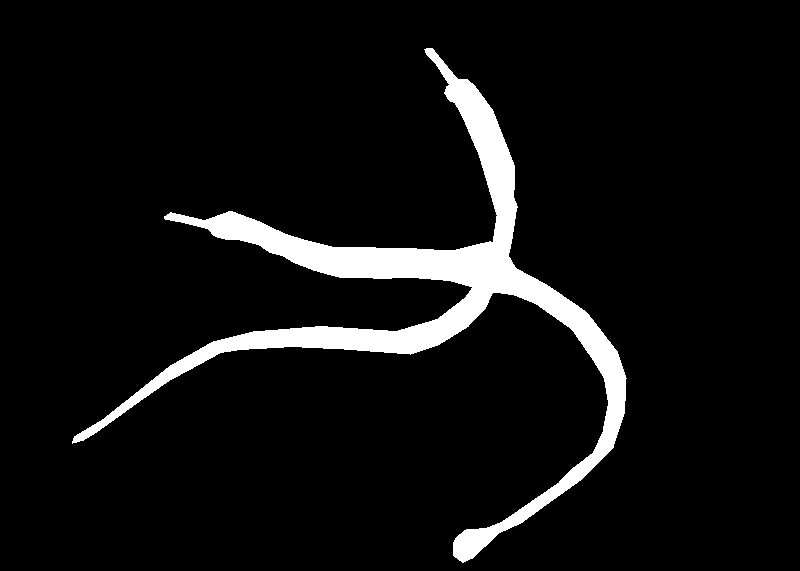}\\
			\vspace{0.01\linewidth}
			\includegraphics[width=1.15\linewidth,height=0.840\textwidth]{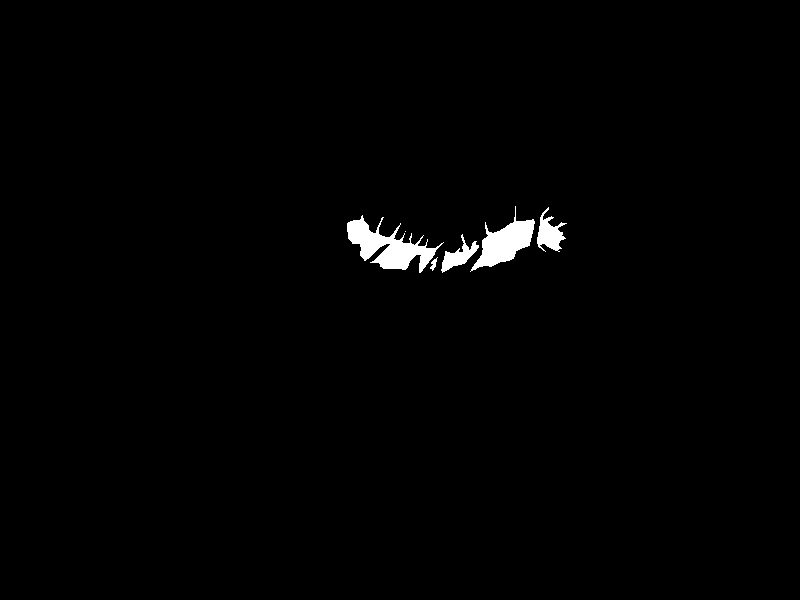}\\
			\vspace{0.01\linewidth}
                \includegraphics[width=1.15\textwidth,height=0.840\textwidth]{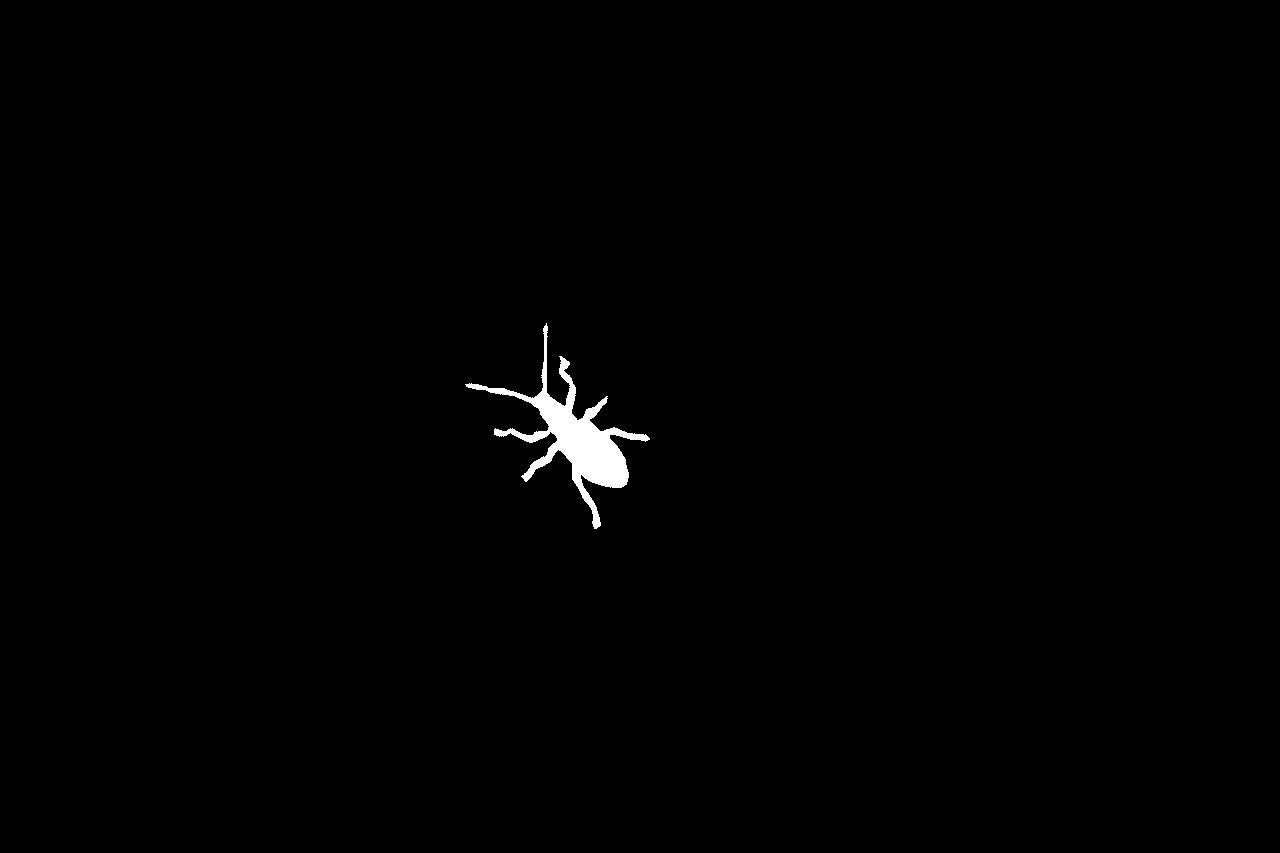}\\
                \vspace{0.01\linewidth}
			\includegraphics[width=1.15\linewidth,height=0.840\textwidth]{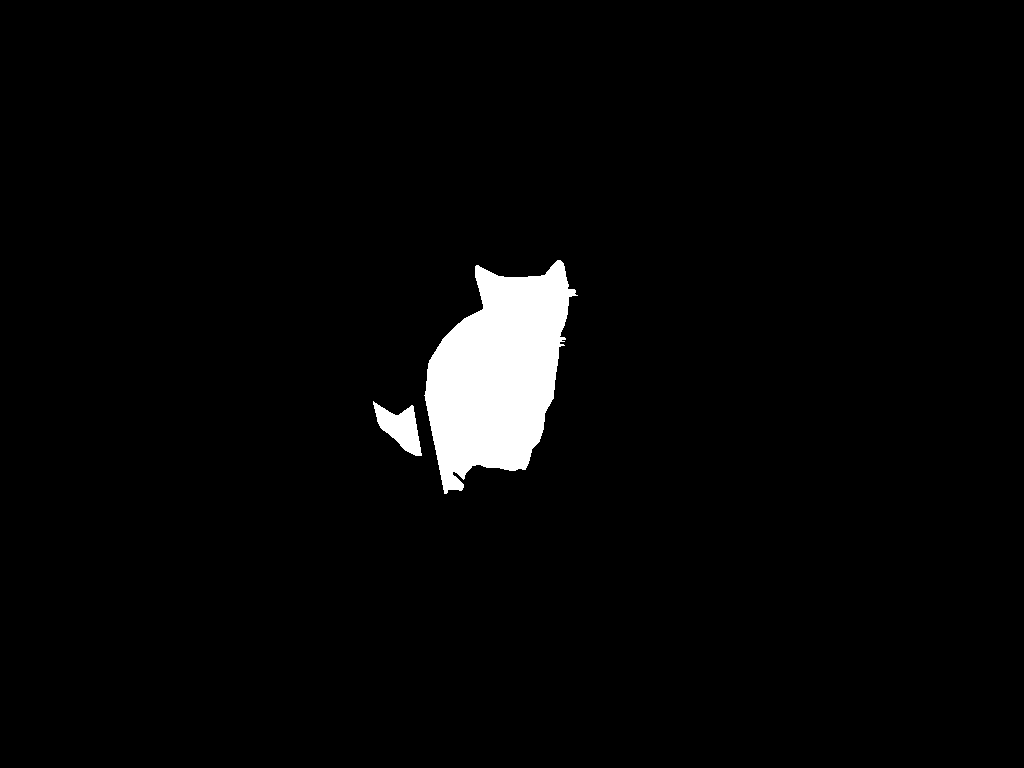}\\
			\vspace{0.08\linewidth}
		\end{minipage}%
	}\hspace{0.018\columnwidth}
	\subfigure[{\scriptsize Ours}]{
		\begin{minipage}[t]{0.1\textwidth}
			\centering
			\includegraphics[width=1.15\linewidth,height=0.84\textwidth]{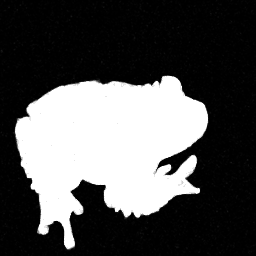}\\
			\vspace{0.01\linewidth}
                \includegraphics[width=1.15\textwidth,height=0.840\textwidth]{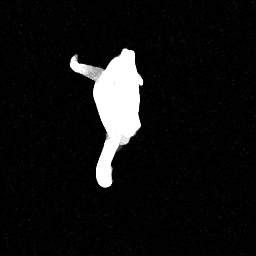}\\
			\vspace{0.01\linewidth}
                \includegraphics[width=1.15\linewidth,height=0.84\textwidth]{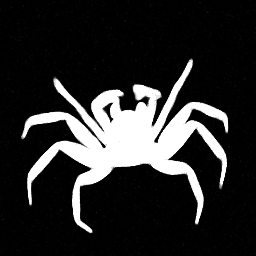}\\
                \vspace{0.01\linewidth}
			\includegraphics[width=1.15\linewidth,height=0.84\textwidth]{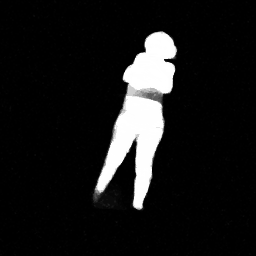}\\
			\vspace{0.01\linewidth}
                \includegraphics[width=1.15\linewidth,height=0.840\textwidth]{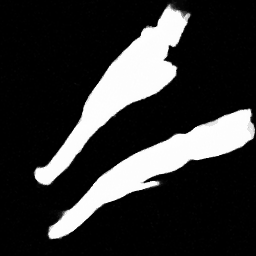}\\
			\vspace{0.01\linewidth}
                \includegraphics[width=1.15\linewidth,height=0.840\textwidth]{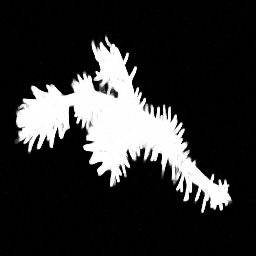}\\
			\vspace{0.01\linewidth}
                \includegraphics[width=1.15\linewidth,height=0.840\textwidth]{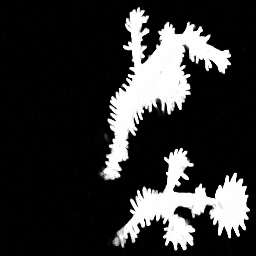}\\
	          \vspace{0.01\linewidth}
                \includegraphics[width=1.15\linewidth,height=0.840\textwidth]{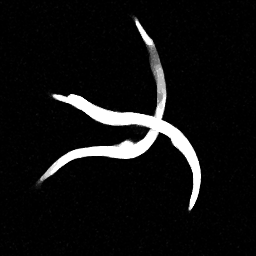}\\
			\vspace{0.01\linewidth}
			\includegraphics[width=1.15\linewidth,height=0.840\textwidth]{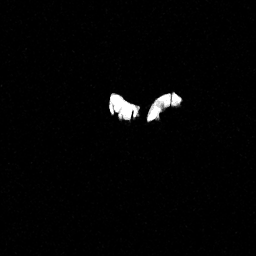}\\
			\vspace{0.01\linewidth}
                \includegraphics[width=1.15\textwidth,height=0.840\textwidth]{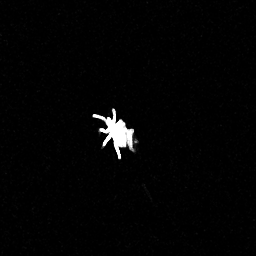}\\
                \vspace{0.01\linewidth}
			\includegraphics[width=1.15\linewidth,height=0.840\textwidth]{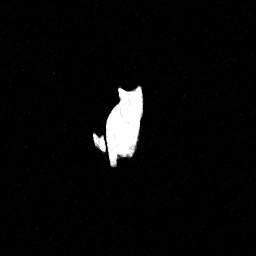}\\
			\vspace{0.08\linewidth}
		\end{minipage}%
	}\hspace{0.018\columnwidth}
	\subfigure[{\scriptsize CRNet~\cite{he2022weakly}}]{
		\begin{minipage}[t]{0.1\textwidth}
			\centering
			\includegraphics[width=1.15\linewidth,height=0.84\textwidth]{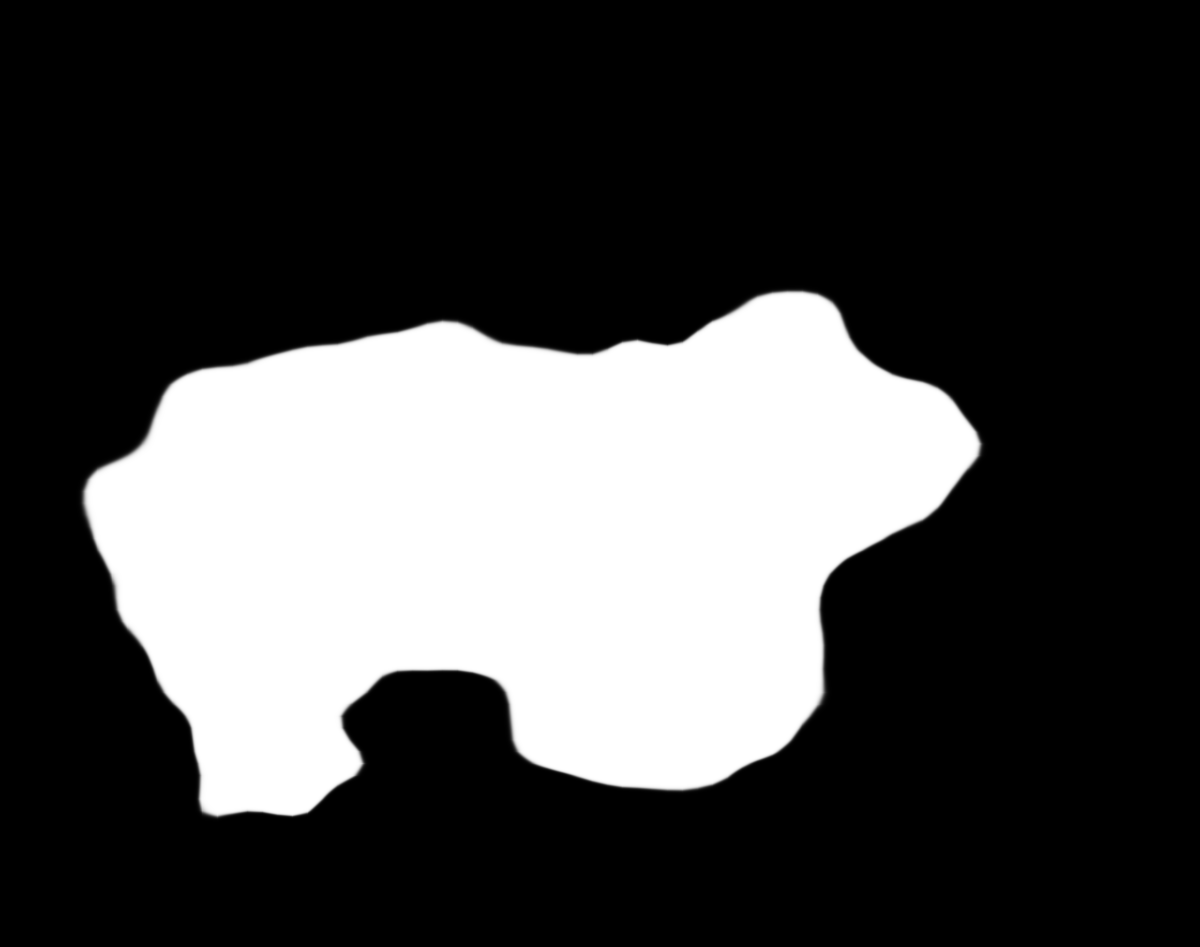}\\
                \includegraphics[width=1.15\textwidth,height=0.840\textwidth]{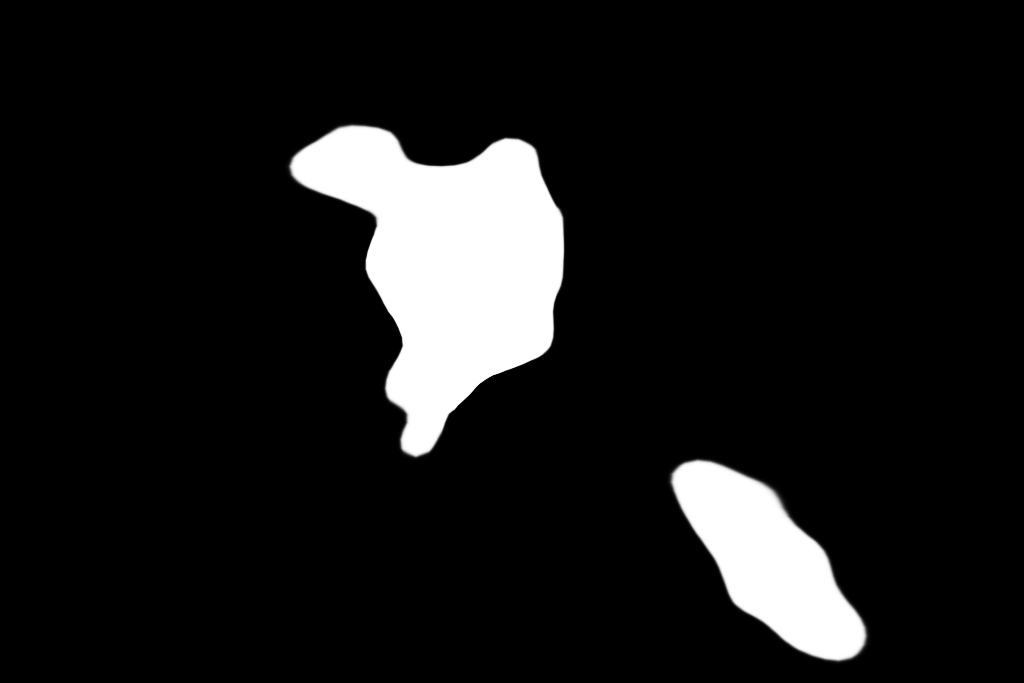}\\
			\vspace{0.01\linewidth}
			\vspace{0.01\linewidth}
                \includegraphics[width=1.15\linewidth,height=0.84\textwidth]{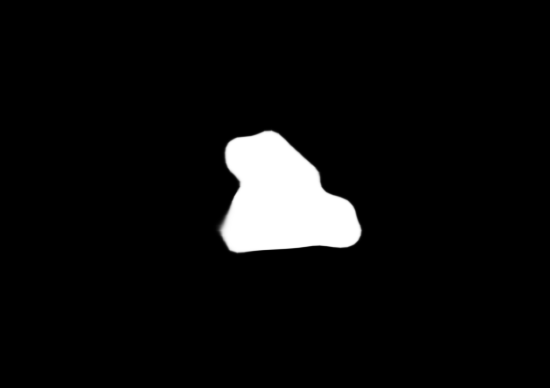}\\
                \vspace{0.01\linewidth}
			\includegraphics[width=1.15\linewidth,height=0.84\textwidth]{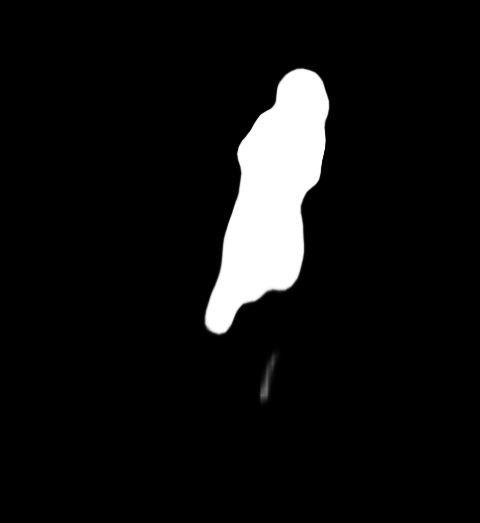}\\
			\vspace{0.01\linewidth}
                \includegraphics[width=1.15\linewidth,height=0.840\textwidth]{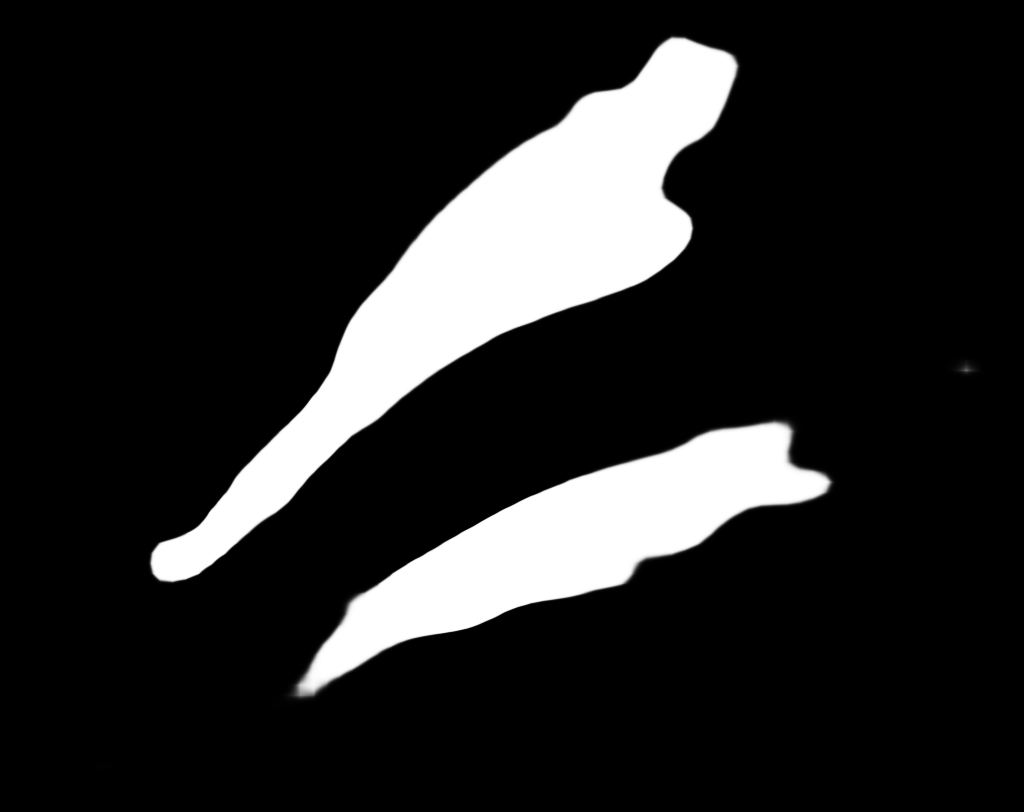}\\
			\vspace{0.01\linewidth}
                \includegraphics[width=1.15\linewidth,height=0.840\textwidth]{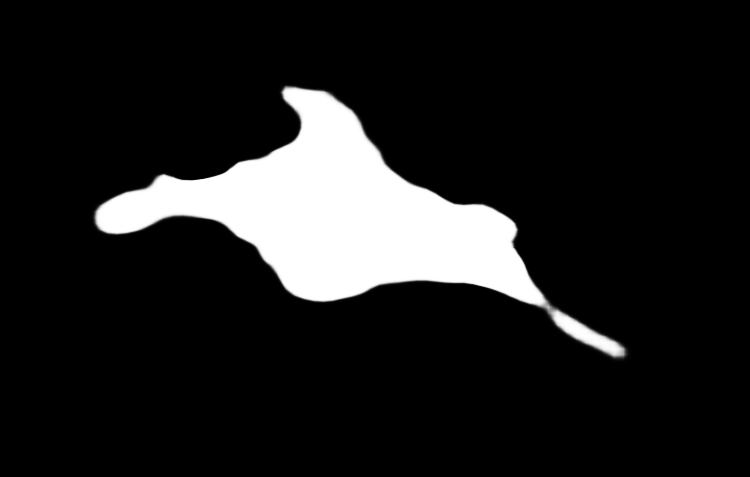}\\
			\vspace{0.01\linewidth}
                \includegraphics[width=1.15\linewidth,height=0.840\textwidth]{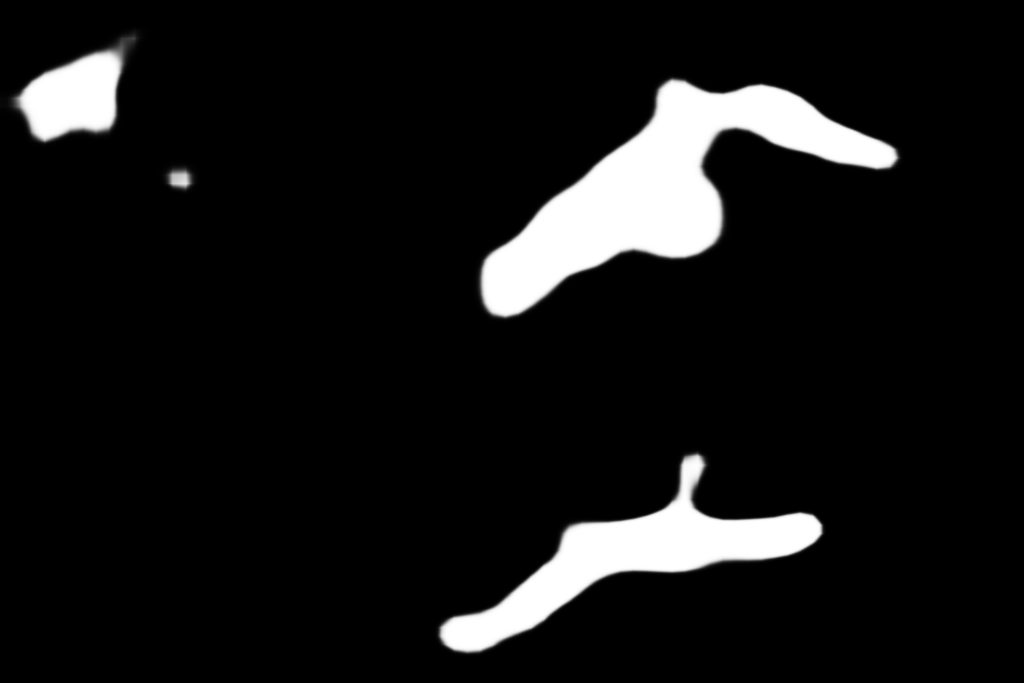}\\
			\vspace{0.01\linewidth}
                \includegraphics[width=1.15\linewidth,height=0.840\textwidth]{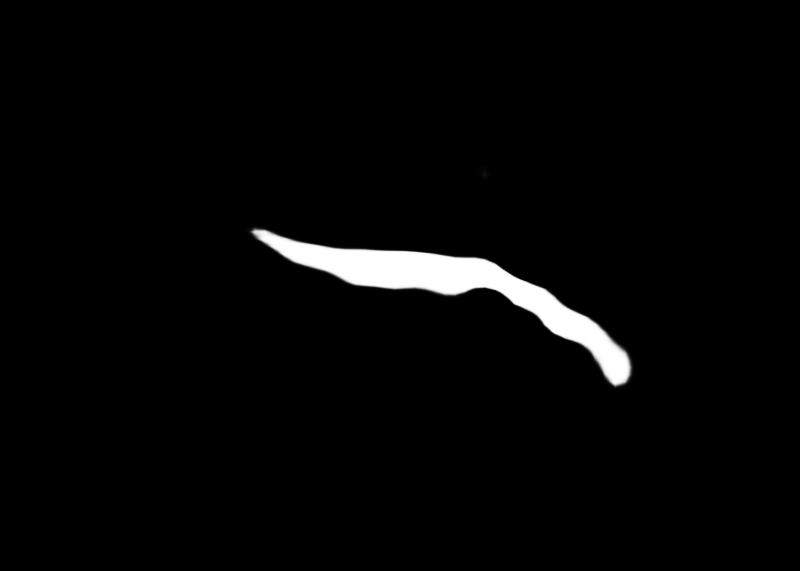}\\
			\vspace{0.01\linewidth}
			\includegraphics[width=1.15\linewidth,height=0.840\textwidth]{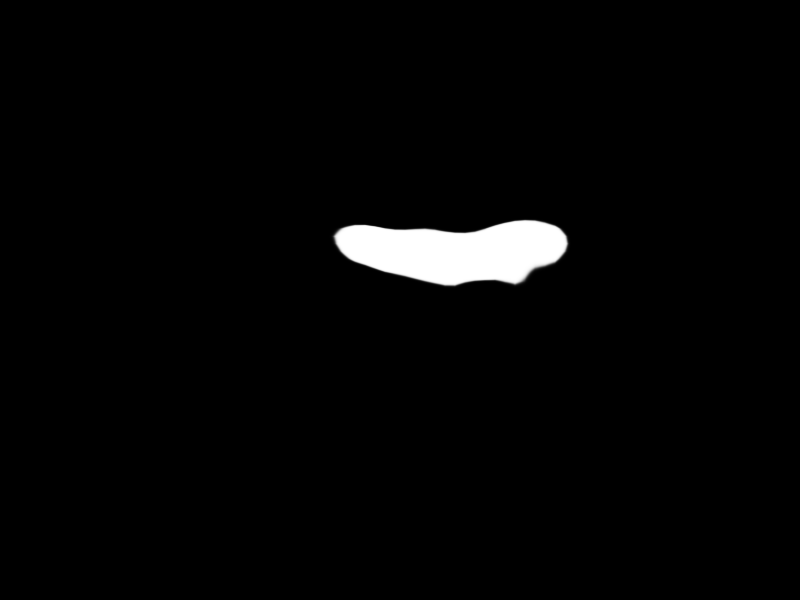}\\
			\vspace{0.01\linewidth}
                \includegraphics[width=1.15\textwidth,height=0.840\textwidth]{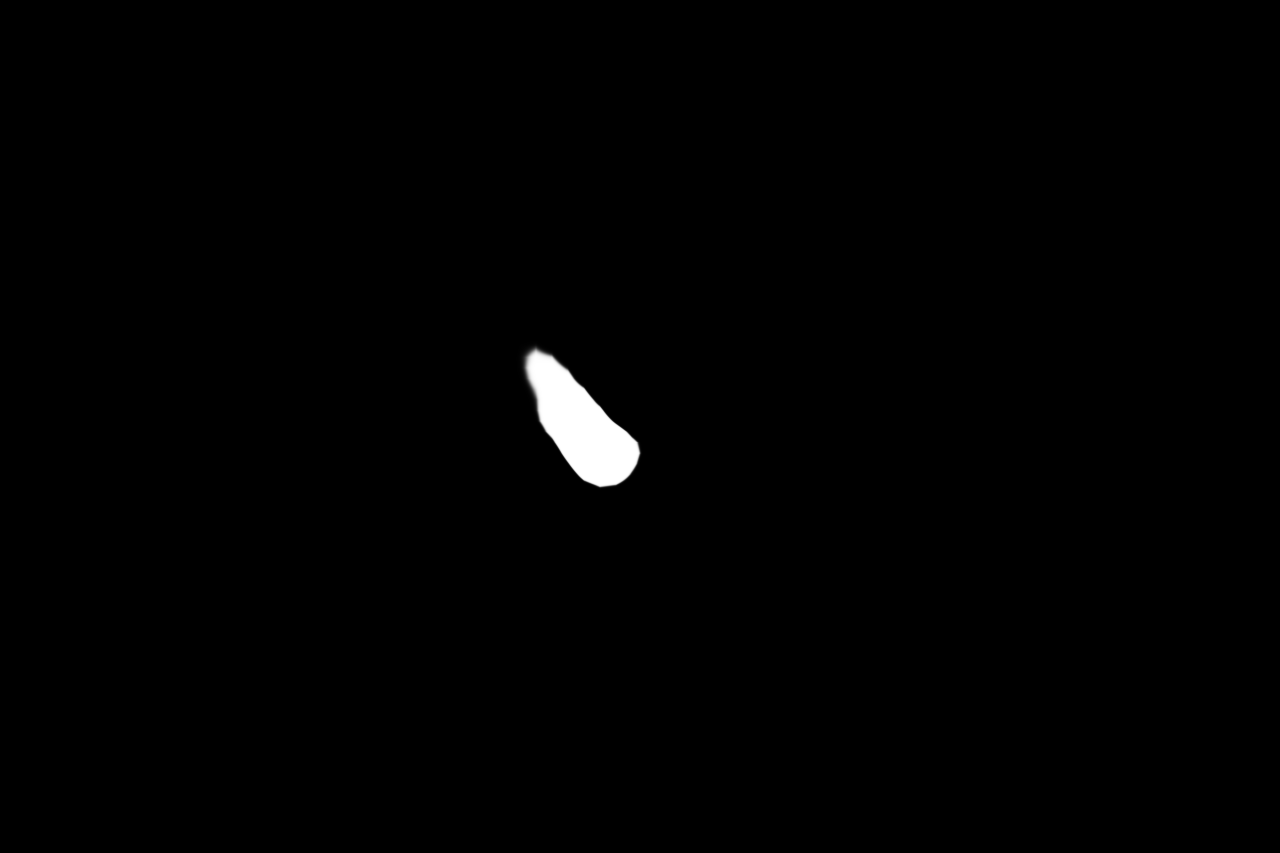}\\
                \vspace{0.01\linewidth}
			\includegraphics[width=1.15\linewidth,height=0.840\textwidth]{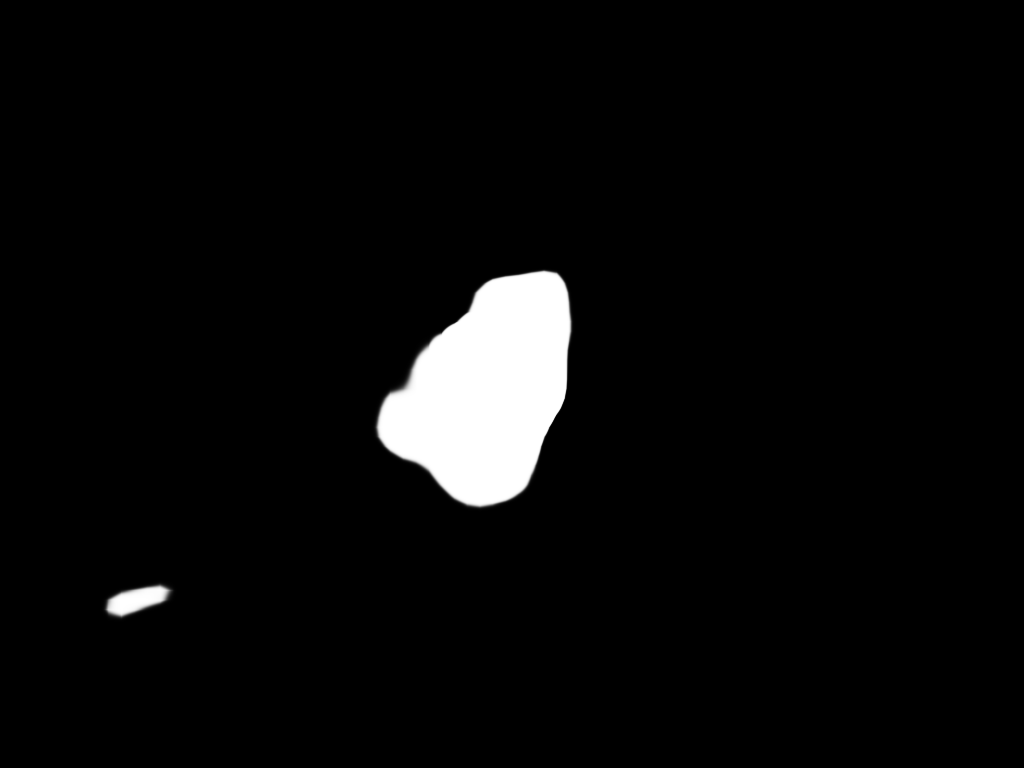}\\
			\vspace{0.08\linewidth}
		\end{minipage}%
	}\hspace{0.018\columnwidth}
	\subfigure[{\scriptsize CubeNet~\cite{zhong2022detecting}}]{
		\begin{minipage}[t]{0.1\textwidth}
			\centering
			\includegraphics[width=1.15\linewidth,height=0.84\textwidth]{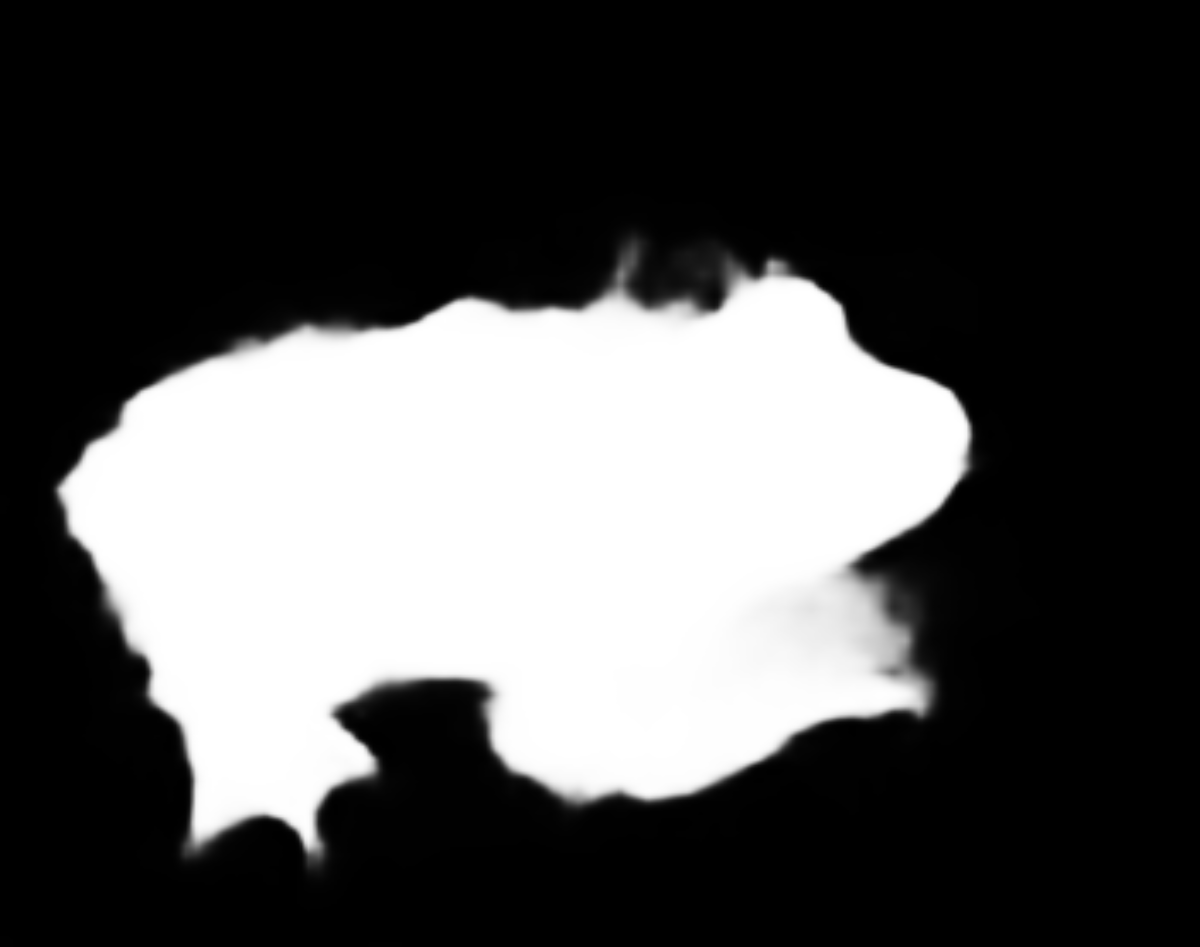}\\
			\vspace{0.01\linewidth}
                \includegraphics[width=1.15\textwidth,height=0.840\textwidth]{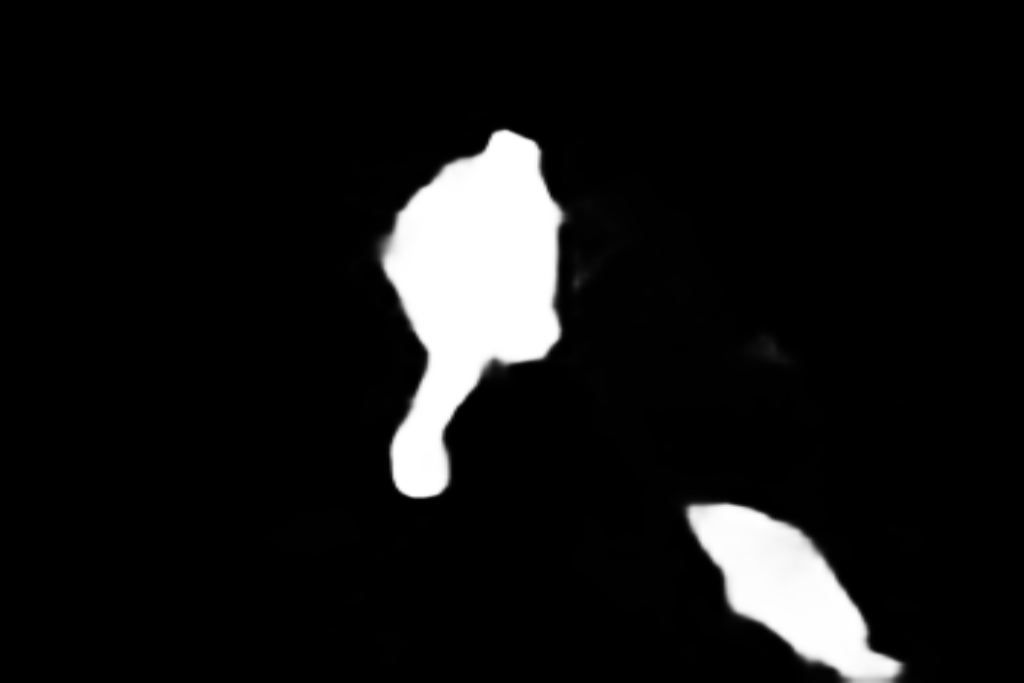}\\
			\vspace{0.01\linewidth}
                \includegraphics[width=1.15\linewidth,height=0.84\textwidth]{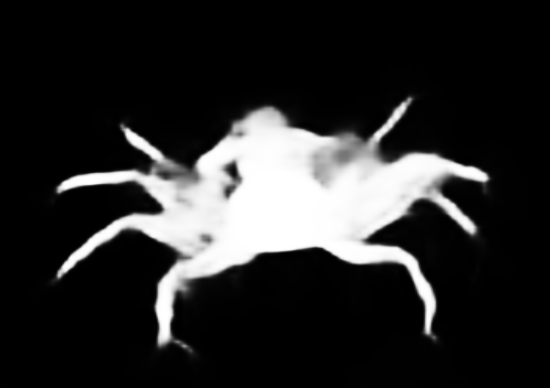}\\
                \vspace{0.01\linewidth}
			\includegraphics[width=1.15\linewidth,height=0.84\textwidth]{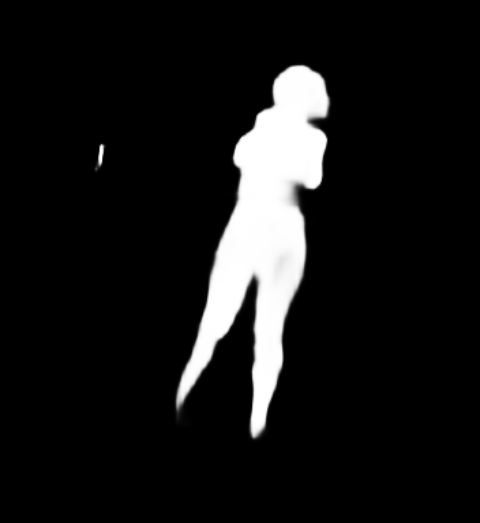}\\
			\vspace{0.01\linewidth}
                \includegraphics[width=1.15\linewidth,height=0.840\textwidth]{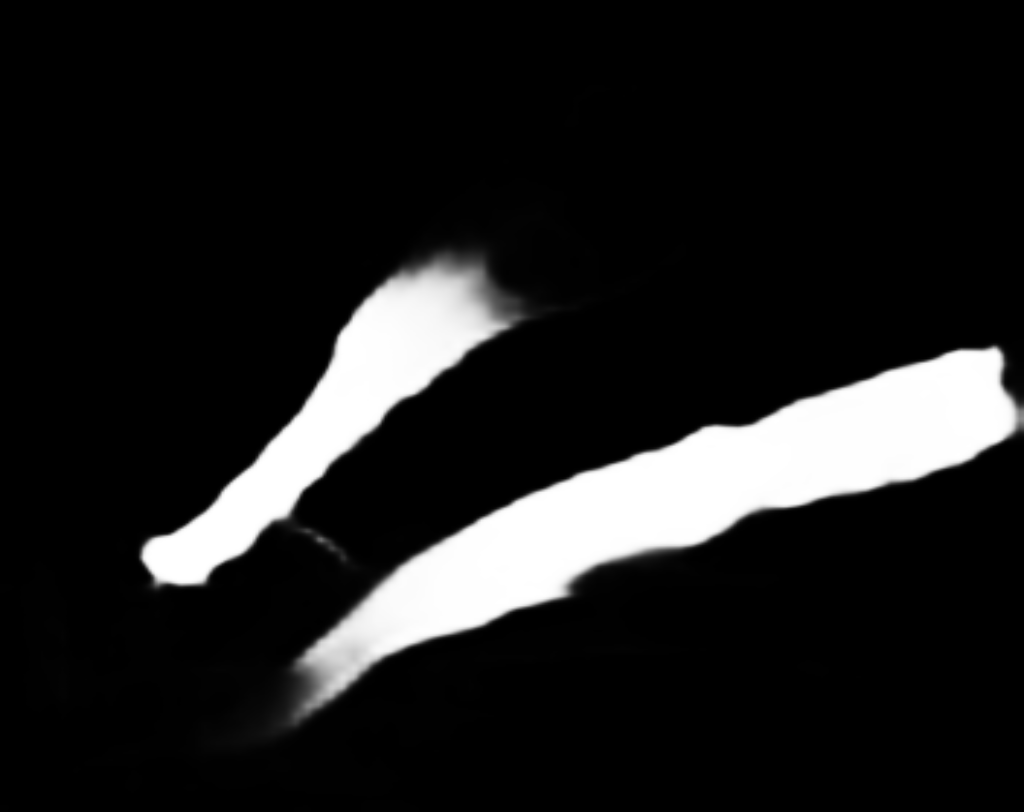}\\
			\vspace{0.01\linewidth}
                \includegraphics[width=1.15\linewidth,height=0.840\textwidth]{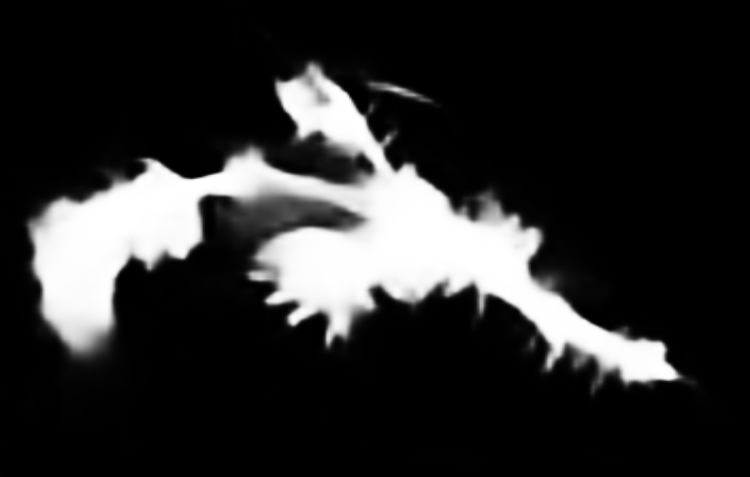}\\
			\vspace{0.01\linewidth}
                \includegraphics[width=1.15\linewidth,height=0.840\textwidth]{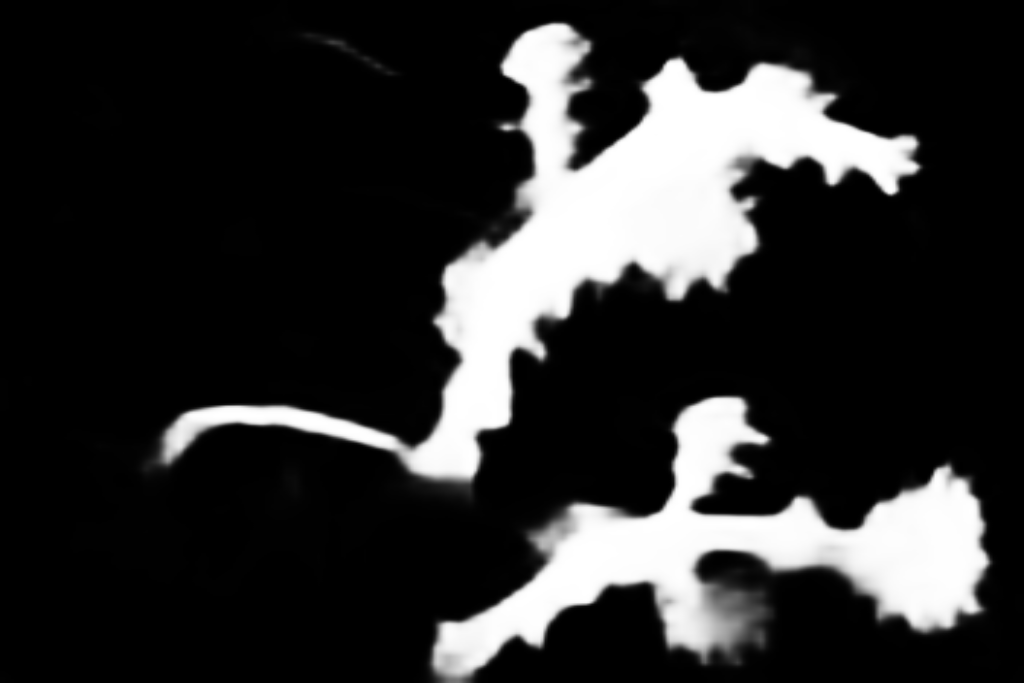}\\
			\vspace{0.01\linewidth}
			\includegraphics[width=1.15\linewidth,height=0.840\textwidth]{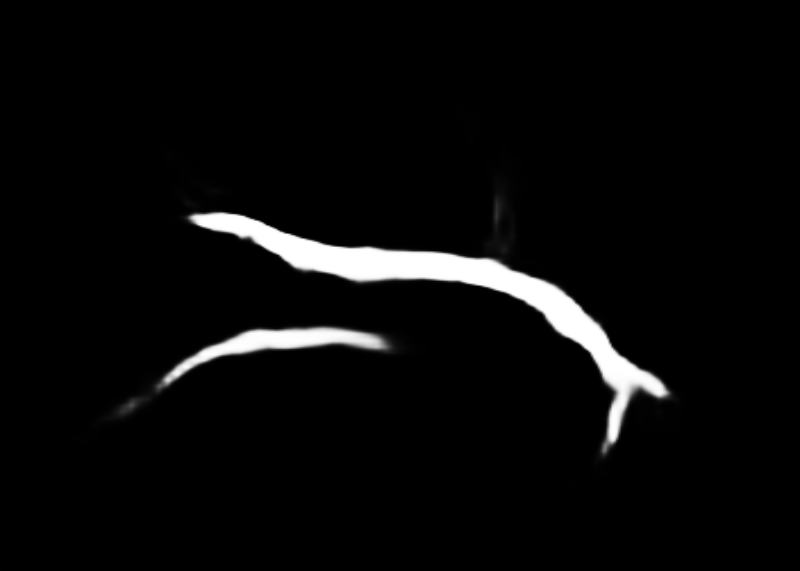}\\
			\vspace{0.01\linewidth}
			\includegraphics[width=1.15\linewidth,height=0.840\textwidth]{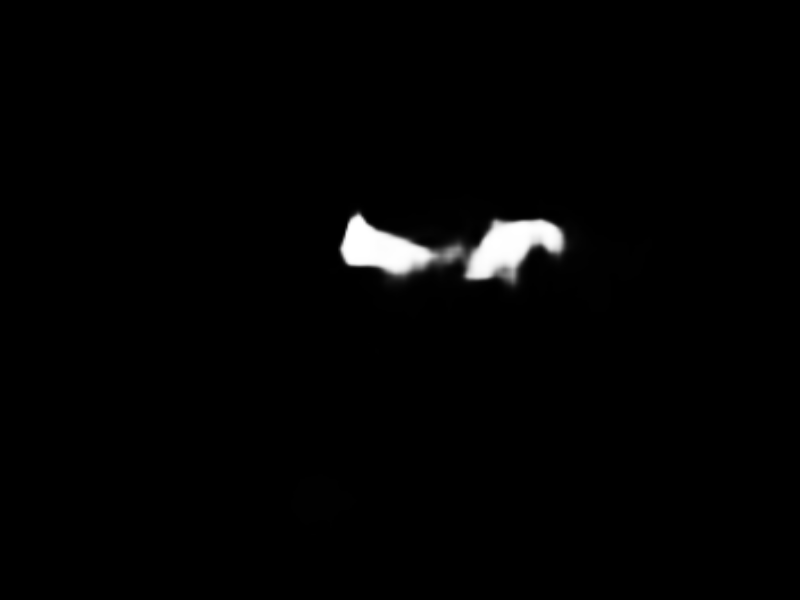}\\
			\vspace{0.01\linewidth}
                \includegraphics[width=1.15\textwidth,height=0.840\textwidth]{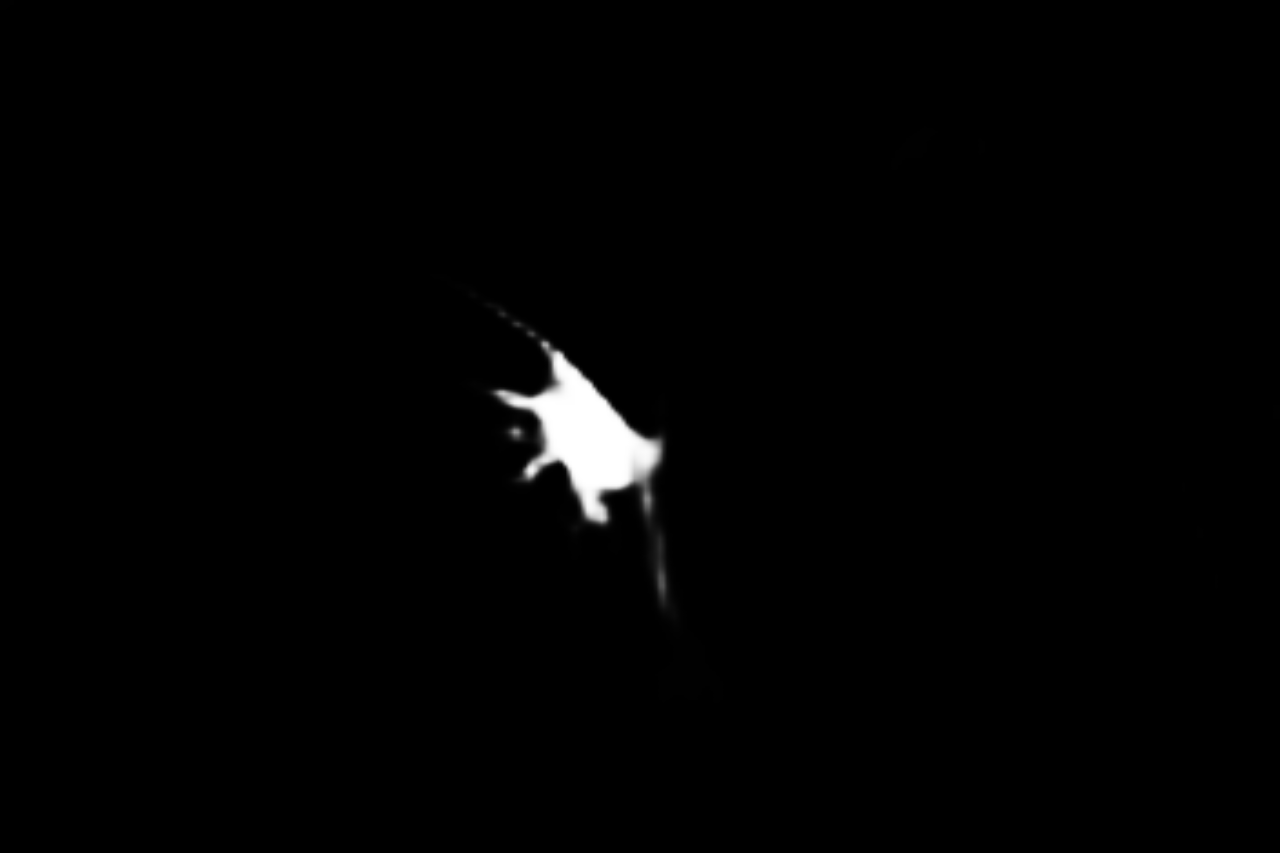}\\
                \vspace{0.01\linewidth}
			\includegraphics[width=1.15\linewidth,height=0.840\textwidth]{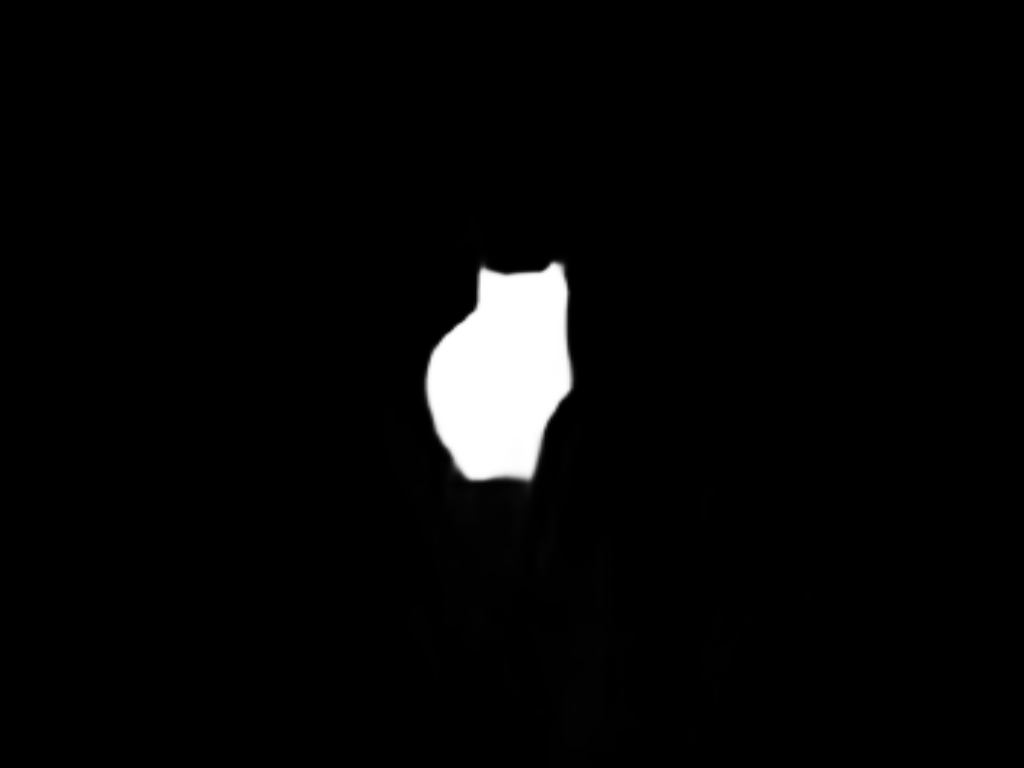}\\
			\vspace{0.08\linewidth}
		\end{minipage}%
	}\hspace{0.018\columnwidth}
	\subfigure[{\scriptsize PFNet~\cite{mei2021camouflaged}}]{
		\begin{minipage}[t]{0.1\textwidth}
			\centering
			\includegraphics[width=1.15\linewidth,height=0.84\textwidth]{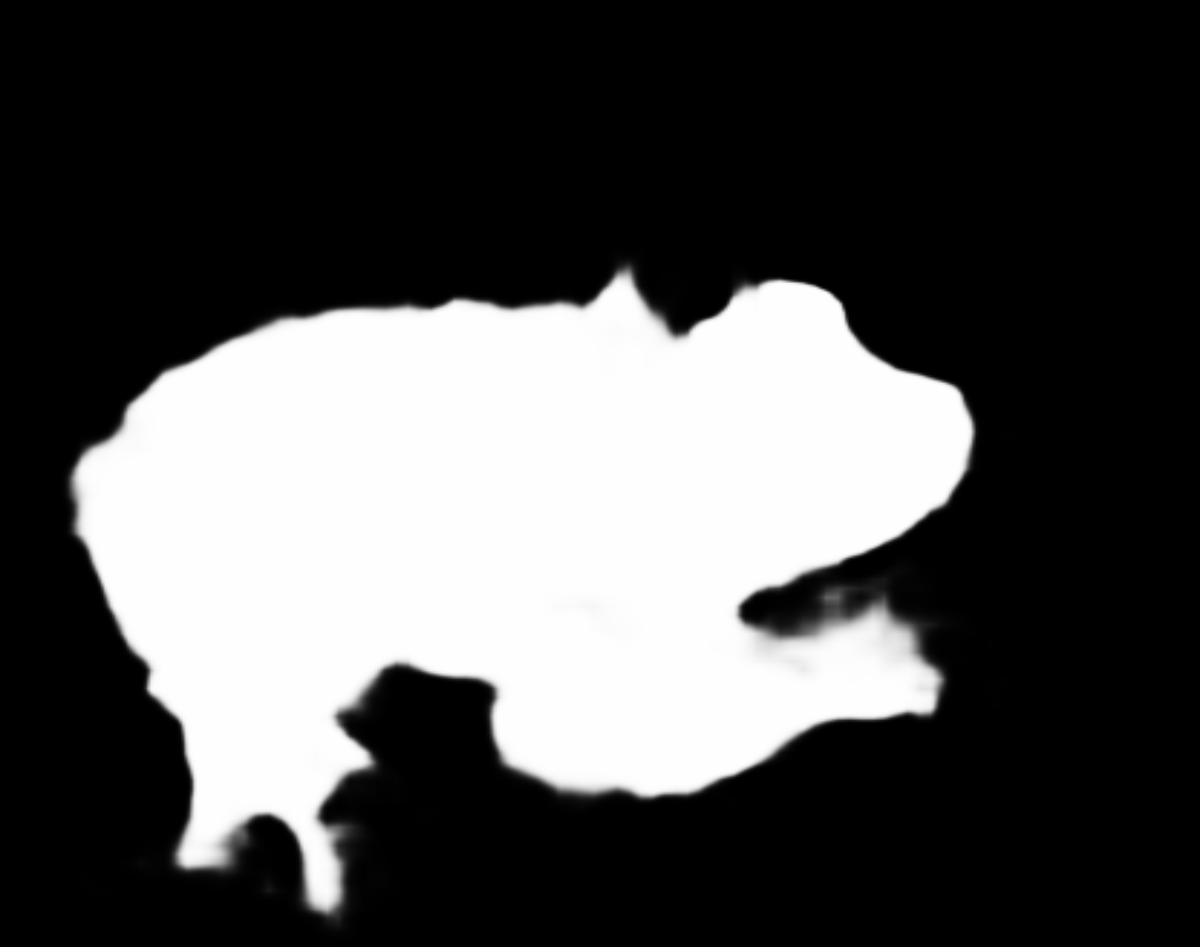}\\
			\vspace{0.01\linewidth}
                \includegraphics[width=1.15\textwidth,height=0.840\textwidth]{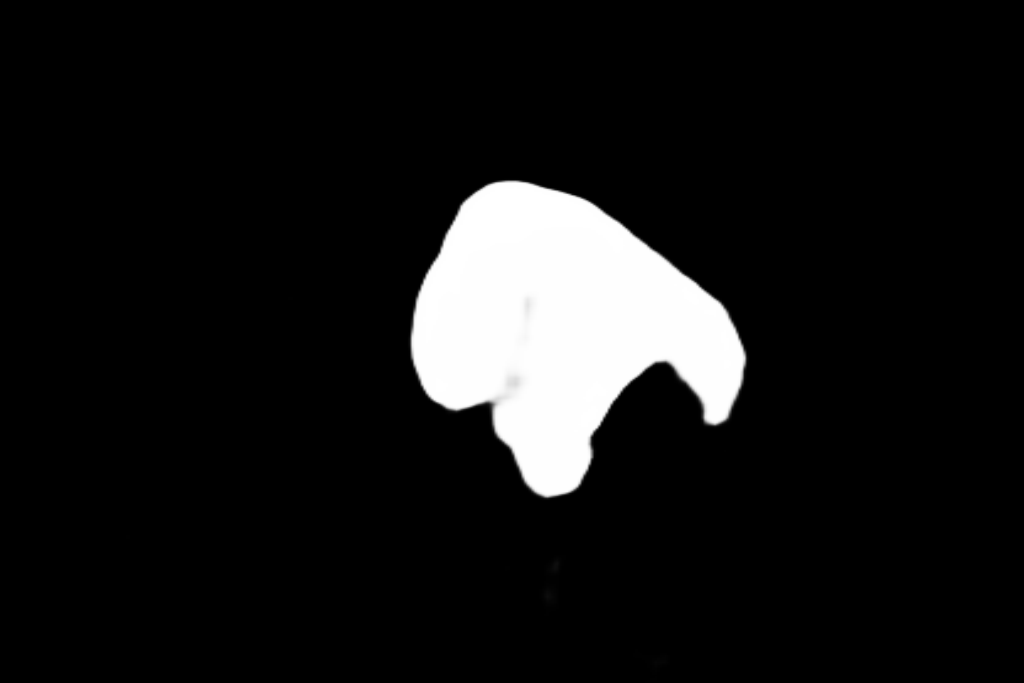}\\
                \vspace{0.01\linewidth}
                \includegraphics[width=1.15\linewidth,height=0.84\textwidth]{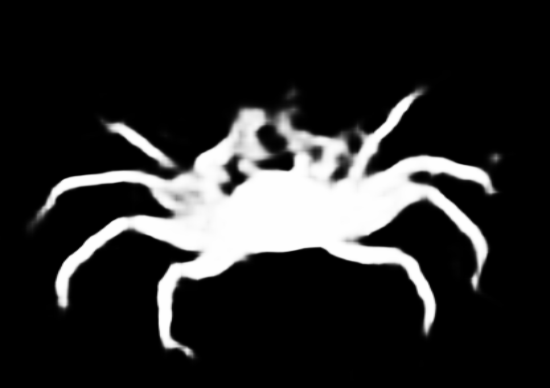}\\
                \vspace{0.01\linewidth}
			\includegraphics[width=1.15\linewidth,height=0.84\textwidth]{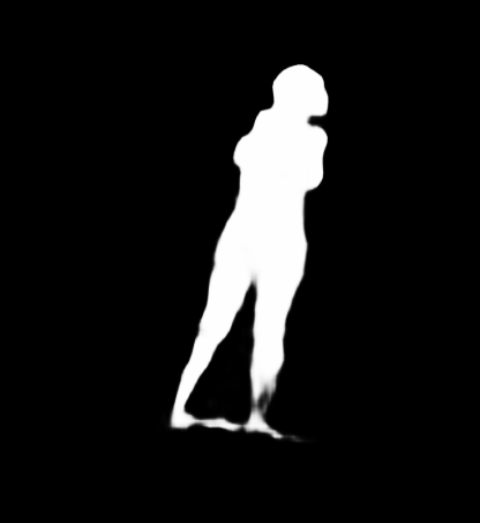}\\
			\vspace{0.01\linewidth}
                \includegraphics[width=1.15\linewidth,height=0.840\textwidth]{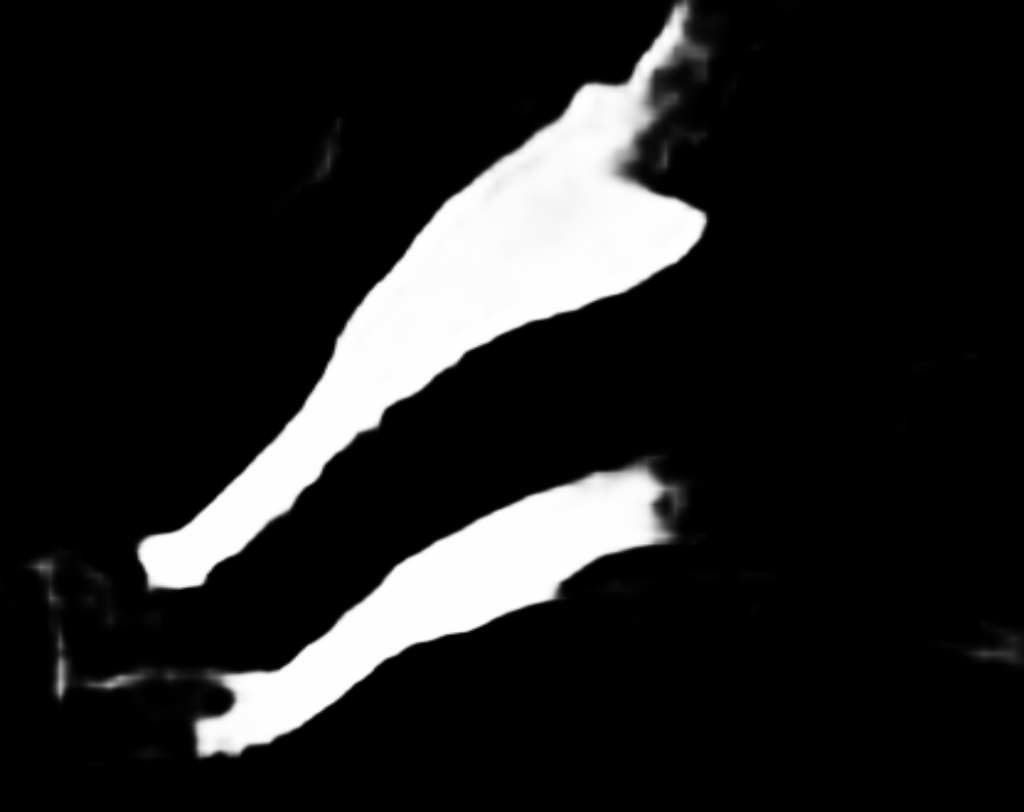}\\
			\vspace{0.01\linewidth}
                \includegraphics[width=1.15\linewidth,height=0.840\textwidth]{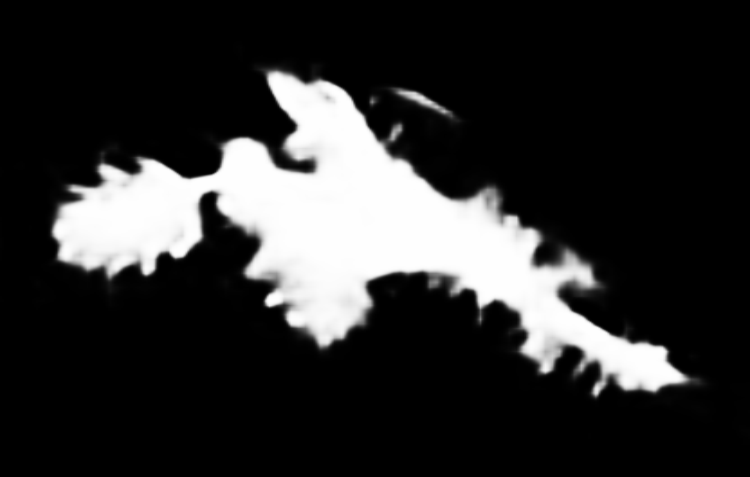}\\
			\vspace{0.01\linewidth}
                \includegraphics[width=1.15\linewidth,height=0.840\textwidth]{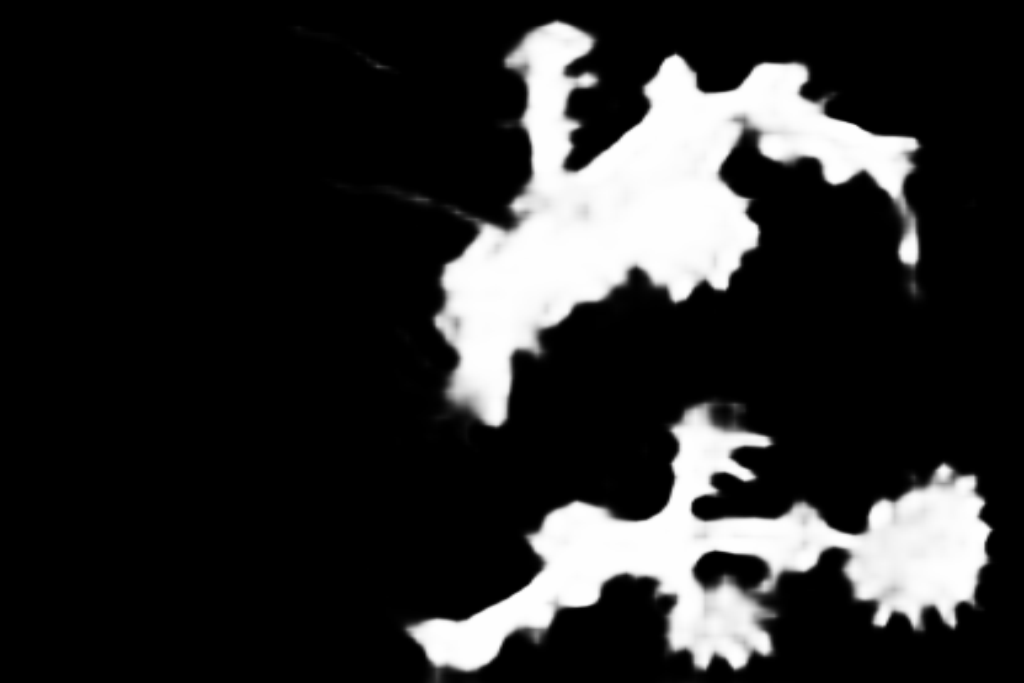}\\
			\vspace{0.01\linewidth}
			\includegraphics[width=1.15\linewidth,height=0.840\textwidth]{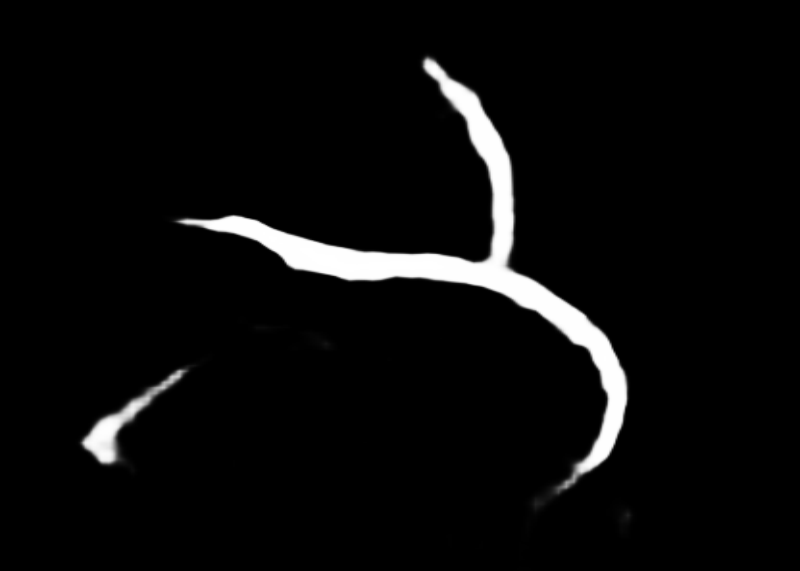}\\
			\vspace{0.01\linewidth}
			\includegraphics[width=1.15\linewidth,height=0.840\textwidth]{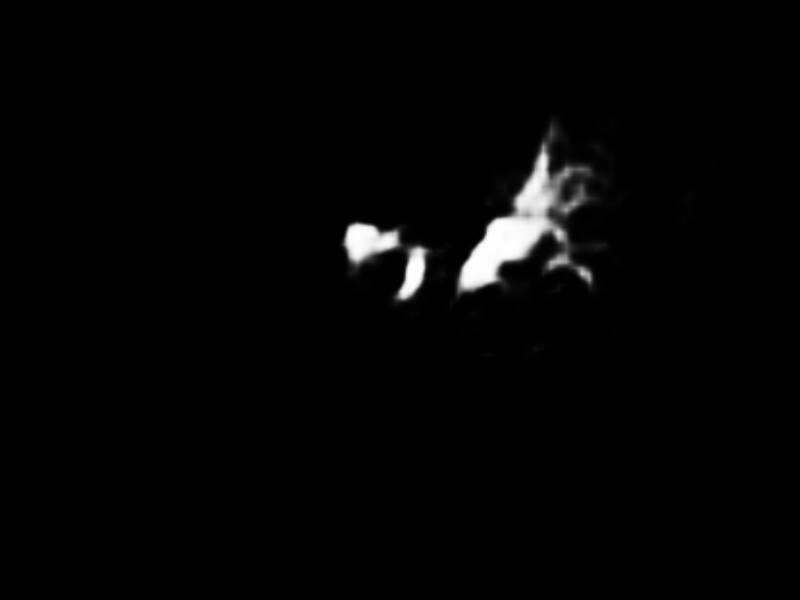}\\
			\vspace{0.01\linewidth}
                \includegraphics[width=1.15\textwidth,height=0.840\textwidth]{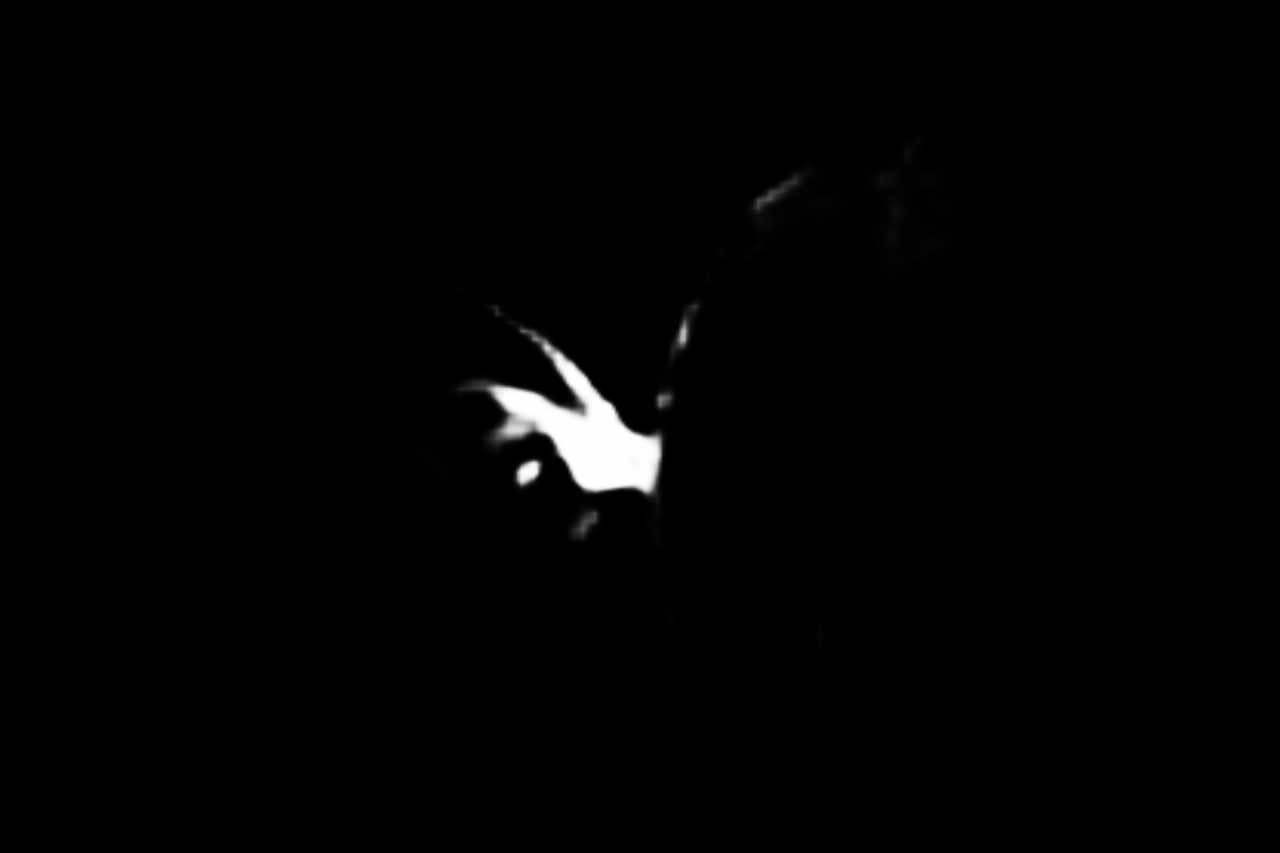}\\
                \vspace{0.01\linewidth}
			\includegraphics[width=1.15\linewidth,height=0.840\textwidth]{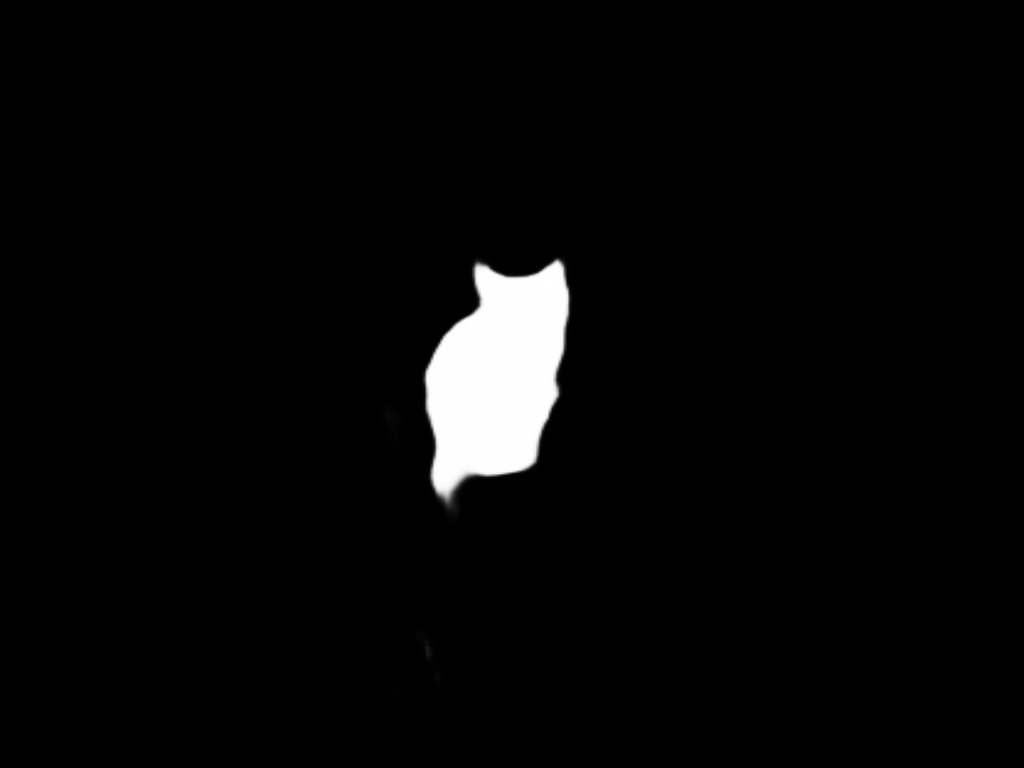}\\
			\vspace{0.08\linewidth}
		\end{minipage}%
	}\hspace{0.018\columnwidth}
	\subfigure[{\scriptsize SINet~\cite{fan2020camouflaged}}]{
		\begin{minipage}[t]{0.1\textwidth}
			\centering
			\includegraphics[width=1.15\linewidth,height=0.84\textwidth]{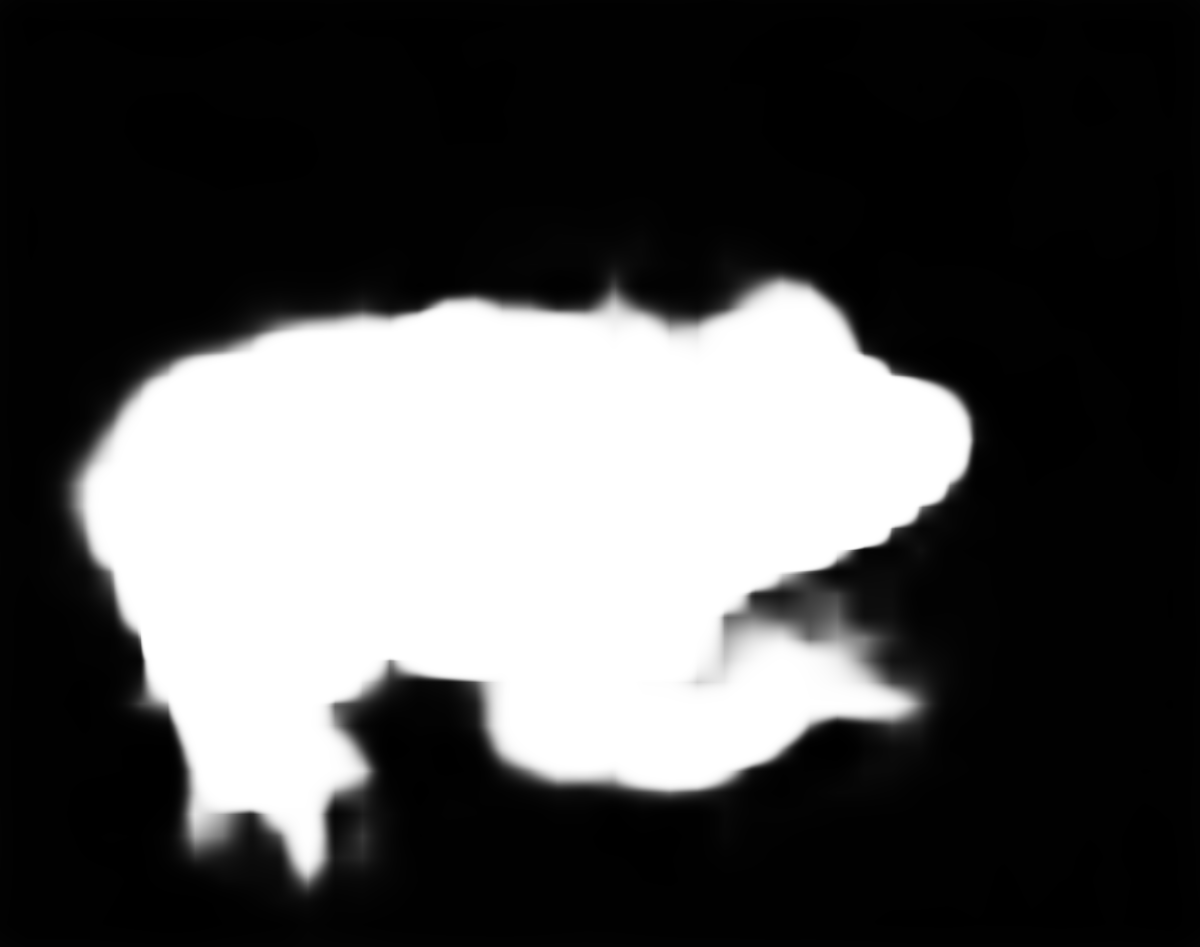}\\
			\vspace{0.01\linewidth}
                \includegraphics[width=1.15\textwidth,height=0.840\textwidth]{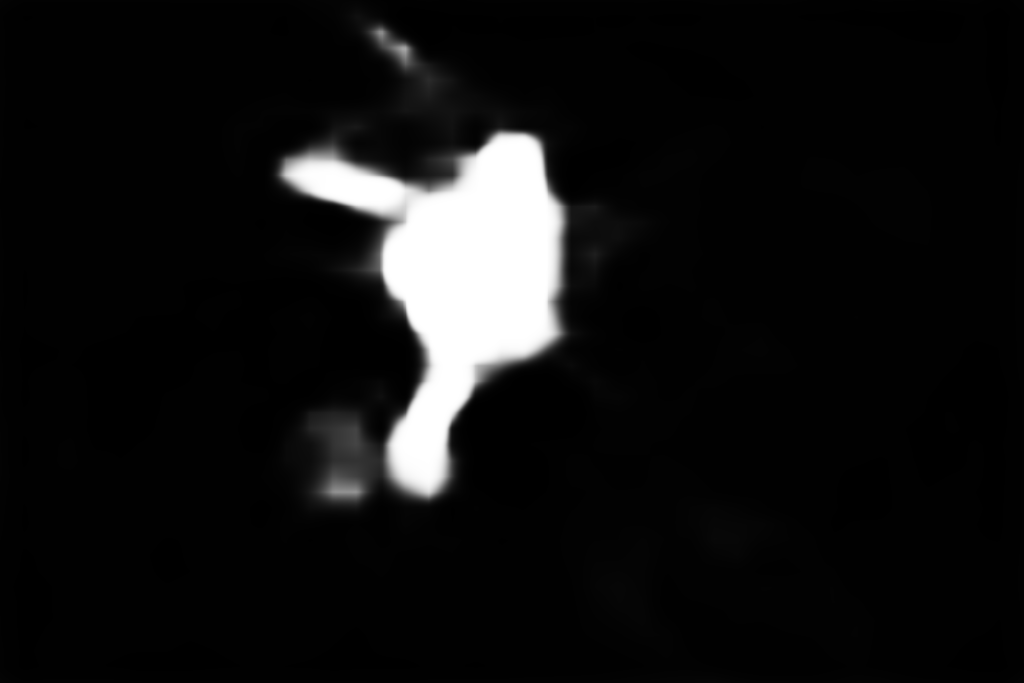}\\
			\vspace{0.01\linewidth}
                \includegraphics[width=1.15\linewidth,height=0.84\textwidth]{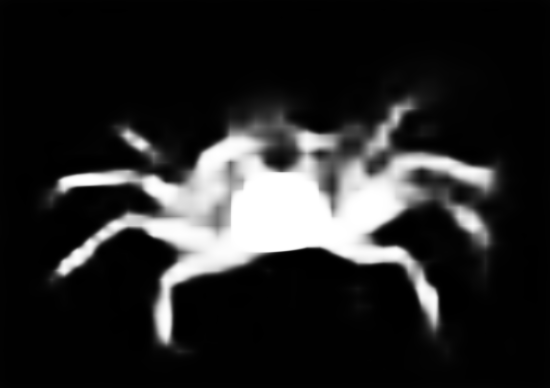}\\
                \vspace{0.01\linewidth}
			\includegraphics[width=1.15\linewidth,height=0.84\textwidth]{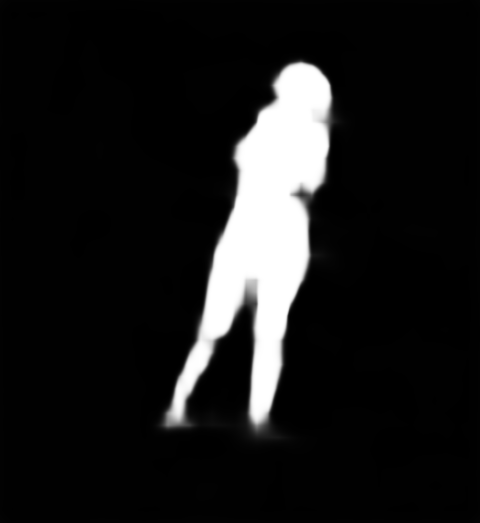}\\
			\vspace{0.01\linewidth}
                \includegraphics[width=1.15\linewidth,height=0.840\textwidth]{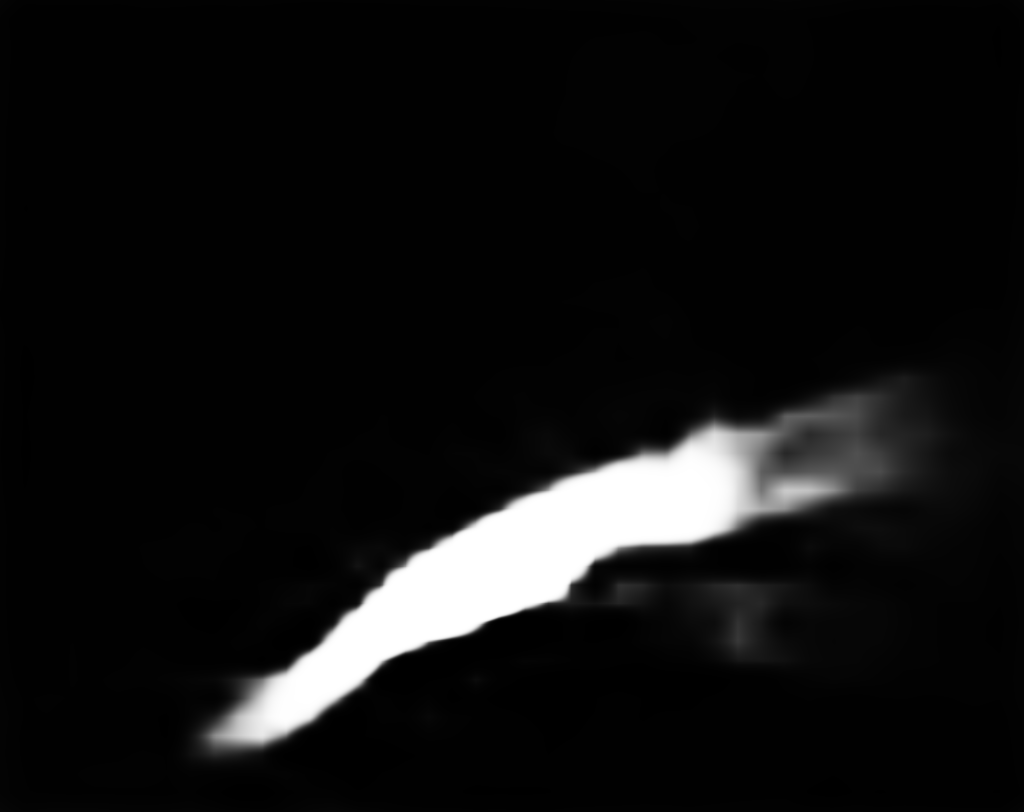}\\
			\vspace{0.01\linewidth}
                \includegraphics[width=1.15\linewidth,height=0.840\textwidth]{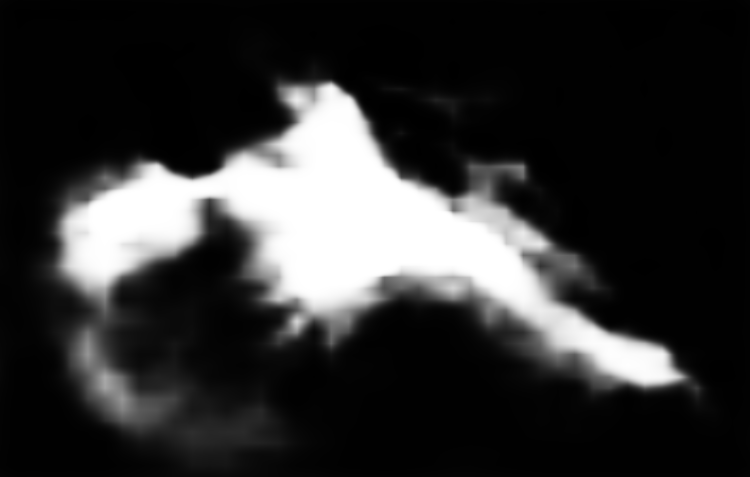}\\
			\vspace{0.01\linewidth}
                \includegraphics[width=1.15\linewidth,height=0.840\textwidth]{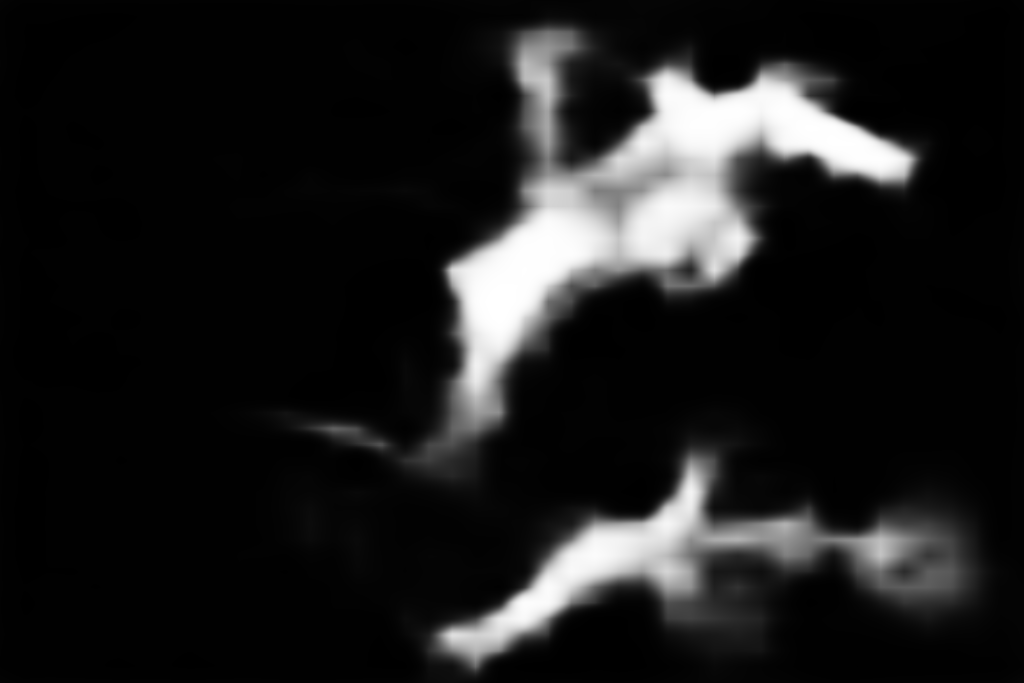}\\
			\vspace{0.01\linewidth}
			\includegraphics[width=1.15\linewidth,height=0.840\textwidth]{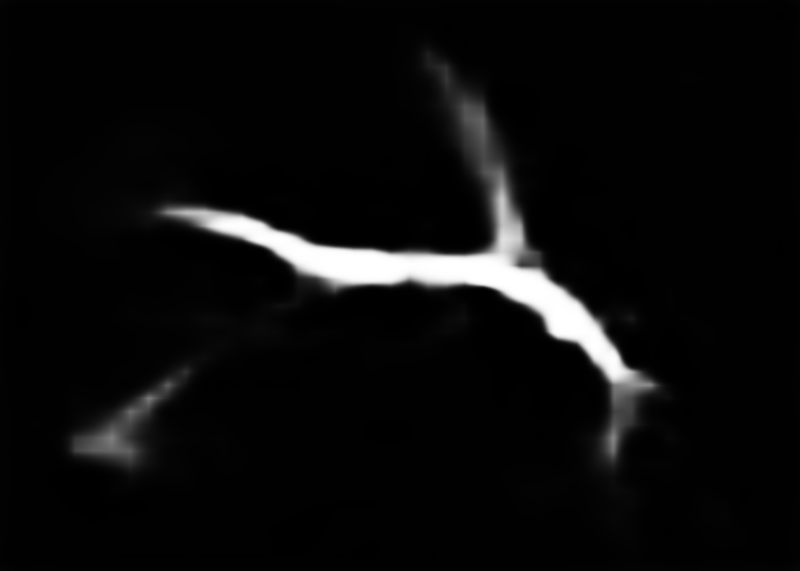}\\
			\vspace{0.01\linewidth}
			\includegraphics[width=1.15\linewidth,height=0.840\textwidth]{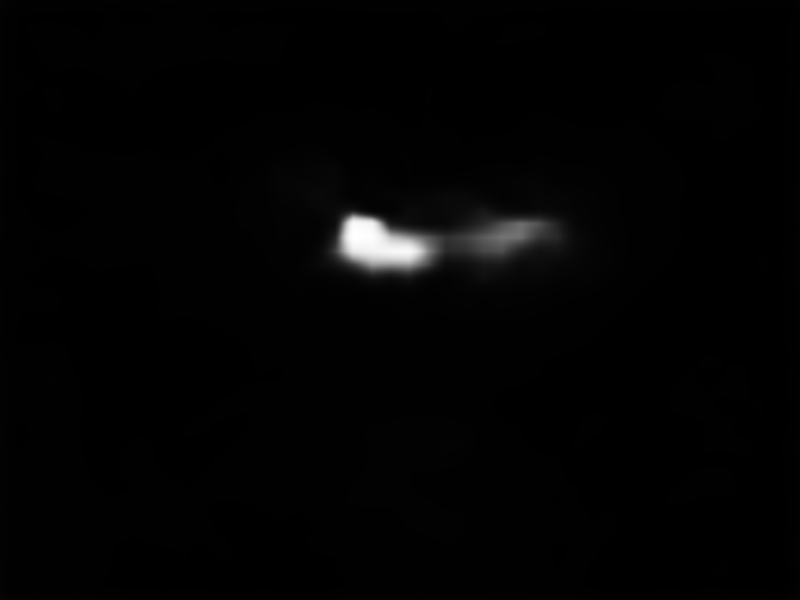}\\
			\vspace{0.01\linewidth}
                \includegraphics[width=1.15\textwidth,height=0.840\textwidth]{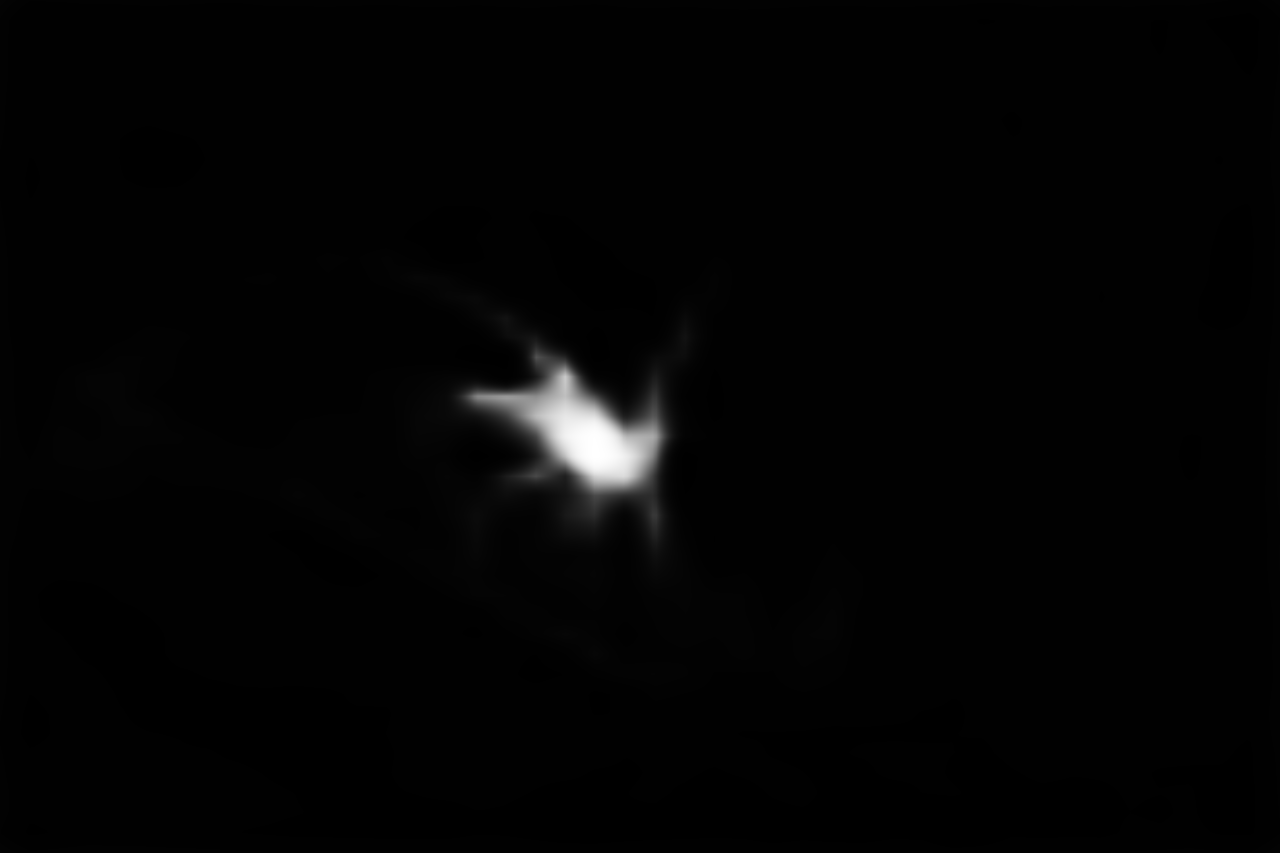}\\
                \vspace{0.01\linewidth}
			\includegraphics[width=1.15\linewidth,height=0.840\textwidth]{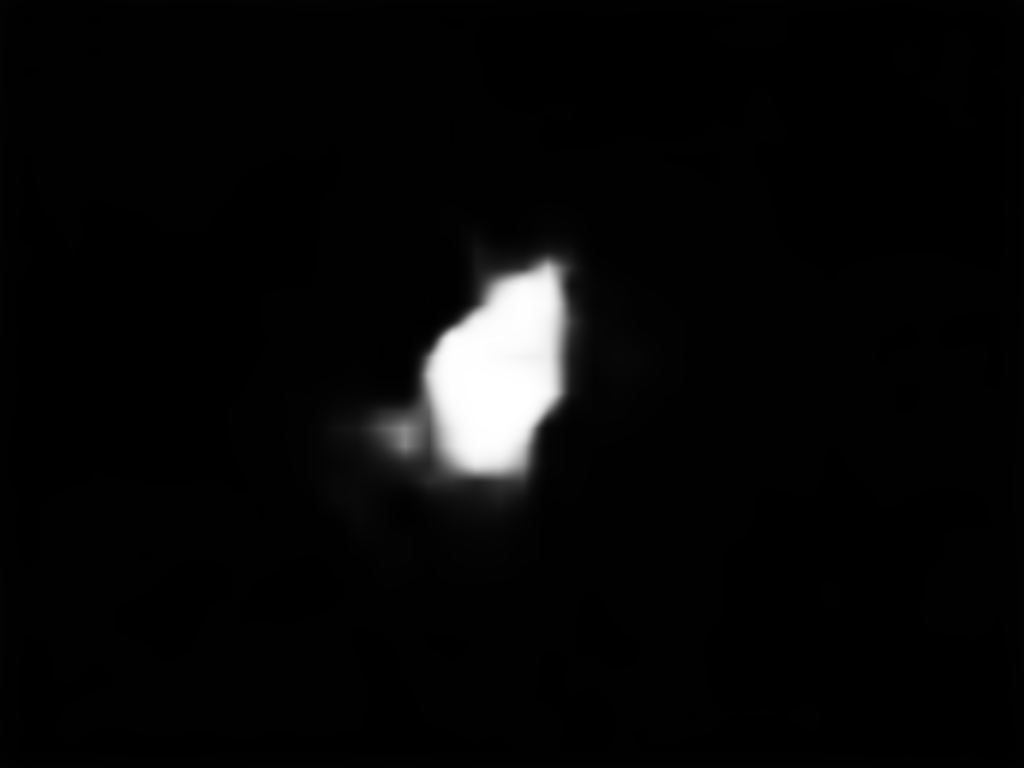}\\
			\vspace{0.08\linewidth}
		\end{minipage}%
	}\hspace{0.018\columnwidth}
        \subfigure[{\scriptsize MINet~\cite{pang2020multi}}]{
		\begin{minipage}[t]{0.1\textwidth}
			\centering
			\includegraphics[width=1.15\linewidth,height=0.84\textwidth]{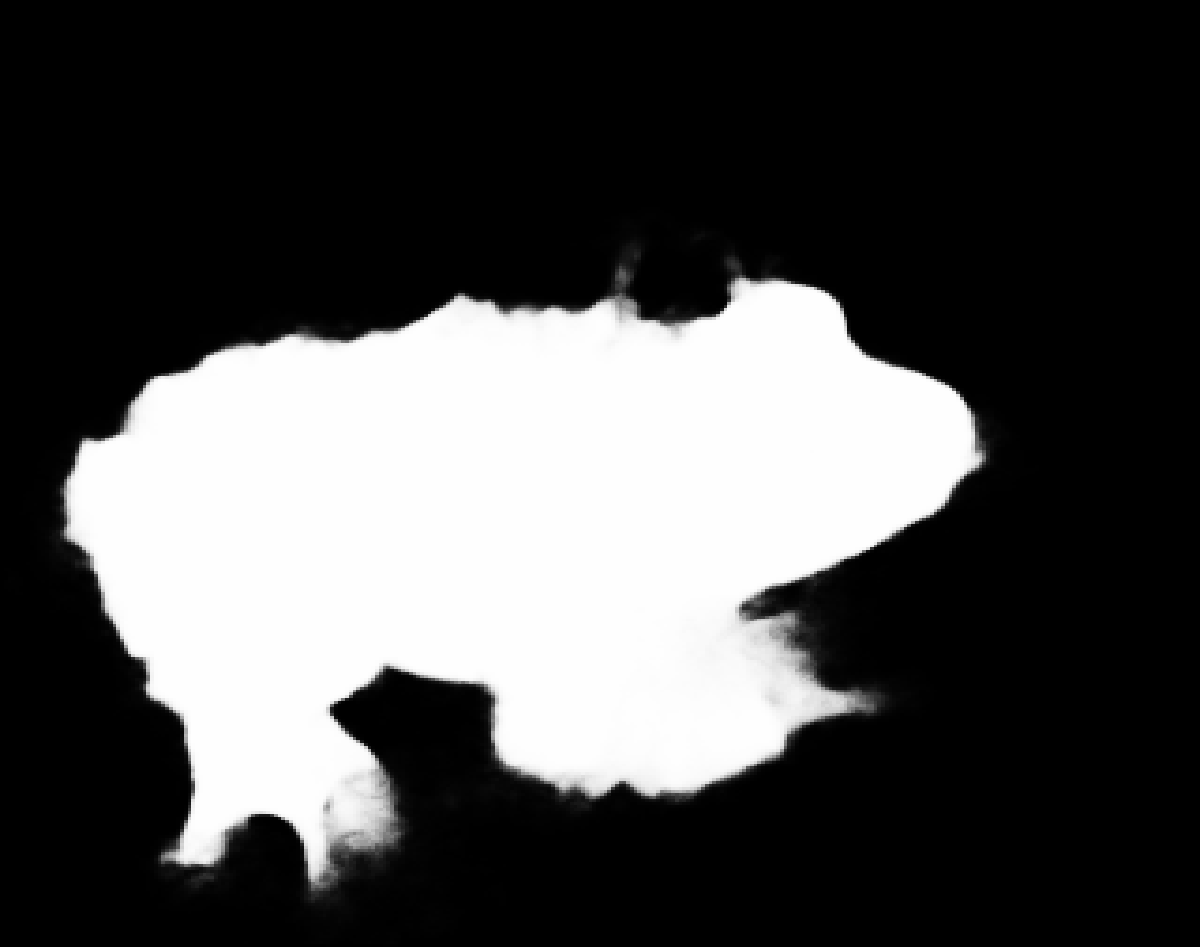}\\
			\vspace{0.01\linewidth}
                \includegraphics[width=1.15\textwidth,height=0.840\textwidth]{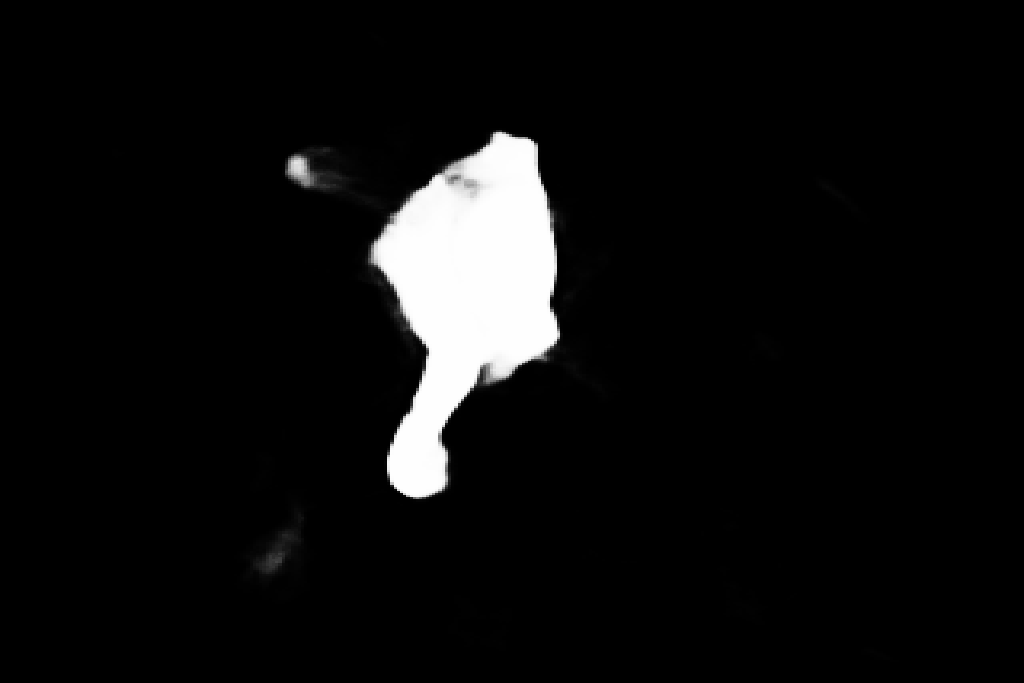}\\
			\vspace{0.01\linewidth}
                \includegraphics[width=1.15\linewidth,height=0.84\textwidth]{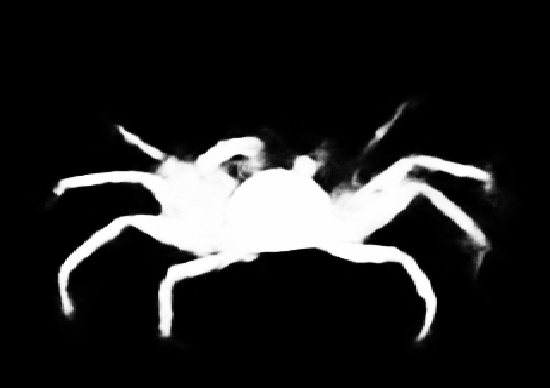}\\
                \vspace{0.01\linewidth}
			\includegraphics[width=1.15\linewidth,height=0.84\textwidth]{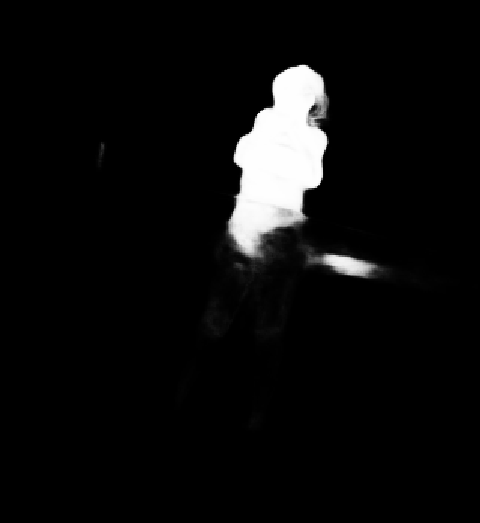}\\
			\vspace{0.01\linewidth}
                \includegraphics[width=1.15\linewidth,height=0.840\textwidth]{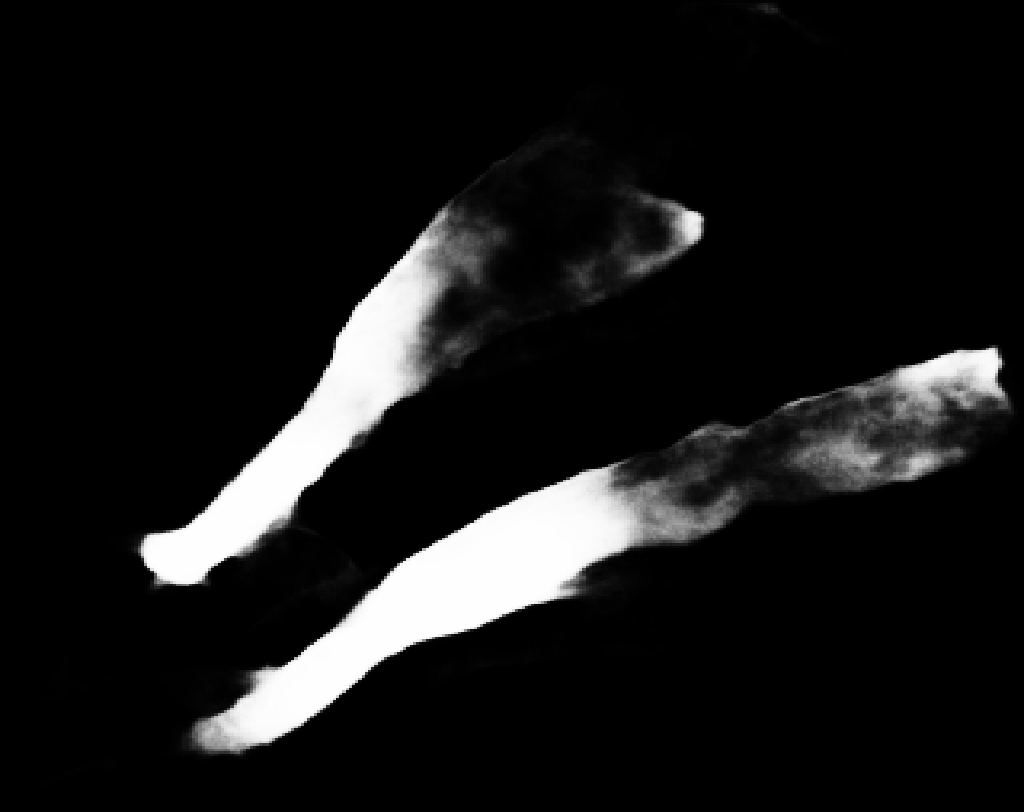}\\
			\vspace{0.01\linewidth}
                \includegraphics[width=1.15\linewidth,height=0.840\textwidth]{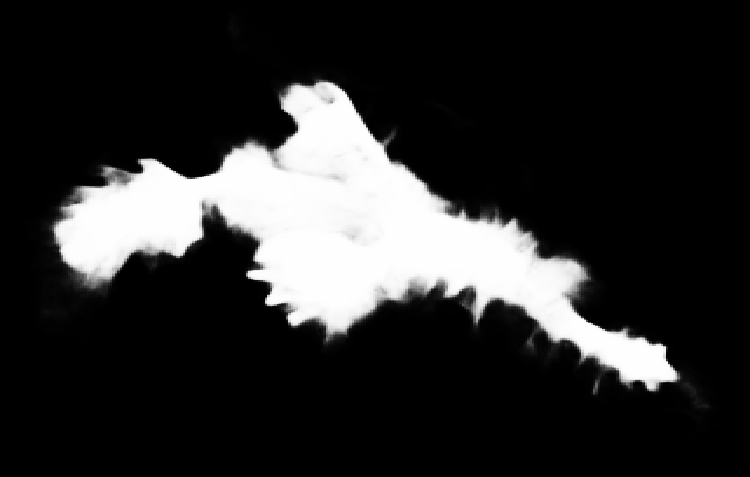}\\
			\vspace{0.01\linewidth}
                \includegraphics[width=1.15\linewidth,height=0.840\textwidth]{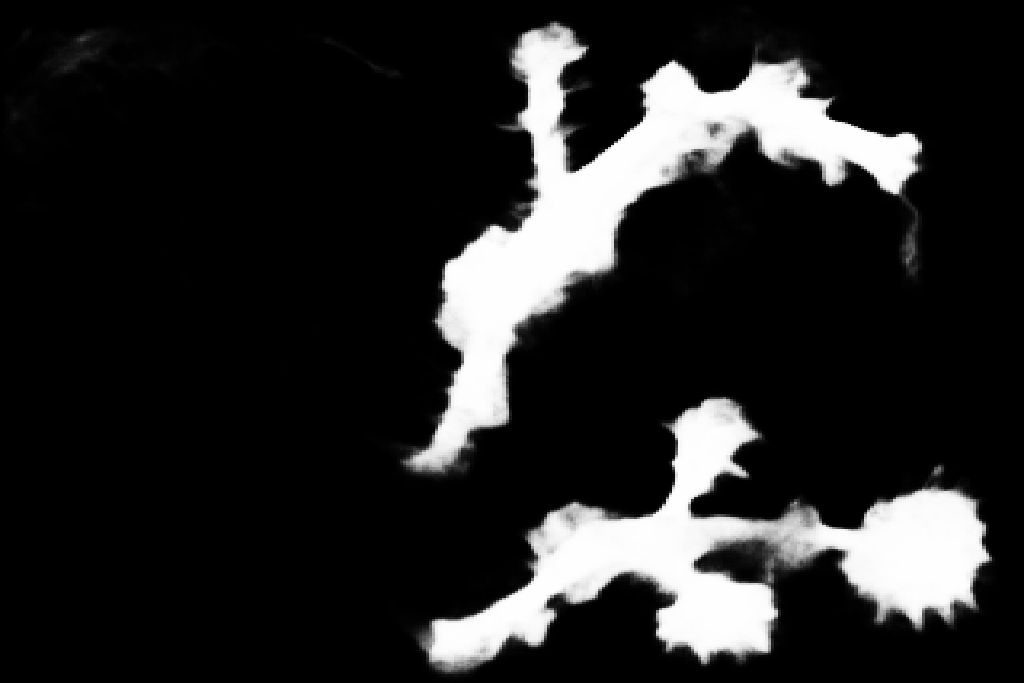}\\
			\vspace{0.01\linewidth}
			\includegraphics[width=1.15\linewidth,height=0.840\textwidth]{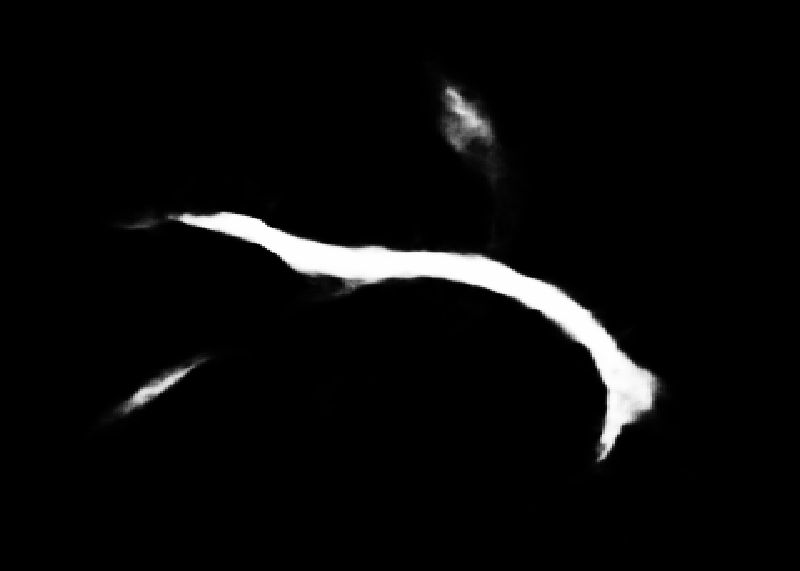}\\
			\vspace{0.01\linewidth}
			\includegraphics[width=1.15\linewidth,height=0.840\textwidth]{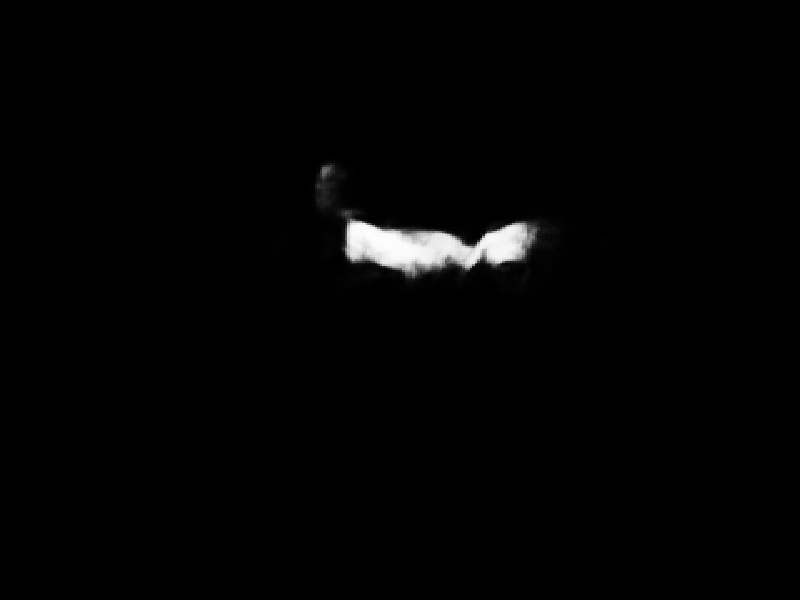}\\
			\vspace{0.01\linewidth}
                \includegraphics[width=1.15\textwidth,height=0.840\textwidth]{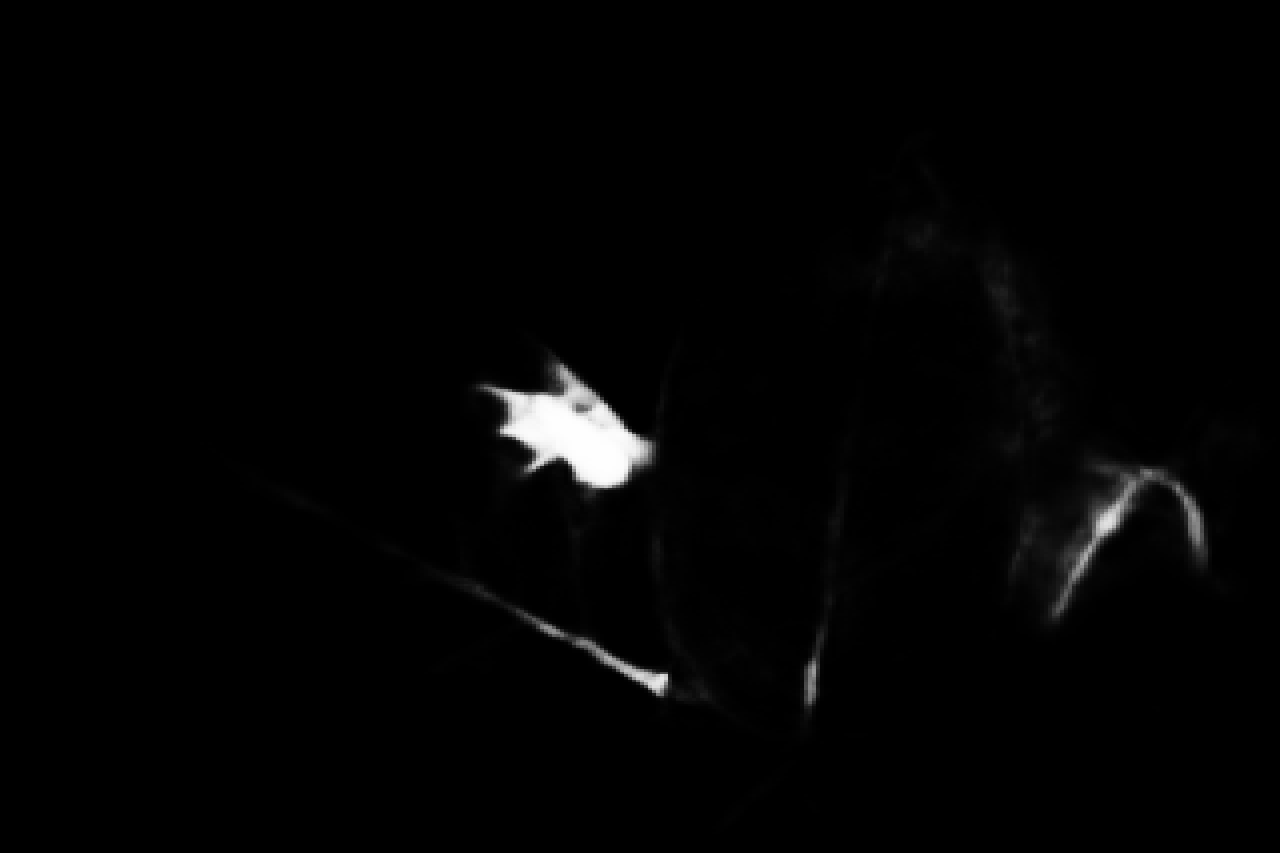}\\
                \vspace{0.01\linewidth}
			\includegraphics[width=1.15\linewidth,height=0.840\textwidth]{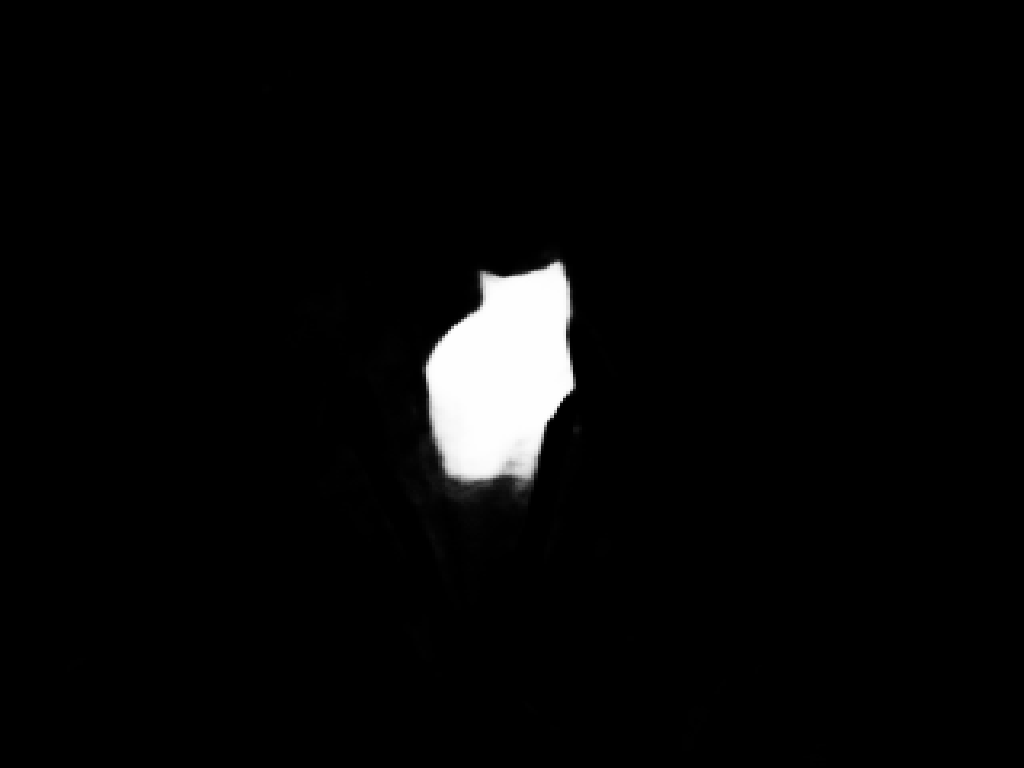}\\
			\vspace{0.08\linewidth}
		\end{minipage}%
	}\hspace{0.018\columnwidth}
	\centering
	\caption{Qualitative comparison of our proposed method and other representative COD methods. Our method provides better performance than all competitors for camouflaged object segmentation in various complex scenes.}
    \label{fig3}
    \vspace{5mm}
\end{figure*}

\subsection{Qualitative Evaluation}
Figure~\ref{fig3} shows a comprehensive visual comparison with current state-of-the-art methods. It can be found that our method achieves competitive visual performance in different types of challenging scenarios. Our diffCOD is able to guarantee the integrity and correctness of recognition even under difficult conditions, such as single object (\textit{e.g.,} row 1-4), multi-objects (\textit{e.g.,} row 5-8), small object (\textit{e.g.}, row 9-11). Nature's camouflaged organisms often have strange traits, such as tentacles, tiny spikes, etc. Past models have blurred the recognition of edge parts even if the location of the target is correctly targeted. However, we are surprised by the advantages of diffCOD in terms of detailed textures. As shown in Figure \ref{fig:detail}, our method is able to accurately identify every subtlety, and it can depict the textures of the object in extremely fine detail, solving the blurring problem of segmentation masks in other methods.

\subsection{Ablation Studies}
\noindent\textbf{Overview.}
We perform ablation studies on key components to verify their effectiveness and analyze their impacts on performance, as shown in Table \ref{tab2}. Experimental results demonstrate that our designed Injection Attention Module (IAM), Feature Fusion (FF), and ViT can improve detection performance. When they are combined to build diffCOD, significant improvements in all evaluation metrics are observed. Note that the Baseline refers to the standard diffusion model.

\begin{table*}[t]
\resizebox{\linewidth}{!}{
\renewcommand{\arraystretch}{1.3}
\begin{tabular}{c|cccc|ccccc|ccccc|ccccc}
\toprule[1pt]
\multirow{2}{*}{No.} & \multicolumn{4}{c|}{\textbf{Component}}                                                                              & \multicolumn{5}{c|}{\textbf{COD10K}}                                                                     & \multicolumn{5}{c|}{\textbf{NC4K}}                                                                                                            & \multicolumn{5}{c}{\textbf{CAMO}}                                                                                              \\ \cline{2-20} 
                     & Baseline      & IAM & FF & ViT  & $S_{\alpha}\uparrow$              & $F_{\beta}^{\omega}\uparrow$              & $F_{m}\uparrow$              & $E_{m}\uparrow$              & $MAE\downarrow$                                   & \multicolumn{1}{c}{$S_{\alpha}\uparrow$} & \multicolumn{1}{c}{$F_{\beta}^{\omega}\uparrow$} & \multicolumn{1}{c}{$F_{m}\uparrow$} & \multicolumn{1}{c}{$E_{m}\uparrow$} & \multicolumn{1}{c|}{$MAE\downarrow$}               & \multicolumn{1}{c}{$S_{\alpha}\uparrow$} & \multicolumn{1}{c}{$F_{\beta}^{\omega}\uparrow$} & \multicolumn{1}{c}{$F_{m}\uparrow$} & \multicolumn{1}{c}{$E_{m}\uparrow$} & \multicolumn{1}{c}{$MAE\downarrow$} \\ \hline
\#1                  & $\checkmark$ &                         &                         &                                                  & 0.761          & 0.604          & 0.657          & 0.845          & \multicolumn{1}{c|}{0.046}          & 0.781                 & 0.687                 & 0.712                 & 0.841                 & \multicolumn{1}{c|}{0.061}          & 0.731                 & 0.607                 & 0.664                 & 0.790                 & 0.097                 \\
\#2                  & $\checkmark$ & $\checkmark$            &                         &                                                  & 0.788          & 0.638          & 0.687          & 0.861          & \multicolumn{1}{c|}{0.041}          & 0.805                 & 0.711                 & 0.747                 & 0.863                 & \multicolumn{1}{c|}{0.056}          & 0.749                 & 0.631                 & 0.694                 & 0.805                 & 0.093                 \\  
\#3                  & $\checkmark$ & $\checkmark$            & $\checkmark$            &                                                  & 0.801          & 0.662          & 0.709          & 0.876          & \multicolumn{1}{c|}{0.039}          & 0.823                 & 0.731                 & 0.772                 & 0.876                 & \multicolumn{1}{c|}{0.054}          & 0.770                 & 0.664                 & 0.718                 & 0.829                 & 0.087                 \\
\#4                  & $\checkmark$ & $\checkmark$            &             & $\checkmark$                                                 & 0.809          & 0.677          & 0.719          & 0.888          & \multicolumn{1}{c|}{0.036}          & 0.835                 & 0.758                 & 0.798                 & 0.889                 & \multicolumn{1}{c|}{0.051}          & 0.792                 & 0.693                 & 0.751                 & 0.849                 & 0.083                 \\
\#5                  & $\checkmark$ &            & $\checkmark$            & $\checkmark$                                     & 0.799          & 0.657          & 0.708          & 0.868          & \multicolumn{1}{c|}{0.039}          & 0.820                 & 0.727                 & 0.770                 & 0.872                 & \multicolumn{1}{c|}{0.054}          & 0.772                 & 0.663                 & 0.722                 & 0.831                 & 0.086                 \\ \cline{1-20}
\#OUR                & $\checkmark$ & $\checkmark$            & $\checkmark$            & $\checkmark$                        & \textbf{0.812} & \textbf{0.684} & \textbf{0.723} & \textbf{0.892} & \multicolumn{1}{c|}{\textbf{0.036}} & \textbf{0.837} & \textbf{0.761} & \textbf{0.802} & \textbf{0.891} & \multicolumn{1}{c|}{\textbf{0.051}} & \textbf{0.795} & \textbf{0.704} & \textbf{0.758} & \textbf{0.852} & \textbf{0.082}       \\ \bottomrule[1pt]
\end{tabular}
}
\caption{Ablation studies of our diffCOD. The best results are marked in $\mathbf{bold}$.}
\label{tab2}
\end{table*}

\begin{figure*}[t]
	\centering
	\subfigure[{\scriptsize Image}]{
		\begin{minipage}[t]{0.118\textwidth}
			\centering
                \includegraphics[width=1.15\textwidth,height=0.840\textwidth]{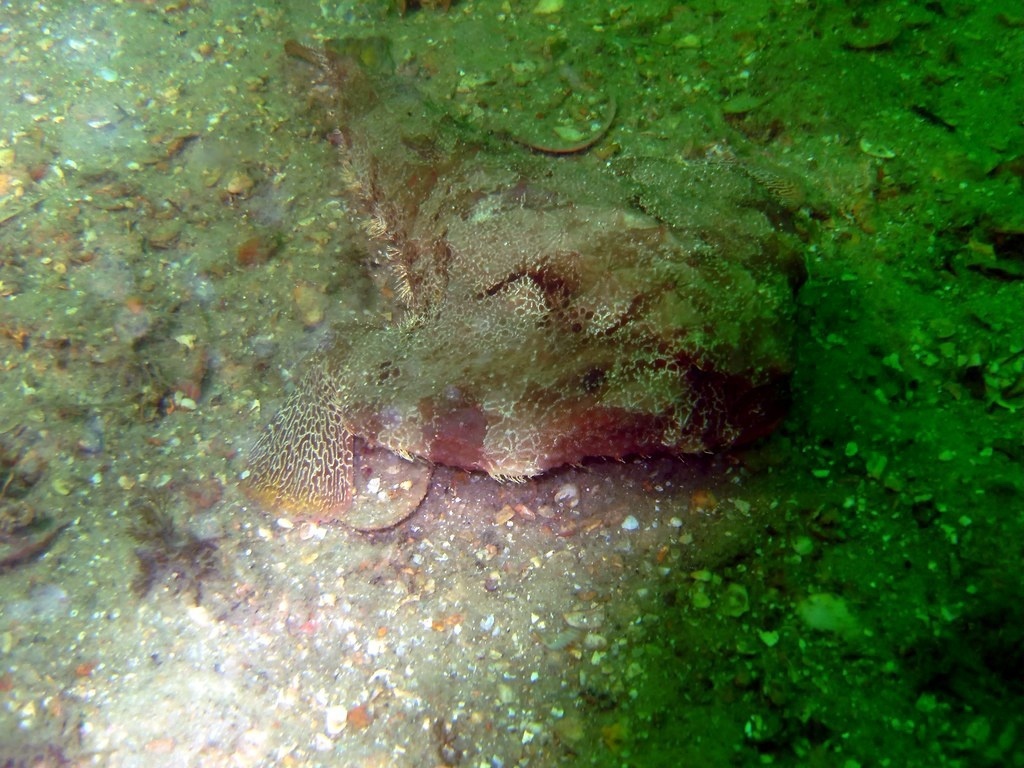}\\
			\vspace{0.01\linewidth}
                \includegraphics[width=1.15\textwidth,height=0.840\textwidth]{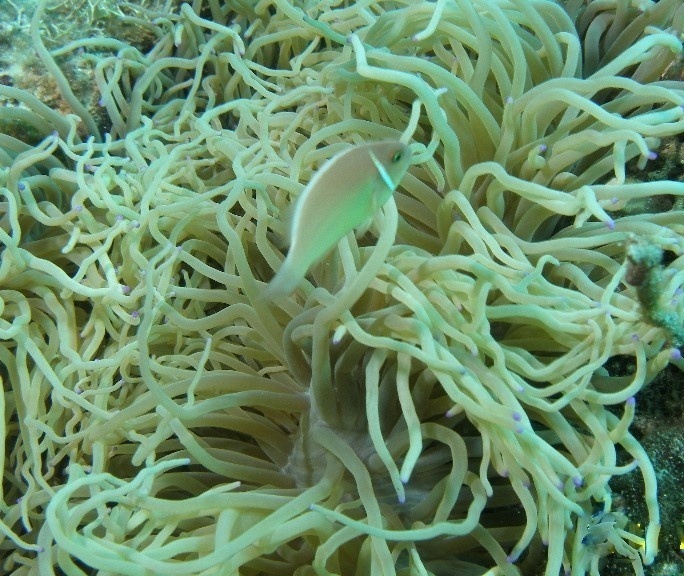}\\
                \vspace{0.01\linewidth}
			\includegraphics[width=1.15\linewidth,height=0.840\textwidth]{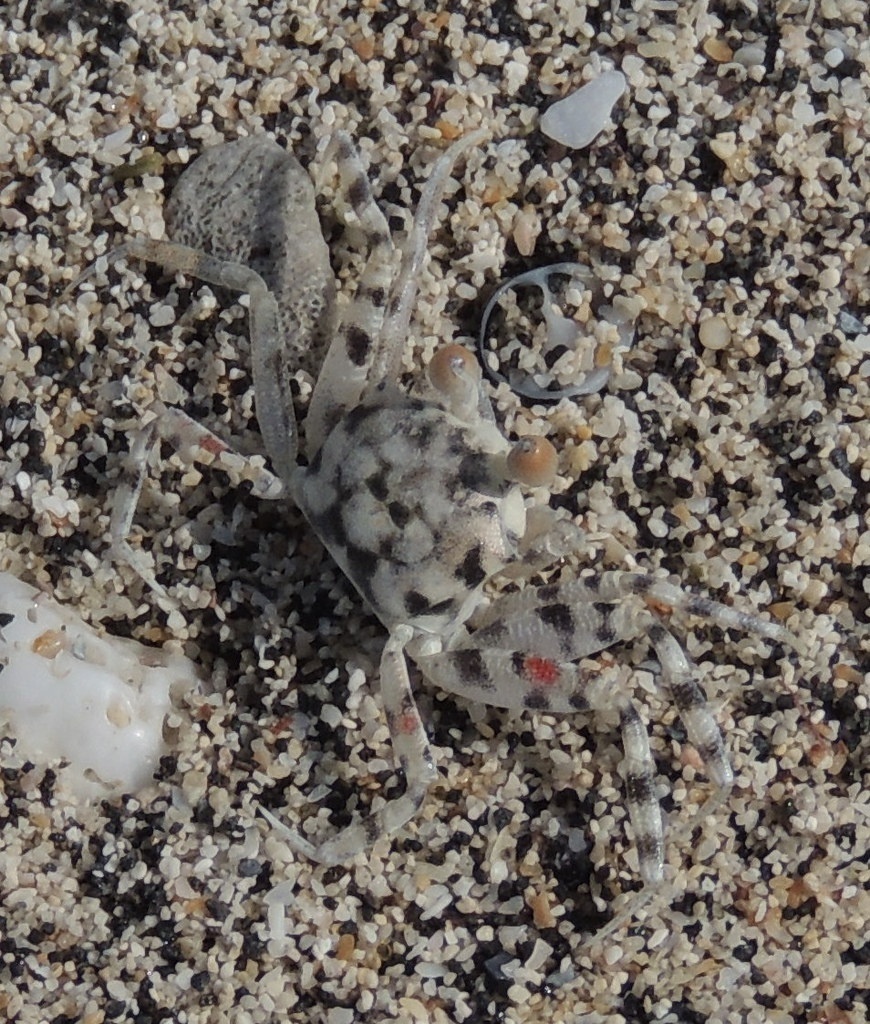}\\
			\vspace{0.01\linewidth}
			\includegraphics[width=1.15\linewidth,height=0.840\textwidth]{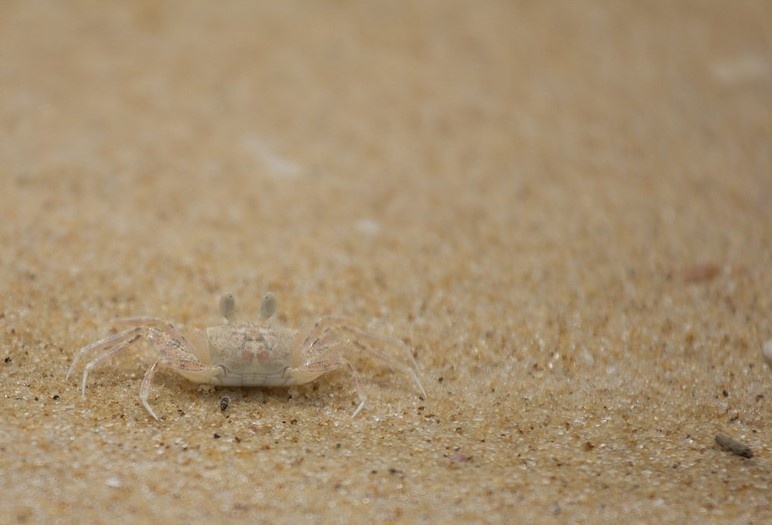}\\
			\vspace{0.08\linewidth}
		\end{minipage}%
	}\hspace{0.03\columnwidth}
	\subfigure[{\scriptsize GT}]{
		\begin{minipage}[t]{0.118\textwidth}
			\centering
                \includegraphics[width=1.15\textwidth,height=0.840\textwidth]{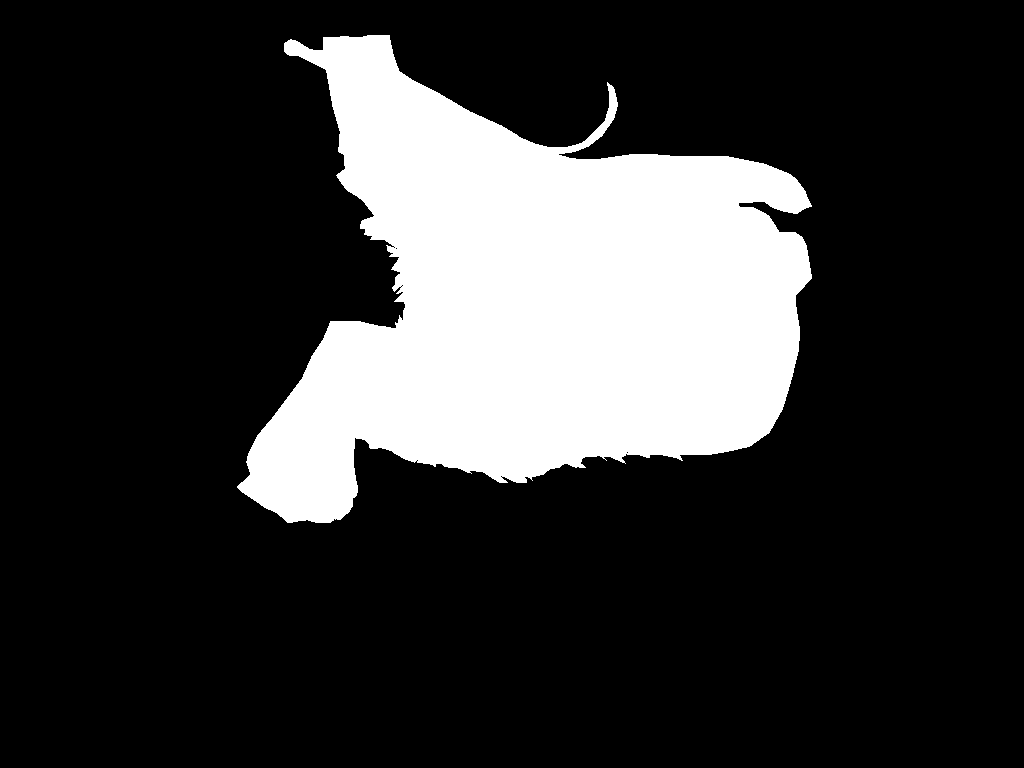}\\
			\vspace{0.01\linewidth}
                \includegraphics[width=1.15\textwidth,height=0.840\textwidth]{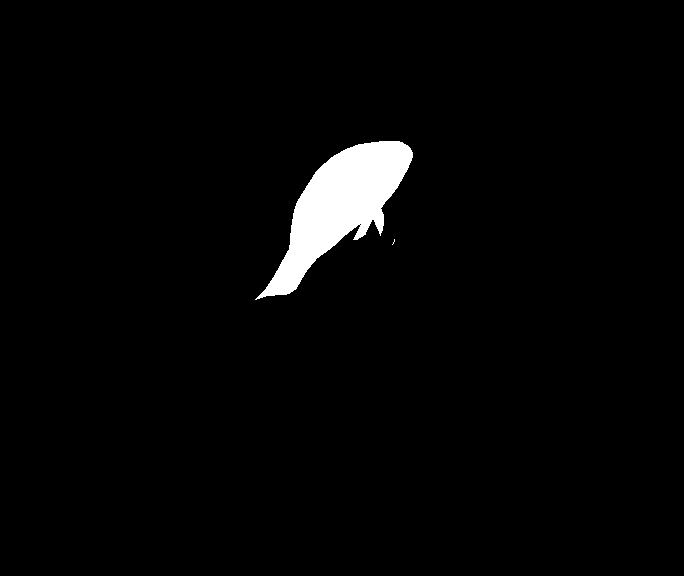}\\
                \vspace{0.01\linewidth}
			\includegraphics[width=1.15\linewidth,height=0.840\textwidth]{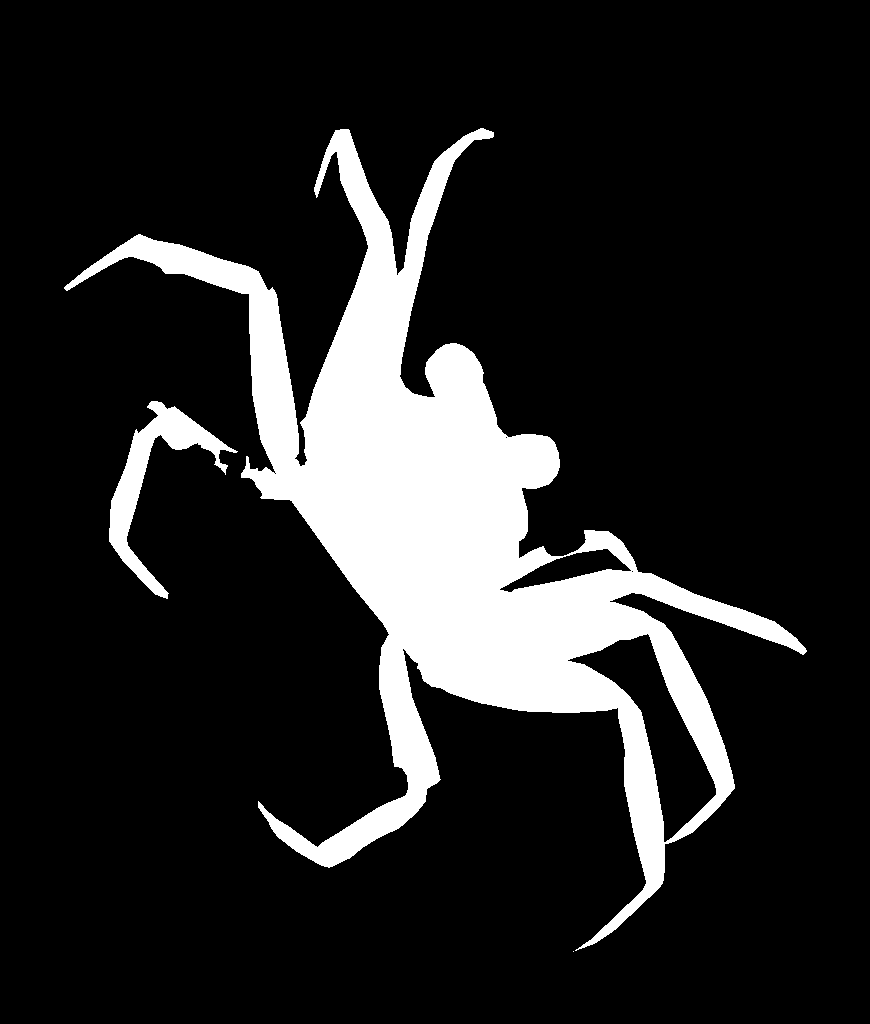}\\
			\vspace{0.01\linewidth}
			\includegraphics[width=1.15\linewidth,height=0.840\textwidth]{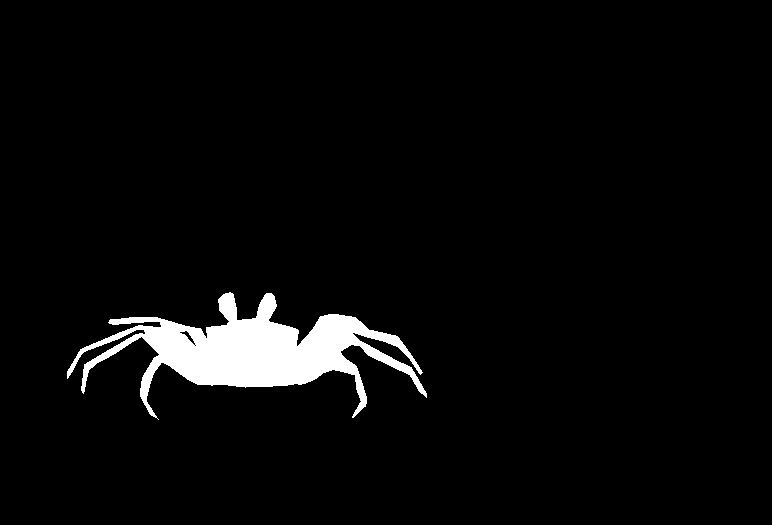}\\
			\vspace{0.08\linewidth}
		\end{minipage}%
	}\hspace{0.03\columnwidth}
	\subfigure[{\scriptsize Sample 1}]{
		\begin{minipage}[t]{0.118\textwidth}
			\centering
                \includegraphics[width=1.15\textwidth,height=0.840\textwidth]{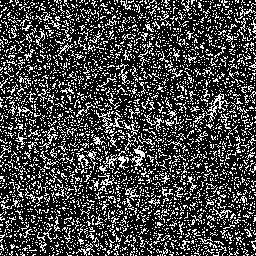}\\
			\vspace{0.01\linewidth}
                \includegraphics[width=1.15\textwidth,height=0.840\textwidth]{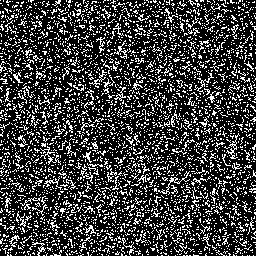}\\
                \vspace{0.01\linewidth}
			\includegraphics[width=1.15\linewidth,height=0.840\textwidth]{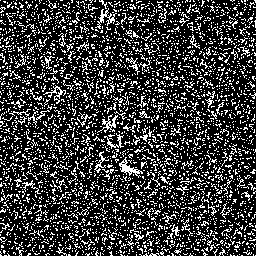}\\
			\vspace{0.01\linewidth}
			\includegraphics[width=1.15\linewidth,height=0.840\textwidth]{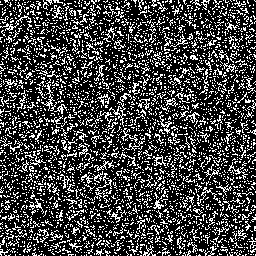}\\
			\vspace{0.08\linewidth}
		\end{minipage}%
	}\hspace{0.03\columnwidth}
	\subfigure[{\scriptsize Sample 2}]{
		\begin{minipage}[t]{0.118\textwidth}
			\centering
                \includegraphics[width=1.15\textwidth,height=0.840\textwidth]{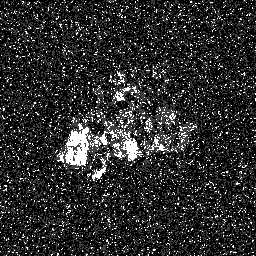}\\
			\vspace{0.01\linewidth}
                \includegraphics[width=1.15\textwidth,height=0.840\textwidth]{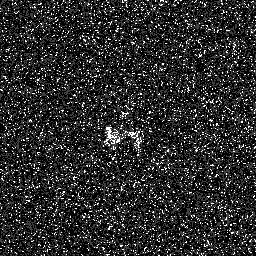}\\
                \vspace{0.01\linewidth}
			\includegraphics[width=1.15\linewidth,height=0.840\textwidth]{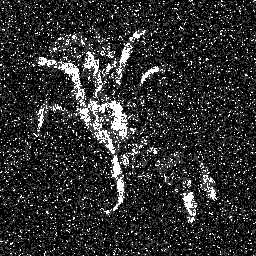}\\
			\vspace{0.01\linewidth}
			\includegraphics[width=1.15\linewidth,height=0.840\textwidth]{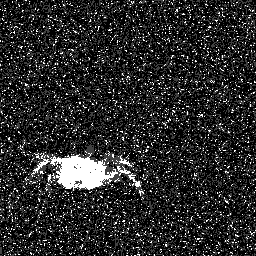}\\
			\vspace{0.08\linewidth}
		\end{minipage}%
	}\hspace{0.03\columnwidth}
	\subfigure[{\scriptsize Sample 3}]{
		\begin{minipage}[t]{0.118\textwidth}
			\centering
                \includegraphics[width=1.15\textwidth,height=0.840\textwidth]{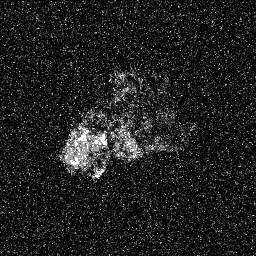}\\
			\vspace{0.01\linewidth}
                \includegraphics[width=1.15\textwidth,height=0.840\textwidth]{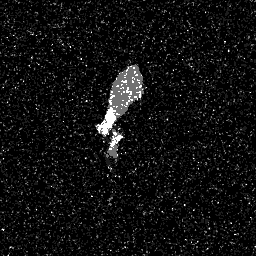}\\
                \vspace{0.01\linewidth}
			\includegraphics[width=1.15\linewidth,height=0.840\textwidth]{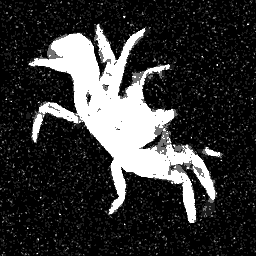}\\
			\vspace{0.01\linewidth}
			\includegraphics[width=1.15\linewidth,height=0.840\textwidth]{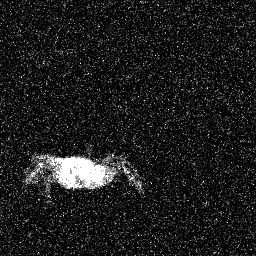}\\
			\vspace{0.08\linewidth}
		\end{minipage}%
	}\hspace{0.03\columnwidth}
	\subfigure[{\scriptsize Sample 4}]{
		\begin{minipage}[t]{0.118\textwidth}
			\centering
                \includegraphics[width=1.15\textwidth,height=0.840\textwidth]{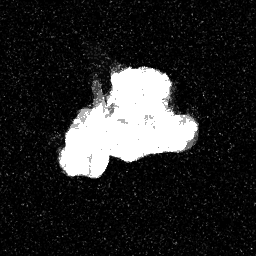}\\
			\vspace{0.01\linewidth}
                \includegraphics[width=1.15\textwidth,height=0.840\textwidth]{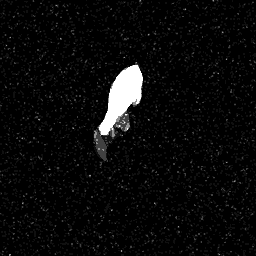}\\
                \vspace{0.01\linewidth}
			\includegraphics[width=1.15\linewidth,height=0.840\textwidth]{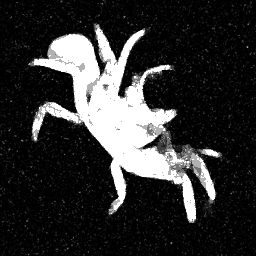}\\
			\vspace{0.01\linewidth}
			\includegraphics[width=1.15\linewidth,height=0.840\textwidth]{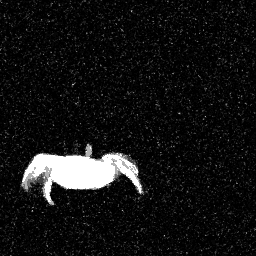}\\
			\vspace{0.08\linewidth}
		\end{minipage}%
	}\hspace{0.03\columnwidth}
	\subfigure[{\scriptsize Sample 5}]{
		\begin{minipage}[t]{0.118\textwidth}
			\centering
                \includegraphics[width=1.15\textwidth,height=0.840\textwidth]{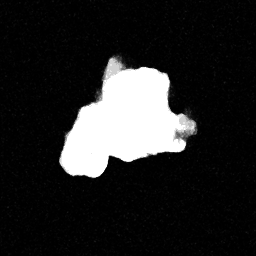}\\
			\vspace{0.01\linewidth}
                \includegraphics[width=1.15\textwidth,height=0.840\textwidth]{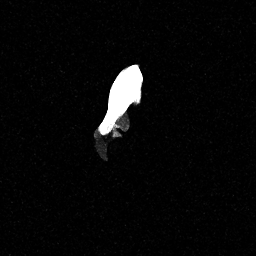}\\
                \vspace{0.01\linewidth}
			\includegraphics[width=1.15\linewidth,height=0.840\textwidth]{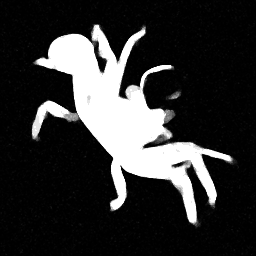}\\
			\vspace{0.01\linewidth}
			\includegraphics[width=1.15\linewidth,height=0.840\textwidth]{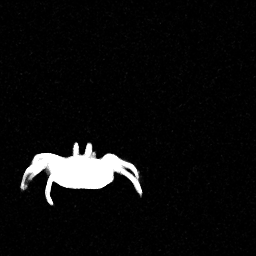}\\
			\vspace{0.08\linewidth}
		\end{minipage}%
	}\hspace{0.03\columnwidth}
	\centering
        \caption{Visual results of the sampling process. (c)-(g) is the diffCOD sampling process. The time step is 200, 400, 600, 800, and 1000, respectively.}
        \label{fig4}
\end{figure*}

\vspace{4mm}
\noindent\textbf{Effectiveness of IAM.}
As can be seen in Table \ref{tab2}, the presence or absence of IAM plays a key role in the performance improvement of the model. Compared to the experiments without this key component, the average improvement of \#2 with IAM over \#1 for $S_{\alpha}$, $F_{\beta}^{\omega}$, $F_{m}$, $E_{m}$ and $MAE$ on the three datasets is 3.0\%, 4.3\%, 4.7\%, 2.1\% and 7.7\%, respectively. Furthermore, \#Our accuracy improvement over \#5 is significant, with an average increase of 6.0\% in $MAE$ metric on the three datasets. This is a good indication that IAM integrates diffusion features and texture features from the backbone perfectly.

\vspace{5mm}
\noindent\textbf{Effectiveness of FF.} 
The main role of FF is to aggregate the multi-scale features. As shown in Table~\ref{tab2}, compared to No.~\#2, No.~\#3 has an average improvement of 2.2\%, 5.0\%, 3.8\%, 2.5\% and 6.0\% for $S_{\alpha}$, $F_{\beta}^{\omega}$, $F_{m}$, $E_{m}$ and $MAE$ on the three datasets, respectively. The performance of \#Ours on $S_{\alpha}$, $F_{\beta}^{\omega}$, $F_{m}$ and $E_{m}$ is 3.2\%, 1.0\%, 0.7\% and 0.3\% higher than that of No.~\#4.

\vspace{5mm}
\noindent\textbf{Effectiveness of ViT.}
To obtain the location information and texture information of the objects in the original features, we use a ViT as a backbone to assist the diffusion process. From Table~\ref{tab2}, we can learn that \#Ours containing rich original features has an average improvement of 2.1\%, 4.5\%, 3.8\%, 2.1\% and 6.3\% over \#3 for $S_{\alpha}$, $F_{\beta}^{\omega}$, $F_{m}$, $E_{m}$ and $MAE$ on the three datasets, respectively. \#2, which contains no original features at all, has an average of 4.0\%, 7.5\%, 6.6\%, 3.9\% and 10.6\% lower than \#4 for $S_{\alpha}$, $F_{\beta}^{\omega}$, $F_{m}$, $E_{m}$ and $MAE$ on the three data sets, respectively. 
In addition, to further demonstrate the significance of conditional semantic features to guide the diffusion process, we visualize the sampling process of diffCOD. From Figure \ref{fig4}, we can see that our model learns part of the location information and texture patterns of the camouflaged objects at the early stage of denoising, and the subsequent inference process gradually refines the final mask by training out the denoising model on this basis. This shows that the key clues extracted by ViT are perfectly integrated into the diffusion process with the help of FF and IAM.

\vspace{2.2mm}
\section{Conclusion}
In this paper, we propose a diffusion-based framework for camouflaged object detection, which changes the previous detection paradigm of the COD community by using a generative model for the segmentation of camouflaged objects to achieve significant performance gains. To the best of our knowledge, this is the first framework that employs a denoising diffusion model for COD tasks. Our approach decouples the task of segmenting camouflaged objects into a series of forward and reverse diffusion processes, and integrates key information from conditional semantic features to guide this process. Extensive experiments show the superiority over 11 other state-of-the-art methods on four datasets. 
As a new paradigm for camouflaged object detection, we hope that our proposed method will serve as a solid baseline and encourage future research. 

\newpage
\bibliography{ecai}

\begin{thebibliography}{10}

\bibitem{amit2021segdiff}
Tomer Amit, Eliya Nachmani, Tal Shaharbany, and Lior Wolf, `Segdiff: Image
  segmentation with diffusion probabilistic models', {\em arXiv preprint
  arXiv:2112.00390}, (2021).

\bibitem{bakr2023hrs}
Eslam~Mohamed Bakr, Pengzhan Sun, Xiaoqian Shen, Faizan~Farooq Khan, Li~Erran
  Li, and Mohamed Elhoseiny, `Hrs-bench: Holistic, reliable and scalable
  benchmark for text-to-image models', {\em arXiv preprint arXiv:2304.05390},
  (2023).

\bibitem{baranchuk2021label}
Dmitry Baranchuk, Andrey Voynov, Ivan Rubachev, Valentin Khrulkov, and Artem
  Babenko, `Label-efficient semantic segmentation with diffusion models', in
  {\em {ICLR}}, (2022).

\bibitem{brempong2022denoising}
Emmanuel~Asiedu Brempong, Simon Kornblith, Ting Chen, Niki Parmar, Matthias
  Minderer, and Mohammad Norouzi, `Denoising pretraining for semantic
  segmentation', in {\em CVPR}, pp. 4175--4186, (2022).

\bibitem{cheng2022implicit}
Xuelian Cheng, Huan Xiong, Deng-Ping Fan, Yiran Zhong, Mehrtash Harandi, Tom
  Drummond, and Zongyuan Ge, `Implicit motion handling for video camouflaged
  object detection', in {\em CVPR}, pp. 13864--13873, (2022).

\bibitem{chung2022come}
Hyungjin Chung, Byeongsu Sim, and Jong~Chul Ye, `Come-closer-diffuse-faster:
  Accelerating conditional diffusion models for inverse problems through
  stochastic contraction', in {\em Proceedings of the IEEE/CVF Conference on
  Computer Vision and Pattern Recognition}, pp. 12413--12422, (2022).

\bibitem{couairon2022diffedit}
Guillaume Couairon, Jakob Verbeek, Holger Schwenk, and Matthieu Cord,
  `Diffedit: Diffusion-based semantic image editing with mask guidance', {\em
  arXiv preprint arXiv:2210.11427}, (2022).

\bibitem{croitoru2023diffusion}
Florinel-Alin Croitoru, Vlad Hondru, Radu~Tudor Ionescu, and Mubarak Shah,
  `Diffusion models in vision: A survey', {\em IEEE TPAMI}, (2023).

\bibitem{daniels2021score}
Max Daniels, Tyler Maunu, and Paul Hand, `Score-based generative neural
  networks for large-scale optimal transport', {\em {NeurIPS}}, {\bf 34},
  12955--12965, (2021).

\bibitem{dhariwal2021diffusion}
Prafulla Dhariwal and Alexander Nichol, `Diffusion models beat gans on image
  synthesis', {\em {NeurIPS}}, {\bf 34},  8780--8794, (2021).

\bibitem{Dong2023a}
Bo~Dong, Jialun Pei, Rongrong Gao, Tian-Zhu Xiang, Shuo Wang, and Huan Xiong,
  `A unified query-based paradigm for camouflaged instance segmentation', in
  {\em {ACM} {MM}}, (2023).

\bibitem{fan2021concealed}
Deng-Ping Fan, Ge-Peng Ji, Ming-Ming Cheng, and Ling Shao, `Concealed object
  detection', {\em IEEE TPAMI}, (2021).

\bibitem{fan2020camouflaged}
Deng-Ping Fan, Ge-Peng Ji, Guolei Sun, Ming-Ming Cheng, Jianbing Shen, and Ling
  Shao, `Camouflaged object detection', in {\em CVPR}, pp. 2777--2787, (2020).

\bibitem{fan2023advances}
Deng-Ping Fan, Ge-Peng Ji, Peng Xu, Ming-Ming Cheng, Christos Sakaridis, and
  Luc Van~Gool, `Advances in deep concealed scene understanding', {\em arXiv
  preprint arXiv:2304.11234}, (2023).

\bibitem{fan2020pranet}
Deng-Ping Fan, Ge-Peng Ji, Tao Zhou, Geng Chen, Huazhu Fu, Jianbing Shen, and
  Ling Shao, `Pranet: Parallel reverse attention network for polyp
  segmentation', in {\em MICCAI}, pp. 263--273. Springer, (2020).

\bibitem{fan2020inf}
Deng-Ping Fan, Tao Zhou, Ge-Peng Ji, Yi~Zhou, Geng Chen, Huazhu Fu, Jianbing
  Shen, and Ling Shao, `Inf-net: Automatic covid-19 lung infection segmentation
  from ct images', {\em IEEE Transactions on Medical Imaging}, {\bf 39}(8),
  2626--2637, (2020).

\bibitem{gu2022diffusioninst}
Zhangxuan Gu, Haoxing Chen, Zhuoer Xu, Jun Lan, Changhua Meng, and Weiqiang
  Wang, `Diffusioninst: Diffusion model for instance segmentation', {\em arXiv
  preprint arXiv:2212.02773}, (2022).

\bibitem{he2022weakly}
Ruozhen He, Qihua Dong, Jiaying Lin, and Rynson~WH Lau, `Weakly-supervised
  camouflaged object detection with scribble annotations', {\em AAAI},
  781--789, (2023).

\bibitem{hertz2022prompt}
Amir Hertz, Ron Mokady, Jay Tenenbaum, Kfir Aberman, Yael Pritch, and Daniel
  Cohen-Or, `Prompt-to-prompt image editing with cross attention control', {\em
  arXiv preprint arXiv:2208.01626}, (2022).

\bibitem{ho2020denoising}
Jonathan Ho, Ajay Jain, and Pieter Abbeel, `Denoising diffusion probabilistic
  models', {\em {NeurIPS}}, {\bf 33},  6840--6851, (2020).

\bibitem{ho2022video}
Jonathan Ho, Tim Salimans, Alexey Gritsenko, William Chan, Mohammad Norouzi,
  and David~J Fleet, `Video diffusion models', {\em arXiv preprint
  arXiv:2204.03458}, (2022).

\bibitem{hu2022high}
Xiaobin Hu, Shuo Wang, Xuebin Qin, Hang Dai, Wenqi Ren, Donghao Luo, Ying Tai,
  and Ling Shao, `High-resolution iterative feedback network for camouflaged
  object detection', {\em AAAI}, (2023).

\bibitem{huang2023feature}
Zhou Huang, Hang Dai, Tian-Zhu Xiang, Shuo Wang, Huai-Xin Chen, Jie Qin, and
  Huan Xiong, `Feature shrinkage pyramid for camouflaged object detection with
  transformers', {\em CVPR}, (2023).

\bibitem{ji2022gradient}
Ge-Peng Ji, Deng-Ping Fan, Yu-Cheng Chou, Dengxin Dai, Alexander Liniger, and
  Luc Van~Gool, `Deep gradient learning for efficient camouflaged object
  detection', {\em Machine Intelligence Research}, (2023).

\bibitem{ji2022fast}
Ge-Peng Ji, Lei Zhu, Mingchen Zhuge, and Keren Fu, `Fast camouflaged object
  detection via edge-based reversible re-calibration network', {\em Pattern
  Recognition}, {\bf 123},  108414, (2022).

\bibitem{ji2023ddp}
Yuanfeng Ji, Zhe Chen, Enze Xie, Lanqing Hong, Xihui Liu, Zhaoqiang Liu, Tong
  Lu, Zhenguo Li, and Ping Luo, `Ddp: Diffusion model for dense visual
  prediction', {\em arXiv preprint arXiv:2303.17559}, (2023).

\bibitem{jia2022segment}
Qi~Jia, Shuilian Yao, Yu~Liu, Xin Fan, Risheng Liu, and Zhongxuan Luo,
  `Segment, magnify and reiterate: Detecting camouflaged objects the hard way',
  in {\em CVPR}, pp. 4713--4722, (2022).

\bibitem{jia2023taming}
Xuhui Jia, Yang Zhao, Kelvin~CK Chan, Yandong Li, Han Zhang, Boqing Gong,
  Tingbo Hou, Huisheng Wang, and Yu-Chuan Su, `Taming encoder for zero
  fine-tuning image customization with text-to-image diffusion models', {\em
  arXiv preprint arXiv:2304.02642}, (2023).

\bibitem{karras2022elucidating}
Tero Karras, Miika Aittala, Timo Aila, and Samuli Laine, `Elucidating the
  design space of diffusion-based generative models', {\em arXiv preprint
  arXiv:2206.00364}, (2022).

\bibitem{kumar2021early}
Karthika~Suresh Kumar and Aamer Abdul~Rahman, `Early detection of locust swarms
  using deep learning', in {\em Advances in machine learning and computational
  intelligence},  303--310, Springer, (2021).

\bibitem{le2019anabranch}
Trung-Nghia Le, Tam~V Nguyen, Zhongliang Nie, Minh-Triet Tran, and Akihiro
  Sugimoto, `Anabranch network for camouflaged object segmentation', {\em
  CVIU}, {\bf 184},  45--56, (2019).

\bibitem{li2021uncertainty}
Aixuan Li, Jing Zhang, Yunqiu Lv, Bowen Liu, Tong Zhang, and Yuchao Dai,
  `Uncertainty-aware joint salient object and camouflaged object detection', in
  {\em CVPR}, pp. 10071--10081, (2021).

\bibitem{li2022srdiff}
Haoying Li, Yifan Yang, Meng Chang, Shiqi Chen, Huajun Feng, Zhihai Xu, Qi~Li,
  and Yueting Chen, `Srdiff: Single image super-resolution with diffusion
  probabilistic models', {\em Neurocomputing}, {\bf 479},  47--59, (2022).

\bibitem{li2022trichomonas}
Lin Li, Jingyi Liu, Shuo Wang, Xunkun Wang, and Tian-Zhu Xiang, `Trichomonas
  vaginalis segmentation in microscope images', in {\em MICCAI}, pp. 68--78.
  Springer, (2022).

\bibitem{li2021mvdi25k}
Lin Li, Jingyi Liu, Fei Yu, Xunkun Wang, and Tian-Zhu Xiang, `Mvdi25k: A
  large-scale dataset of microscopic vaginal discharge images', {\em
  BenchCouncil Transactions on Benchmarks, Standards and Evaluations}, {\bf
  1}(1),  100008, (2021).

\bibitem{lv2021simultaneously}
Yunqiu Lv, Jing Zhang, Yuchao Dai, Aixuan Li, Bowen Liu, Nick Barnes, and
  Deng-Ping Fan, `Simultaneously localize, segment and rank the camouflaged
  objects', in {\em CVPR}, pp. 11591--11601, (2021).

\bibitem{mei2021camouflaged}
Haiyang Mei, Ge-Peng Ji, Ziqi Wei, Xin Yang, Xiaopeng Wei, and Deng-Ping Fan,
  `Camouflaged object segmentation with distraction mining', in {\em CVPR}, pp.
  8772--8781, (2021).

\bibitem{meng2021sdedit}
Chenlin Meng, Yang Song, Jiaming Song, Jiajun Wu, Jun-Yan Zhu, and Stefano
  Ermon, `Sdedit: Image synthesis and editing with stochastic differential
  equations', {\em arXiv preprint arXiv:2108.01073}, (2021).

\bibitem{nichol2021improved}
Alexander~Quinn Nichol and Prafulla Dhariwal, `Improved denoising diffusion
  probabilistic models', in {\em ICML}, pp. 8162--8171, (2021).

\bibitem{pang2022zoom}
Youwei Pang, Xiaoqi Zhao, Tian-Zhu Xiang, Lihe Zhang, and Huchuan Lu, `Zoom in
  and out: A mixed-scale triplet network for camouflaged object detection', in
  {\em CVPR}, pp. 2160--2170, (2022).

\bibitem{pang2020multi}
Youwei Pang, Xiaoqi Zhao, Lihe Zhang, and Huchuan Lu, `Multi-scale interactive
  network for salient object detection', in {\em CVPR}, pp. 9413--9422, (2020).

\bibitem{prakash2021multi}
Aditya Prakash, Kashyap Chitta, and Andreas Geiger, `Multi-modal fusion
  transformer for end-to-end autonomous driving', in {\em CVPR}, pp.
  7077--7087, (2021).

\bibitem{aimon2023ambiguous}
Aimon Rahman, Jeya Maria~Jose Valanarasu, Ilker Hacihaliloglu, and Vishal~M
  Patel, `Ambiguous medical image segmentation using diffusion models', in {\em
  CVPR}, (2023).

\bibitem{ronneberger2015u}
Olaf Ronneberger, Philipp Fischer, and Thomas Brox, `U-net: Convolutional
  networks for biomedical image segmentation', in {\em MICCAI}, pp. 234--241.
  Springer, (2015).

\bibitem{skurowski2018animal}
Przemys{\l}aw Skurowski, Hassan Abdulameer, J~B{\l}aszczyk, Tomasz Depta, Adam
  Kornacki, and P~Kozie{\l}, `Animal camouflage analysis: Chameleon database',
  {\em Unpublished manuscript}, {\bf 2}(6), ~7, (2018).

\bibitem{sohl2015deep}
Jascha Sohl-Dickstein, Eric Weiss, Niru Maheswaranathan, and Surya Ganguli,
  `Deep unsupervised learning using nonequilibrium thermodynamics', in {\em
  International Conference on Machine Learning}, pp. 2256--2265. PMLR, (2015).

\bibitem{song2019generative}
Yang Song and Stefano Ermon, `Generative modeling by estimating gradients of
  the data distribution', {\em {NeurIPS}}, {\bf 32}, (2019).

\bibitem{song2020score}
Yang Song, Jascha Sohl-Dickstein, Diederik~P Kingma, Abhishek Kumar, Stefano
  Ermon, and Ben Poole, `Score-based generative modeling through stochastic
  differential equations', {\em arXiv preprint arXiv:2011.13456}, (2020).

\bibitem{sun2021context}
Yujia Sun, Geng Chen, Tao Zhou, Yi~Zhang, and Nian Liu, `Context-aware
  cross-level fusion network for camouflaged object detection', {\em IJCAI},
  1025--1031, (2021).

\bibitem{sun2022boundary}
Yujia Sun, Shuo Wang, Chenglizhao Chen, and Tian-Zhu Xiang, `Boundary-guided
  camouflaged object detection', {\em IJCAI},  1335--1341, (2022).

\bibitem{tabernik2020segmentation}
Domen Tabernik, Samo {\v{S}}ela, Jure Skvar{\v{c}}, and Danijel Sko{\v{c}}aj,
  `Segmentation-based deep-learning approach for surface-defect detection',
  {\em Journal of Intelligent Manufacturing}, {\bf 31}(3),  759--776, (2020).

\bibitem{wang2021pvtv2}
Wenhai Wang, Enze Xie, Xiang Li, Deng-Ping Fan, Kaitao Song, Ding Liang, Tong
  Lu, Ping Luo, and Ling Shao, `Pvtv2: Improved baselines with pyramid vision
  transformer', {\em Computational Visual Media}, {\bf 8}(3),  1--10, (2022).

\bibitem{wu2022medsegdiff}
Junde Wu, Huihui Fang, Yu~Zhang, Yehui Yang, and Yanwu Xu, `Medsegdiff: Medical
  image segmentation with diffusion probabilistic model', {\em MIDL}, (2023).

\bibitem{wu2023medsegdiff}
Junde Wu, Rao Fu, Huihui Fang, Yu~Zhang, and Yanwu Xu, `Medsegdiff-v2:
  Diffusion based medical image segmentation with transformer', {\em arXiv
  preprint arXiv:2301.11798}, (2023).

\bibitem{wu2023diffumask}
Weijia Wu, Yuzhong Zhao, Mike~Zheng Shou, Hong Zhou, and Chunhua Shen,
  `Diffumask: Synthesizing images with pixel-level annotations for semantic
  segmentation using diffusion models', {\em ICCV}, (2023).

\bibitem{wu2019cascaded}
Zhe Wu, Li~Su, and Qingming Huang, `Cascaded partial decoder for fast and
  accurate salient object detection', in {\em CVPR}, pp. 3907--3916, (2019).

\bibitem{xu2023open}
Jiarui Xu, Sifei Liu, Arash Vahdat, Wonmin Byeon, Xiaolong Wang, and Shalini
  De~Mello, `Open-vocabulary panoptic segmentation with text-to-image diffusion
  models', {\em CVPR}, (2023).

\bibitem{yin2022camoformer}
Bowen Yin, Xuying Zhang, Qibin Hou, Bo-Yuan Sun, Deng-Ping Fan, and Luc
  Van~Gool, `Camoformer: Masked separable attention for camouflaged object
  detection', {\em arXiv preprint arXiv:2212.06570}, (2022).

\bibitem{zhai2021mutual}
Qiang Zhai, Xin Li, Fan Yang, Chenglizhao Chen, Hong Cheng, and Deng-Ping Fan,
  `Mutual graph learning for camouflaged object detection', in {\em CVPR}, pp.
  12997--13007, (2021).

\bibitem{zhang2022camouflaged}
Cong Zhang, Kang Wang, Hongbo Bi, Ziqi Liu, and Lina Yang, `Camouflaged object
  detection via neighbor connection and hierarchical information transfer',
  {\em CVIU}, {\bf 221},  103450, (2022).

\bibitem{zhang2022preynet}
Miao Zhang, Shuang Xu, Yongri Piao, Dongxiang Shi, Shusen Lin, and Huchuan Lu,
  `Preynet: Preying on camouflaged objects', in {\em ACM MM}, pp. 5323--5332,
  (2022).

\bibitem{Zhao_2019_ICCV}
Jia-Xing Zhao, Jiang-Jiang Liu, Deng-Ping Fan, Yang Cao, Jufeng Yang, and
  Ming-Ming Cheng, `Egnet: Edge guidance network for salient object detection',
  in {\em ICCV}, pp. 8778--8787, (October 2019).

\bibitem{zhong2022detecting}
Yijie Zhong, Bo~Li, Lv~Tang, Senyun Kuang, Shuang Wu, and Shouhong Ding,
  `Detecting camouflaged object in frequency domain', in {\em CVPR}, pp.
  4504--4513, (2022).

\bibitem{zhu2022can}
Hongwei Zhu, Peng Li, Haoran Xie, Xuefeng Yan, Dong Liang, Dapeng Chen,
  Mingqiang Wei, and Jing Qin, `I can find you! boundary-guided separated
  attention network for camouflaged object detection', in {\em AAAI}, pp.
  3608--3616, (2022).

\bibitem{zhu2021inferring}
Jinchao Zhu, Xiaoyu Zhang, Shuo Zhang, and Junnan Liu, `Inferring camouflaged
  objects by texture-aware interactive guidance network', in {\em AAAI}, pp.
  3599--3607, (2021).

\bibitem{zhuge2022cubenet}
Mingchen Zhuge, Xiankai Lu, Yiyou Guo, Zhihua Cai, and Shuhan Chen, `Cubenet:
  X-shape connection for camouflaged object detection', {\em Pattern
  Recognition}, {\bf 127},  108644, (2022).

\end{thebibliography}
\end{document}


\begin{frontmatter}

\title{Supplementary Material: \\Diffusion Model for Camouflaged Object Detection}
\author{Anonymous Submission}

\end{frontmatter}

\newpage
\bibliography{ecai}